\documentclass{IEEEtran}

\usepackage{graphicx}
\graphicspath{ {./Figures/} }
\usepackage{placeins}
\usepackage{cuted}
\usepackage{caption}
\usepackage{graphics} % for pdf, bitmapped graphics files
\usepackage{epsfig} % for postscript graphics files
\usepackage{mathptmx} % assumes new font selection scheme installed
\usepackage{times} % assumes new font selection scheme installed
\usepackage{amsmath} % assumes amsmath package installed
\usepackage{booktabs}
\usepackage{tabularx}
\usepackage{siunitx} % if you want nice units; optional

\usepackage[
backend=bibtex,
style=ieee,
sorting=none,
citestyle=numeric-comp
]{biblatex}
\AtBeginBibliography{\footnotesize}

\usepackage{amssymb}  % assumes amsmath package installed
\usepackage{bm}
\usepackage{color}
\usepackage{graphicx}
\usepackage{graphics} % for pdf, bitmapped graphics files

\usepackage{amssymb}
\usepackage{wrapfig}
\usepackage{multirow}
\usepackage{tabularx,booktabs}
\usepackage{algorithm} 
\usepackage{algpseudocode} 
\usepackage{chngcntr}

\newcolumntype{C}{>{\centering\arraybackslash}X} 
\usepackage{lipsum} % filler text
\newcommand{\rev}[1]{\textcolor{black}{#1}}

\newcommand{\mrev}[1]{\textcolor{black}{#1}}
\usepackage{mdframed}
\usepackage{float}
\usepackage{amsthm}
\newtheorem{theorem}{Theorem}
\newtheorem{remarknn}{Remark}
\newtheorem{assumption}{Assumption}
\newtheorem{lemma}{Lemma}

\ifCLASSOPTIONcompsoc
\else
\fi
\ifCLASSINFOpdf
\else
\fi
	\usepackage[table,xcdraw,dvipsnames]{xcolor} % enables color names like 'teal', 'RoyalBlue', etc.
	\usepackage[colorlinks=true,
	linkcolor=blue,
	citecolor=blue,
	urlcolor=magenta]{hyperref}
	\usepackage{xr-hyper}
	\usepackage[nameinlink,capitalize,noabbrev]{cleveref} % optional but nice

\begin{document}
		%
		% paper title
		% Titles are generally capitalized except for words such as a, an, and, as,
		% at, but, by, for, in, nor, of, on, or, the, to and up, which are usually
		% not capitalized unless they are the first or last word of the title.
		% Linebreaks \\ can be used within to get better formatting as desired.
		% Do not put math or special symbols in the title.
		\title{Curriculum-Adapted Robust Reinforcement Learning for UAV Deconfliction in Adversarial Environments}
		%
		%
		% author names and IEEE memberships
		% note positions of commas and nonbreaking spaces ( ~ ) LaTeX will not break
		% a structure at a ~ so this keeps an author's name from being broken across
		% two lines.
		% use \thanks{} to gain access to the first footnote area
		% a separate \thanks must be used for each paragraph as LaTeX2e's \thanks
		% was not built to handle multiple paragraphs
		%
		%
		%\IEEEcompsocitemizethanks is a special \thanks that produces the bulleted
		% lists the Computer Society journals use for "first footnote" author
		% affiliations. Use \IEEEcompsocthanksitem which works much like \item
		% for each affiliation group. When not in compsoc mode,
		% \IEEEcompsocitemizethanks becomes like \thanks and
		% \IEEEcompsocthanksitem becomes a line break with idention. This
		% facilitates dual compilation, although admittedly the differences in the
		% desired content of \author between the different types of papers makes a
		% one-size-fits-all approach a daunting prospect. For instance, compsoc 
		% journal papers have the author affiliations above the "Manuscript
		% received ..."  text while in non-compsoc journals this is reversed. Sigh.
		
		\author{Deepak Kumar Panda, Adolfo Perrusqu\'ia and Weisi Guo% <-this % stops a space
			\thanks{This work was supported by the Royal Academy of Engineering and the Office of the Chief Science Adviser for National Security under the UK Intelligence Community Postdoctoral Research Fellowship programme}% <-this % stops a space
			\thanks{$^{1}$D. Kumar Panda, A. Perrusqu\'ia and W. Guo are with the Faculty of Engineering and Applied Sciences,
				Cranfield University, MK43 0AL Cranfield, U.K
				{\tt\small Deepak.Panda@cranfield.ac.uk, Adolfo.Perrusquia-Guzman@cranfield.ac.uk, weisi.guo@cranfield.ac.uk}}%
		}
		
		% note the % following the last \IEEEmembership and also \thanks - 
		% these prevent an unwanted space from occurring between the last author name
		% and the end of the author line. i.e., if you had this:
		% 
		% \author{....lastname \thanks{...} \thanks{...} }
		%                     ^------------^------------^----Do not want these spaces!
		%
		% a space would be appended to the last name and could cause every name on that
		% line to be shifted left slightly. This is one of those "LaTeX things". For
		% instance, "\textbf{A} \textbf{B}" will typeset as "A B" not "AB". To get
		% "AB" then you have to do: "\textbf{A}\textbf{B}"
		% \thanks is no different in this regard, so shield the last } of each \thanks
	% that ends a line with a % and do not let a space in before the next \thanks.
	% Spaces after \IEEEmembership other than the last one are OK (and needed) as
	% you are supposed to have spaces between the names. For what it is worth,
	% this is a minor point as most people would not even notice if the said evil
	% space somehow managed to creep in.

	% The paper headers
	\markboth{Journal of \LaTeX\ Class Files,~Vol.~14, No.~8, August~2015}%
	{Shell \MakeLowercase{\textit{et al.}}: Bare Advanced Demo of IEEEtran.cls for IEEE Computer Society Journals}
	% The only time the second header will appear is for the odd numbered pages
	% after the title page when using the twoside option.
	% 
	% *** Note that you probably will NOT want to include the author's ***
	% *** name in the headers of peer review papers.                   ***
	% You can use \ifCLASSOPTIONpeerreview for conditional compilation here if
	% you desire.

	% The publisher's ID mark at the bottom of the page is less important with
	% Computer Society journal papers as those publications place the marks
	% outside of the main text columns and, therefore, unlike regular IEEE
	% journals, the available text space is not reduced by their presence.
	% If you want to put a publisher's ID mark on the page you can do it like
	% this:
	%\IEEEpubid{0000--0000/00\$00.00~\copyright~2015 IEEE}
	% or like this to get the Computer Society new two part style.
	%\IEEEpubid{\makebox[\columnwidth]{\hfill 0000--0000/00/\$00.00~\copyright~2015 IEEE}%
		%\hspace{\columnsep}\makebox[\columnwidth]{Published by the IEEE Computer Society\hfill}}
	% Remember, if you use this you must call \IEEEpubidadjcol in the second
	% column for its text to clear the IEEEpubid mark (Computer Society journal
	% papers don't need this extra clearance.)

	% use for special paper notices
	%\IEEEspecialpapernotice{(Invited Paper)}

	% for Computer Society papers, we must declare the abstract and index terms
	% PRIOR to the title within the \IEEEtitleabstractindextext IEEEtran
	% command as these need to go into the title area created by \maketitle.
	% As a general rule, do not put math, special symbols or citations
	% in the abstract or keywords.
	\IEEEtitleabstractindextext{%
		\begin{abstract}
		Autonomous unmanned aerial vehicles (UAVs) increasingly rely on reinforcement learning (RL) for navigation. However, global navigation satellite system (GNSS) spoofing attacks can induce out-of-distribution observation shifts that corrupt value estimation and degrade mission performance. Existing robust RL approaches typically improve resilience against specific attack models but often fail to generalize to attacks not encountered during training. To address this limitation, we propose a curriculum-guided adaptation framework that progressively exposes a robust policy to gradient-based adversarial observation perturbations of increasing intensity while aligning temporal-difference (TD) error distributions across curriculum stages. Rather than adapting to a particular attack model, the proposed approach preserves TD-error consistency to promote transferability across attack conditions. We further derive a TD-space generalization certificate showing that if the TD-error distribution induced by a test-time attack remains sufficiently close to that of the final curriculum stage, the resulting performance degradation is bounded. The framework is evaluated in a UAV deconfliction environment with dynamic 3D obstacles under previously unseen fixed and dynamic GNSS spoofing attacks. \mrev{Under fixed spoofing conditions, the curriculum-adapted (CurAd) policy maintained approximately 99-100\% mission success across the evaluated aerial-traffic densities, compared with approximately 20-24\% success rates obtained from the best-performing robust baseline, while non-robust vanilla baseline achieved higher but considerably lower than CurAd success rates. Under dynamic obstacle-luring spoofing attacks, it achieved the highest episodic rewards while reducing the median mission-completion steps by up to approximately 22\% relative to the best performing robust-RL baseline.} These results indicate that adversarial curriculum adaptation can transfer resilience from synthetic adversarial perturbations to realistic GNSS spoofing attacks without attack-specific retraining, improving autonomous navigation under previously unseen attack conditions.
		\end{abstract}
		
		% Note that keywords are not normally used for peerreview papers.
		}

	% make the title area
	\maketitle

	% To allow for easy dual compilation without having to reenter the
	% abstract/keywords data, the \IEEEtitleabstractindextext text will
	% not be used in maketitle, but will appear (i.e., to be "transported")
	% here as \IEEEdisplaynontitleabstractindextext when compsoc mode
	% is not selected <OR> if conference mode is selected - because compsoc
	% conference papers position the abstract like regular (non-compsoc)
	% papers do!
	\IEEEdisplaynontitleabstractindextext
	% \IEEEdisplaynontitleabstractindextext has no effect when using
	% compsoc under a non-conference mode.

	% For peer review papers, you can put extra information on the cover
	% page as needed:
	% \ifCLASSOPTIONpeerreview
	% \begin{center} \bfseries EDICS Category: 3-BBND \end{center}
	% \fi
	%
	% For peerreview papers, this IEEEtran command inserts a page break and
	% creates the second title. It will be ignored for other modes.
	\IEEEpeerreviewmaketitle

	\ifCLASSOPTIONcompsoc
	\IEEEraisesectionheading{\section{Introduction}\label{sec:introduction}}
	\else
	\label{sec:introduction}
	\fi
	Unmanned aerial vehicles (UAVs) are increasingly deployed across  civilian domains such as logistics, surveillance, and disaster response \cite{perrusquia2024uncovering}. \rev{Reinforcement learning (RL) has been extensively used for UAV path-planning in complex and dynamic environments \cite{razzaghi2024survey}}. However, they face significant security challenges such as GNSS spoofing and sensor falsification \cite{kim2017ads, ying2019detecting, panda2024action}, which cause \rev{unexpected} distribution shift in the observation-space.  
	
	\rev{Early studies employed deep Q-networks (DQN) and double DQN for tactical collision avoidance of the UAVs in dense airspace, while actor-critic variants such as proximal policy optimization (PPO), deep deterministic policy gradient (DDPG), and soft actor–critic (SAC) demonstrated improved safety in continuous control settings by learning optimal heading and altitude maneuvers in real time among cooperative and non-cooperative UAVs within urban air mobility environments \cite{razzaghi2024survey}. However, these navigation policies tend to become fragile under adversarial observation-space shifts \cite{ilahi2021challenges}.} Recent work has primarily addressed action-space noise or benign domain shifts, neglecting the reality that adversaries can actively distort sensor inputs, such as GNSS spoofing \cite{ying2019detecting, panda2024action, leonardi2017ads}. As a result, RL policies often suffer from catastrophic forgetting, manifesting as value over-estimation leading to unsafe behaviors in autonomous systems.
	
	To address this critical gap, we present a \textit{curriculum adaptation framework} (\mrev{CurAd}), against  adversarial observations of increasing strength \rev{for improved test-time generalization}.  Our adaptation therefore operates in TD-space: TD-error distributions are aligned across curriculum stages, which improves generalization to unseen spoofing attacks. We introduce a formal definition of catastrophic forgetting under adversarial conditions, quantifying value function degradation using the Wasserstein-1 distance between temporal-difference (TD) error distributions. This enables both theoretical characterization and practical mitigation of fragility in RL-based UAV navigation, moving beyond conventional robust RL to provide empirically and provably robust adaptation in cyber-physical systems.
	\subsection{Related Works}
	\begin{table*}[htbp] 
		\centering 
		\caption{Comparison with existing approaches and research gaps addressed by the proposed framework.} 
		\begin{tabularx}{\textwidth}{c|l|X|X} 
			\hline 
			\textbf{\rev{No.}} & 
			\rev{\textbf{Existing Methods}} & 
			\rev{\textbf{Description}} & 
			\rev{\textbf{Research Gap Addressed in the Paper}} \\ 
			\hline 
			
			\rev{1} & 
			\rev{Robust RL \cite{shi2024distributionally, tessler2019action}} &  
			\rev{Hardens policies against adversarial disturbances using uncertainty sets, adversarial training, or worst-case optimization.} & 
			\rev{Typically assumes fixed or bounded disturbances and does not account for TD-value drift under unseen attacks, which can lead to catastrophic forgetting. Moreover, no test-time robustness certificates are provided.} \\ 
			\hline 
			
			\rev{2} & 
			\rev{Continual RL \cite{nayyar2025autonomous, rostami2021lifelong}} & 
			\rev{Adapts policies to evolving environments while mitigating catastrophic forgetting across tasks.} & 
			\rev{Primarily focuses on task transitions rather than maintaining value-distribution consistency under adversarial attacks.} \\ 
			\hline 
			
			\rev{3} & 
			\rev{Existing Curriculum RL and Unsupervised Domain Adaptation Techniques \cite{huang2022curriculum, wang2023curriculum}} & 
			\rev{Improves learning by progressively increasing task difficulty while incorporating distribution alignment.} & 
			\rev{Does not provide generalization guarantees against unseen domains or adversarial attacks.} \\ 
			\hline 
			
			\mrev{4} & 
			\mrev{Model-based safe RL \cite{berkenkamp2017safe, turchetta2016safe}} & 
			\mrev{Learns dynamics and safety models with uncertainty estimates to support safe exploration.} & 
			\mrev{Primarily addresses model uncertainty and safety constraints but does not directly characterize robustness against unseen adversarial attacks.} \\ 
			\hline 
			
			\rev{5} & 
			\rev{Existing GNSS Spoofing Attack Defenses \cite{dasgupta2022sensor, psiaki2016gnss}} & 
			\rev{Detects or mitigates spoofing using sensor redundancy, filtering, or signal-level analysis.} & 
			\rev{Often attack-specific and reliant on predefined spoofing signatures or additional sensing modalities.} \\ 
			\hline 
		\end{tabularx} 
		\label{tab:research_gap} 
	\end{table*}
	\subsubsection{Existing Defense against Spoofing Attacks}
	Existing GNSS spoofing countermeasures primarily focus on signal authentication and anomaly detection \cite{psiaki2016gnss}, sensor redundancy, and cross-modal consistency checking between GNSS, inertial, and vision-based measurements \cite{dasgupta2022sensor}. While these approaches improve attack detection and state-estimation accuracy, they remain largely detection-centric and often rely on predefined attack signatures or alternative sensing modalities. Consequently, relatively little attention has been devoted to maintaining decision-making performance once corrupted observations have propagated into the autonomy stack. This limitation motivates the development of learning-based approaches that can adapt to adversarially induced distribution shifts and maintain robust navigation performance under spoofed operating conditions without compromising the mission efficiency.
	
	\subsubsection{Robust Reinforcement Learning}
	Robust RL has received significant attention as a means to improve the resilience of autonomous RL agents against adversarial attacks and environmental uncertainty. Existing approaches employ adversarial training, uncertainty-set formulations, distributionally robust optimization, and action-robust learning to improve safety in applications such as autonomous navigation and air mobility \cite{zhang2020robust,tessler2019action,panda2024action}. Recent works have extended robust RL to online, offline, and corruption-aware settings, providing stronger theoretical guarantees under predefined uncertainty models through total-variation, Wasserstein, and KL-divergence ambiguity sets \cite{lu2024distributionally,clavier2024near,shi2024distributionally,pollatos2025corruption}. These advances have significantly improved robustness against disturbances drawn from the uncertainty families considered during training. However, existing research mostly evaluate robustness within the same perturbation model used for policy training and provide limited insight into whether robustness can transfer across fundamentally different perturbation mechanisms. As attack characteristics change, policies may experience shifts in value estimation that lead to performance degradation and catastrophic forgetting \cite{korbak2022reinforcement}. This naturally connects to the objectives of continual RL, which addresses adaptation and knowledge retention in non-stationary environments.
	
	\subsubsection{\mrev{Model-Based Safe Reinforcement Learning}}
	\mrev{Model-based safe RL provides a complementary approach to reliable autonomous control by learning predictive dynamics or safety models and incorporating their uncertainty during policy learning and planning. GP-based approaches combine learned dynamics with Lyapunov analysis to provide probabilistic stability certificates, while safe-exploration methods use GP uncertainty to characterize safely reachable regions of the state space \cite{berkenkamp2017safe,turchetta2016safe}. More recent model-based approaches incorporate uncertainty-aware reachability analysis to account for errors in learned dynamics models \cite{yu2023safe}. These approaches primarily address reliability under model uncertainty and explicit safety constraints. However, uncertainty bounds on learned dynamics do not directly characterize adversarial observation-space distribution shifts, where degraded sensor data due to adversarial attack can alter the observations available for decision-making without necessarily changing the underlying physical dynamics. Consequently, robustness transfer under previously unseen sensor-induced attack distributions remains a distinct reliability challenge.}
	
	\subsubsection{Continual Learning in Reinforcement Learning}
	In contrast to conventional RL frameworks that assume stationary Markov decision processes and evaluate agents based on steady-state performance, continual RL focuses on adaptation in non-stationary environments. Recent work has reformulated RL for continual settings using history-process formalisms and deviation-regret metrics to capture how agent performance changes over time \cite{elelimy2025rethinking}. A central challenge in this setting is catastrophic forgetting, where policies lose previously acquired knowledge when exposed to new tasks. Early approaches such as progressive neural networks used architectural isolation to reduce interference between tasks \cite{rusu2016progressive,khetarpal2022towards}, while subsequent research have explored pseudo-replay, neuron restoration, pre-trained encoders, compositional policies, and hierarchical option reuse to improve knowledge retention and forward transfer \cite{rostami2021transfer,xu2020forget,malagon2025self,nayyar2025autonomous}. These advances provide important mechanisms for learning under non-stationarity observation space, but they are primarily designed for task-level changes, where transitions between domains are either explicit or can be treated as changes in task structure. In adversarial UAV navigation, the challenge is different: spoofing attacks may induce gradual and previously unseen distribution shifts without clear task boundaries, while corrupting the value estimates used for downstream decision-making. Existing continual RL methods therefore provide limited insight into how value-function consistency should be preserved when the source of non-stationarity is malicious, sensor-induced, and potentially different from the perturbations encountered during training. 
	
	\subsubsection{Curriculum Reinforcement Learning and Unsupervised Domain Adaptation}
	Curriculum RL provides a structured approach for training agents in non-stationary environments by exposing them to a sequence of tasks with gradually increasing difficulty \cite{klink2024benefit}. Rather than learning directly on the target task, curriculum strategies guide the agent through intermediate tasks that improve learning efficiency, stability, and generalization. Recent work has formalized curriculum generation through optimal-transport formulations, where Wasserstein distances and barycentric interpolation are used to construct smooth transitions between auxiliary and target task distributions \cite{huang2022curriculum}. Other studies have developed multi-stage curricula for robot navigation, progressively increasing environmental complexity from static obstacles to scenarios involving moving obstacles \cite{wang2023curriculum}. Similarly, hierarchical curriculum frameworks learn reusable skills, options, and state abstractions that facilitate transfer across long-horizon tasks and improve scalability in sparse-reward settings \cite{nayyar2025autonomous}. Despite these advances, existing curriculum RL methods primarily focus on designing learning trajectories that improve sample efficiency, exploration, and final task performance, with the aim of solving difficult tasks within a common task family.  Much less attention has been devoted to understanding whether robustness acquired through such curricula transfers beyond the training distribution to fundamentally different perturbation encountered during deployment. 
	
	A related line of research is unsupervised domain adaptation (UDA), which seeks to transfer knowledge across domains with different data distributions without requiring labelled target-domain data \cite{ganin2015unsupervised}. Existing UDA approaches typically reduce distribution mismatch by learning domain-invariant representations, aligning latent feature distributions, or minimizing divergence-based metrics such as maximum mean discrepancy (MMD) and Wasserstein distance \cite{sankaranarayanan2018generate,long2018conditional,huang2022curriculum}. Similar ideas have been explored in RL, where shared latent representations are used to facilitate policy transfer across domains \cite{xing2021domain}. Although UDA methods can successfully align latent feature distributions across domains, distributional similarity in representation space does not necessarily translate into similarity in value functions or policy behaviour. This distinction is particularly important in sequential decision-making problems, where small discrepancies in value estimation can accumulate over time and lead to substantial performance degradation. As a result, existing UDA frameworks provide limited insight into how robustness should be transferred across adversarial domains while preserving decision-making performance within a given acceptable limit.

	\subsection{Research Gap and Contributions}
	\subsubsection{Research Gap}
	Despite recent progress, critical gaps remain in enabling autonomous agents to maintain reliable decision-making under adversarially induced distribution shifts unseen during training as given in Table \ref{tab:research_gap}. We can summarize three key limitations in the existing state-of-the-art. \textit{First, robustness learned during training does not generalize to previously unseen attack mechanisms.} Robust RL often assume bounded disturbances and may exhibit substantial performance degradation when confronted with attack strategies that differ from those encountered during training. This limits the development of adaptation mechanisms capable of preserving decision-making performance under  attack conditions different from what observed during training. \textit{Second, existing curriculum learning and domain adaptation approaches provide limited guarantees for adversarial generalization.} Although curriculum-based training and distribution alignment have demonstrated improved performance under gradual domain shifts, they are primarily designed for benign distribution changes in the environment and do not establish formal conditions under which robustness can transfer across fundamentally different attack domains. In particular, there remains a lack of theoretical guarantees linking robustness acquired from synthetic adversarial perturbations to performance under previously unseen cyber-physical attacks.
	
	\subsubsection{Contributions}
	To address these challenges, we propose a curriculum-guided adaptation (\mrev{CurAD}) framework that aims to preserve policy performance under progressively changing adversarial conditions. The key idea is that while attacks may differ in how they corrupt observations, their impact is ultimately reflected in the value distribution space used by the agent for decision-making. Therefore, rather than adapting directly in the observation space, we focus on maintaining consistency in the TD-error distributions that characterize the agent learning process. Starting from a robust policy, the agent is progressively exposed to stronger adversarial perturbations through a curriculum-learning strategy. At each adaptation stage, the critic  from the previous stage acts as a reference, encouraging the agent to retain useful knowledge while adapting to the new perturbation regime. In the adaptation process, we aim to minimize the Wasserstein distance between TD-error distributions across consecutive stages, reducing value-function drift and mitigating catastrophic forgetting.
	
	Unlike conventional robust RL methods that seek robustness within a predefined uncertainty set and often require retraining for each new attack model, the proposed framework investigates whether \textbf{robustness acquired under one class of adversarial perturbations can generalize to previously unseen cyber-physical attack mechanisms without task-specific retraining}. To provide theoretical support for this capability, we derive a TD-space generalization certificate that establishes conditions under which robustness preserved through curriculum-guided adaptation transfers across attack domains. We further validate the proposed framework through extensive experiments that evaluate navigation performance under varying aerial traffic conditions and both fixed and dynamic GNSS spoofing strategies. Hence, the contributions of this paper are summarized as follows:
	\begin{itemize}
		\item \textbf{TD-space curriculum adaptation for RL:} We propose a novel adversarial adaptation framework that operates in the TD-error distribution space, motivated by the concept that adversarial perturbations ultimately influence policy behavior through their effect on value estimation. Leveraging this perspective, we develop a curriculum-guided adaptation \mrev{CurAd} strategy that progressively increases perturbation strength while enforcing consistency between TD-error distributions across adaptation stages. We additionally establish convergence guarantees showing that the TD-error distribution remains bounded throughout adaptation.
		\item \textbf{Generalization to spoofing attacks:} We establish a theoretical connection between TD-error distribution consistency and robustness transfer. Specifically, we derive a TD-space generalization certificate showing that when the TD-error distribution induced by a previously unseen attack remains sufficiently close to that of the final curriculum stage, the resulting performance degradation remains bounded. This provides a formal explanation for when robustness learned under one attack family can transfer to a different attack family.
		\item \textbf{Empirical validation in a UAV deconfliction against 3D obstacles under fixed and adaptive spoofing conditions:}
		We validate the proposed framework in a UAV deconfliction against 3D obstacles where adaptation is performed using gradient-based adversarial perturbations and evaluation is conducted under previously unseen fixed and dynamic GNSS spoofing attacks. The results demonstrate improved mission performance, reward retention, while minimizing the impact on the flight times compared with standard and robust RL baselines.
	\end{itemize}
The outline of the manuscript is as follows. Section \ref{Section II} describes the UAV deconfliction as a RL agent. Section \ref{section-III} introduces the RL action-robust action approach to obtain the expert critic for curriculum adaptation. In section \ref{section-IV} the curriculum adaptation strategy is defined along with the theoretical guarantees for TD-error bound across the curriculum \rev{along with test-time robustness guarantees for unseen adversarial attacks}. Section \ref{spoofing_evaluation} describes the nature of the fixed and dynamic spoofing attacks used to evaluate the curriculum guided adaptation strategy. Section \ref{section-V} reports on the simulation studies and the conclusions are presented in Section \ref{concl_section}.
In order to develop the above scheme, first we consider the UAV deconfliction for airspace environment as explained in the next section.
\section{UAV Deconfliction Environment}\label{Section II}
\subsection{Transition Model}
RL is used to plan point-to-point airspace navigation while deconflicting against 3D obstacle trajectories. In this context, UAV navigation is influenced by the kinematics and geometry of 3D obstacles, which generate a repulsive force that disrupts the attractive flow field guiding the UAV toward its target. Effective navigation requires balancing these opposing forces to ensure successful mission completion. The attractive and repulsive fields have been formulated as per \cite{zhang2021adaptive}.
First, the attractive field between the UAV and its target destination point is determined. Let the UAV's position vector be ${P} =\left ( x, y, z \right)$, and its corresponding velocity be denoted as ${v}({P})$. The velocity is computed as follows:
	\begin{equation} \label{eq:1} 
		{v}({P}) = -\left[ \frac{C(x - x_d)}{d({P}, {P}_g)}, \frac{C(y - y_d)}{d({P}, {P}_g)}, \frac{C(z - z_d)}{d({P}, {P}_g)} \right]^\top. \end{equation}
	Here, $C$ represents a scaling constant, ${P}_g = (x_d, y_d, z_d)$ is the target position, and $d({P}, {P}_g)$ denotes the Euclidean distance between the current position of the UAV and its target. Consider the velocity of an obstacle at the position of the UAV ${P}$, denoted by ${v}_{\text{obs}}\left ({P} \right )$. The resulting velocity of the UAV, influenced by the disturbance in the attractive flow field caused by the repulsive flow field of the obstacles expressed as
	\begin{equation} \label{eq:2}
		\bar{{v}}({P}) = \overline{M} \left ( \rho, \varsigma, \vartheta \right ) \left[ {v}({P}) - {v}_{\text{obs}}({P}) \right] + {v}_{\text{obs}}({P}),
	\end{equation}
	where, $\overline{M}(\rho, \varsigma, \vartheta)$ denotes the modulation function that characterizes the interaction between the desired velocity of the UAV toward the target destination and the obstacle-induced velocity. The parameters $\rho$, $\varsigma$ and $\vartheta$ govern the modulation behaviour as detailed in \cite{zhang2021adaptive} and explained in Appendix \ref{IFDS}. $\varsigma$ controls the repulsive force from obstacles, $\rho$ adjusts the attractive force toward the goal, and $\vartheta$ refers to the direction of the UAV.  The UAV position update during the time interval $\Delta T$, based on its velocity from (\ref{eq:2}) is given by:
	\begin{equation} \label{eq:3}
		{P}^{t+1} = {P}^{t} + \bar{{v}}({P}^{t}) \Delta T.
	\end{equation}
	The model in (\ref{eq:3}) acts as discrete-time transition model when an action $ \left ( \rho, \varsigma, \vartheta \right )$ is chosen from the RL policies, thus updating its position in the next time-step. The state and action-space description of the agent is provided in the next subsection.
	\subsection{States} The UAV, equipped with an RL agent for navigation, receives real-time information through automatic dependent surveillance-broadcast (ADS-B) \cite{ince2024sense} from other moving obstacles similar to a UAV with fixed trajectory required for tasks such as infrastructure inspection \cite{liu2022industrial}, road traffic monitoring, and surveillance \cite{huang2021decentralized}. The UAV obtains the position of the obstacle ${P}^t_{\textup{obs}}$ and the velocity ${v}_{\textup{obs}}^t$ at time $t$. The position of the UAV ${P}^t$ is assumed to be obtained from the built-in GNSS sensor. The agent input state at the time $t$ includes the UAV’s relative position to both the target and nearby obstacles, which is formally represented as
	\begin{equation} \label{eq:4}
		\Phi^t = \left\{ {P}_{\text{g}} - {P}^{t}, {P}^t_{\text{obs}} - {P}^{t}, {v}^t_{\text{obs}} \right\}.
	\end{equation}
	\subsection{Actions} 
	The actions of the UAV, as illustrated are represented by the vector $a_t = \left (\rho, \varsigma, \vartheta \right )$, where $\left (\rho, \varsigma \right )$ determine the influence of the repulsive force of the obstacles and the attractive force toward the target respectively, on UAV navigation and $\vartheta$ denotes the heading angle that the UAV must adjust to reach the designated destination.
	\subsection{Reward} \label{reward_section}  The agent reward for a given state $\Phi_t$ upon selecting the action $a_t$ is given by, 
	\begin{equation} \label{eq:5}
		R^t \left( a^t \mid \Phi^t \right) = \lambda_1 \cdot R_{\text{col}} + \lambda_2 \cdot R_{\text{path}} + \lambda_3 \cdot R_{\text{threat}}.
	\end{equation}
	where, $\lambda_1, \lambda_2, \lambda_3 > 0$ are weighing factors.  Each reward component $R_{\text{col}}$, $R_{\text{path}}$, and $R_{\text{threat}}$ is defined as follows:
	\begin{itemize}
		\item \textbf{Penalty reward} that penalizes the agent if the transitioned position ${P}^{t+1}$ falls within a certain radius of the dynamic obstacle, as represented by: 
		\begin{equation} \label{eq:6}
			R_{\text{col}} = -\frac{\left\|{P}^{t+1} - {P}_{\textup{obs}}\right\|}{\mathfrak{R}_{\textup{obs}}}, \quad \text{if} \left\|{P}^{t+1} - {P}_{\textup{obs}}\right\| \leq \mathfrak{R}_{\textup{obs}},
		\end{equation}
		where, $\mathfrak{R}_{\textup{obs}}$ represents the radius around the obstacle.
		\item \textbf{Path optimization reward:} that incentivizes path optimization, while considering the proximity to the goal and the obstacles: 
		\begin{equation} \label{eq:7}
			R_{\text{path}} = 
			\begin{cases} 
				-\frac{\| {P}^{t+1} - {P}_{g} \|}{\| {P}_{\text{obs}} - {P}_{g} \|}, & \text{if } \| {P}^{t+1} - {P}_{g} \| > \upsilon, \\
				-\frac{\| {P}^{t+1} - {P}_{g} \|}{\| {P}_{\text{obs}} - {P}_{g} \|} + C_1, & \text{if } \| {P}^{t+1} - {P}_{g} \| \leq \upsilon.
			\end{cases}
		\end{equation}
		In (\ref{eq:7}), $\upsilon$ denotes goal-proximity threshold used in path-optimization reward.
		\item \textbf{Obstacle threat penalty} imposing a penalty if the planned trajectory intersects a defined threat zone around an obstacle:
		\begin{equation} \label{eq:8}
			\begin{gathered}
				R_{\text{threat}} = \frac{\| {P}^{t+1} - {P}_{\text{obs}} - (\mathfrak{R}_{\text{obs}} + r) \|}{\mathfrak{R}_{\text{obs}}} - C_2,
				\quad \\ \text{if } \mathfrak{R}_{\text{obs}} < \| {P}^{t+1} - {P}_{\text{obs}} \| < \mathfrak{R}_{\text{obs}} + r.
			\end{gathered}
		\end{equation}
	\end{itemize}
	\mrev{The three reward components are equally weighted, i.e. $\lambda_1 = \lambda_2 = \lambda_3 = 1$, while $C_1 = 3$ and $C_2 = 0.3$. The threat-zone extension is $r = 0.4$. The goal-proximity threshold $\nu = 0.5$.}

	\section{Expert Critic Formulation via Robust RL} \label{section-III}
	As illustrated in Figure \ref{figure2_schematic}, the initial robust RL is developed, compensating the fixed distributional shift in the observation with adversarial policies. It involves the concurrent design and training of the adversarial policy $\pi^{\textup{adv}}_{\omega}$ and the agent policy $\pi^{\textup{agent}}_{\theta}$ which are parameterized by $\left \{ \omega, \theta \right \}$ respectively. Both policies share a common critic function $Q_{\textup{AR}}\left ( \Phi^t, a^t \right )$ parameterized by $\phi$. The adversarial policies, represented by $\pi^{\textup{adv}}_{\omega}$, introduce perturbations to the action space, controlled by the parameter $\alpha \in \left [0,1 \right]$. When the action $a_t$ is executed in the environment, the agent receives the reward $r_t$, as detailed in subsection \ref{reward_section}. This results in the agent state transitioning from $\Phi^t$ to $\Phi^{t+1}$, as described in \eqref{eq:3}. The experience $e_t = \left ( \Phi^t, a^t, r^t, \Phi^{t+1} \right )$ is then stored in the replay buffer $\mathcal{D}$. A batch of experiences, $\mathcal{D}_{\textup{batch}} = \left (e_1, \cdots, e_N \right )$, is sampled to train the shared critic function $Q_{\textup{AR}}\left ( \Phi^t, a^t \right )$, which is used to update both the agent policy $\pi^{\textup{agent}}_{\theta}$ and the adversarial policy $\pi^{\textup{adv}}_{\omega}$. The reward at time $t$ depends on the state $\Phi^t$ and the joint action $a^t$, which consists of the actions of both players and governs the transition dynamics. The whole process is modelled as a Markov decision process (MDP) $\mathcal{M}_2 = (\mathcal{S}, \mathcal{A}_{\textup{agent}}, \mathcal{A}_{\textup{adv}}, T_2, \gamma, \mathcal{R}, p_0)$, where $\mathcal{A}_{\textup{agent}}$ and $\mathcal{A}_{\textup{adv}}$ represent the continuous action space for the agent and adversary policies, respectively. $T_2$ represents the transition function, $p_0$ represents the initial-state distribution, while $\mathcal{R}$ represents the reward function. The state transition function is given by $T_2 : \mathcal{S} \times \mathcal{A}_{\textup{agent}} \times \mathcal{A}_{\textup{adv}} \times \mathcal{S} \rightarrow \mathbb{R}$, and the reward function is defined by $\mathcal{R} : \mathcal{S} \times \mathcal{A}_{\textup{agent}} \times \mathcal{A}_{\textup{adv}} \rightarrow \mathbb{R}$. In this framework, the agent follows the policy $\pi^{\textup{agent}}_{\theta} : \mathcal{S} \rightarrow \mathcal{A}_{\textup{agent}}$ and the adversary follows the policy $\pi^{\textup{adv}} : \mathcal{S} \rightarrow \mathcal{A}_{\textup{adv}}$. At each time step, $t$, both players observe the state $\Phi^t$, and their respective actions are given by $a_{\textup{agent}}^t = \pi^{\textup{agent}}_{\theta}(\Phi^t)$ and $a_{\textup{adv}}^t = \pi^{\textup{adv}}_{\omega}(\Phi^t)$. According to the zero-sum game structure, the agent receives the reward $r^t = \mathcal{R} \left  (\Phi^t, a_{\textup{agent}}^t, a_{\textup{adv}}^t \right )$, while the adversary receives a reward of $-r^t$.
	Hence, we can say that the optimal value function of the two-player zero-sum game is given by maximizing the worst-case value function due to adversarial policies. As provided in \cite{kamalaruban2020robust}, the following performance objective is considered in the adversarial game,
	\begin{equation} \label{eq:9}
		J\left ( \pi^{\textup{agent}}_{\theta}, \pi^{\textup{adv}}_{\omega} \right ) = \mathbb{E} 
		\left [ \left.\begin{matrix}
			\sum_{t=1}^{\infty} \gamma^{t-1} r^t
		\end{matrix}\right| \pi^{\textup{agent}}_{\theta}, \pi^{\textup{adv}}_{\omega}, \mathcal{M}_2 \right ]. 
	\end{equation}
	where $\sum_{t=1}^{\infty} \gamma^{t-1} r^t$ is the random cumulative return. If we write, $J\left ( \theta, \omega \right ) = J \left ( \pi^{\textup{agent}}_{\theta}, \pi^{\textup{adv}}_{\omega} \right )$ the following objective has to be considered for analysis,
	\begin{equation} \label{eq:10}
		\max_{\theta \in \Theta} \min_{\omega \in \Omega} J\left ( \theta, \omega \right ).
	\end{equation}
	In equation (\ref{eq:11}), $J$ is nonconvex and nonconcave in both $\theta$ and $\omega$. When a mixed strategy is applied to solve (\ref{eq:11}), the optimal solution can be determined over the set of all probability distributions of $\Theta$ and $\Omega$, as given by
	\begin{equation} \label{eq:11}
		\max_{p \in \mathcal{P} \left ( \Theta \right )} \min_{q \in \mathcal{P}\left ( \Omega \right ) } f\left ( p,q \right )  : = \mathbb{E}_{\theta \sim p} \left [ \mathbb{E}_{\omega \sim q} \left [ J\left ( \theta, \omega \right ) \right ] \right ].
	\end{equation}
	The objective in (\ref{eq:11}) can be solved using the algorithm for finding the mixed-Nash equilibrium in generative adversarial networks (GANs) \cite{hsieh2019finding}, as demonstrated for robust RL problems in \cite{kamalaruban2020robust}. In this scenario, the agent and adversarial policy networks can be viewed as the generator and discriminator networks in a GAN, respectively. Sampling-based methods are employed to derive the mixed-Nash equilibrium from the objective function in equation (\ref{eq:11}), as detailed in \cite{hsieh2019finding}. The solution methods for (\ref{eq:11}) has been provided in \cite{panda2024action}. Although this formulation focuses on action-space perturbations, its outcomes generalize across uncertain environments. Robust agents trained under these conditions tend to exhibit more conservative and stable policies, which are resilient to shifts in the input distribution. This property makes the shared critic $Q_{\textup{AR}}$ a valuable expert baseline for adapting to observation-space adversarial attacks (e.g., GPS spoofing or state perturbations).
	\section{Curriculum Guided Adversarial Distributional Adaptation} \label{section-IV}
	\begin{figure}[thpb]
		\centering
		\includegraphics[scale=0.55]{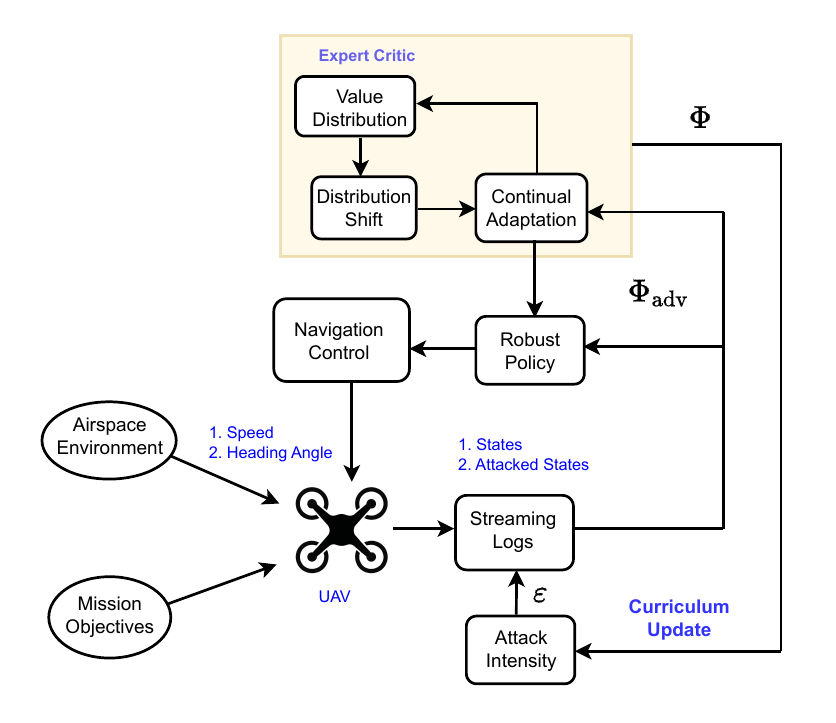}
		\caption{Curriculum guided adaptation scheme for the action robust RL policies against incremental adversarial observations.}
		\label{figure2_schematic}
	\end{figure}
	As illustrated in Figure \ref{figure2_schematic}, the adaptation strategy for the robust RL policies has been shown. The approach employs curriculum-based adversarial observations $\Phi_{\text{adv}}$, with progressively increasing perturbation strength $\epsilon$. The initial action-robust critic is considered as the expert critic. For adaptation, the TD-error is first computed using the original observations $\Phi$ from the expert critic. Using the same critic, the TD error is then computed with adversarial observations $\Phi_{\text{adv}}$. The critic of the robust policy is updated with the Wasserstein distance between these two TD error distributions as depicted in Figure \ref{figure2_schematic}. This adaptation loop is executed for a fixed number of episodes, after which the curriculum is updated with increasing adversarial perturbation strength $\epsilon$. At each curriculum stage, the adapted critic from the previous perturbation level becomes the expert critic for the current stage. The curriculum progresses until the TD error between the expert critic of two successive perturbation strengths falls below a predefined threshold. The adversarial observations employed for both adaptation robust policy is described in the following subsection.
	\subsection{Adversarial Observations for Curriculum Adaptation}
 	We incorporate a curriculum-based adversary during training that gradually increases the intensity of  perturbations encountered by the policy. Adversarial states are generated using a multi-step projected gradient ascent procedure that maximizes the policy loss while constraining perturbations within a $\epsilon$ bounded norm ball \cite{carlini2017towards} given as:
	\begin{equation} \label{eq:12}
		\Phi_{n+1} = \operatorname{Proj} \left[ \Phi_n + \frac{\epsilon}{N_{\textup{step}}} \cdot \textup{sgn} \left( \nabla_{\Phi_n} \mathcal{L}_\pi \right) \right].
	\end{equation}
	Here \( \epsilon \) denotes the perturbation budget, \( \textup{sgn}(\cdot) \) is the sign function, and \( \Pi \) projects the perturbed observation onto the intersection of the valid observation space and the $\epsilon-$ bounded perturbation set around the original observation. Here $\pi$ represents the nominal DDPG policy trained from the agent described in Section \ref{section-III}, while $\mathcal{L}_\pi$ represents the policy loss value. The adversarial curriculum progressively increases \( \epsilon \) throughout the training, exposing the policy to increasingly challenging observation distributions. As the perturbation strength increases, these distributions diverge from the normal observation distribution  encountered during the training. The overall algorithm for adversarial state generation is provided in Algorithm \ref{alg:adv_state} in Appendix \ref{adv_state}. 
	\subsection{Adaptation Framework}
	Generally, predictive likelihood function is used to differentiate different task distributions for a batch of samples $\left ( X,Y \right)$ \cite{wang2022dirichlet}.
	Considering $X = \left ( \Phi^t, a^t \right )$, then the target $Y =  r^t + \gamma Q_{\phi} \left ( \Phi^{t+1}, \pi_{\theta}\left ( \Phi^{t+1} \right ) \right ) $, the likelihood $p_{\xi}(Y \mid X)$ parameterized by $\xi$  is given by,
	\begin{equation} \label{eq:13}
		\begin{aligned}
			p_{\xi}(Y \mid X) &= \prod_i \mathcal{N} \left( y_i, \hat{y}_i, \sigma^2 \right), \\
			&= \prod_i \mathcal{N} \left( r^t_i + \gamma Q_{\phi}(\Phi_i^{t+1}, \pi_{\theta}(\Phi_i^{t+1})) ; Q_{\phi}(\Phi^t_i, a^t_i), \sigma^2 \right).
		\end{aligned}
	\end{equation}
	In \eqref{eq:13}, $\sigma$ refers to the standard deviation of the error distribution.
	\mrev{The representation in (\mrev{\ref{eq:13}}) represents probabilistic characterization of the Bellman prediction error under a Gaussian error model. The predictive likelihood itself is not considered the TD error in the subsequent analysis. The TD error is defined as the signed difference between Bellman target and current critic value.} Hence, for the state action pair $\left ( \Phi^t_i, a^t_i \right )$ following the DDPG policy $\pi_{\textup{DDPG}}$, with the critic value function parameterized by $\phi$ given by $Q_{\phi}$, the TD error is defined as,
	\begin{equation} \label{eq:14}
		\mrev{\scalebox{0.92}{$
				\mathbf{TD}^{\textup{DDPG}}_{\Phi}
				= r^t_i
				+ \gamma Q\left(
				\Phi_i^{t+1},
				\pi_{\textup{DDPG}}\left(\Phi_i^{t+1}\right)
				\right)
				- Q\left(\Phi^t_i, a^t_i\right).
				$}}
	\end{equation}
	Similarly, TD error $\mathbf{TD}^{\textup{AR}}_{\Phi^{\epsilon}_{\textup{adv}}}$ for the action robust critic $Q_{\textup
		{AR}}$ with adversarial experience at time $t$, $ \left\{ \Phi^{\epsilon}_{\textup{adv}}, a^{\epsilon}_{\textup{adv}}, \Phi^{t+1},r^{\epsilon}_{\textup{adv}} \right\}$ is given as,
	\begin{equation}\label{eq:15} 
		\begin{gathered}                          
			\mrev{
				\mathbf{TD}^{\textup{AR}}_{\Phi^{\epsilon}_{\textup{adv}}} 
				= 
				r^{\epsilon}_{\textup{adv}_i} 
				+ \gamma Q_{\textup{AR}}\left( 
				\Phi_i^{t+1}, 
				(1-\alpha)\cdot\pi^{\textup{agent}}_{\theta}
				\left(\Phi_i^{t+1}\right) 
				\right. }\\
			\mrev{
				\left.
				+\alpha\cdot\pi^{\textup{adv}}_{\omega}
				\left(\Phi_i^{t+1}\right) 
				\right) 
				- Q_{\textup{AR}}\left( 
				\Phi^{\epsilon}_{\textup{adv}_i}, 
				a^{\epsilon}_{\textup{adv}_i} 
				\right).
			}
		\end{gathered} 
	\end{equation}
	\mrev{Although Eqs (\ref{eq:14})-(\ref{eq:15}) are instantiated using DDPG as the baseline algorithm in this paper, the proposed adaptation mechanism is not inherently restricted to this algorithm; it requires a learned value/critic function from which TD-error distributions can be constructed and aligned across curriculum stages as shown in the next subsection. Hence, the framework can be extended to other off-policy actor–critic algorithms, such as TD3 and SAC, while on-policy methods would require an appropriate reformulation of the TD-error estimation. DDPG is adopted here as the underlying implementation for the continuous-control UAV deconfliction problem.} In order to analyze the TD error, \mrev{across consecutive curriculum stages, certain regularity assumptions are introduced. These assumptions are considered over the bounded observation region encountered during curriculum adaptation and are not general properties of arbitrary adversarial attacks. The bisimulation-based results in \cite{ferns2011bisimulation, castro2020scalable} provide related theoretical motivation for connecting changes in state-transition distributions to changes in value functions; however, the assumptions below are modelling conditions specific to the present formulation.}
	\begin{assumption} (\mrev{\textbf{Curriculum reward degradation}}) \label{assump_1} 
	\mrev{Let $\epsilon_k$ and $\epsilon_{k+1}$, with $\epsilon_{k+1}>\epsilon_k$, denote two consecutive perturbation budgets in the adversarial curriculum. Over the curriculum considered during adaptation, increasing the strength of the self-induced adversarial perturbation does not improve the expected reward of the policy i.e. 
	\begin{equation}\label{eq:16}
		\mathbb{E}[r_{\epsilon_{k+1}}]-\mathbb{E}[r_{\epsilon_k}] \leq 0.
	\end{equation}
	This assumption applies only to progressively strengthened gradient-based adversarial perturbation.}
	\end{assumption} 
	\begin{assumption}(\mrev{\textbf{Regularization of local value-function}}) \label{assump_2}
		\mrev{Over the bounded observation region considered during curriculum adaptation, the value function under policy $\pi_k^{}$ is assumed to be locally Lipschitz continuous with respect to the observation-space metric, such that}
		\mrev{
		\begin{equation}\label{eq:17}
		\left|Q^{\pi_k^{}}(\Phi_{\epsilon_k})-
		Q^{\pi_k^{*}}(\Phi'_{\epsilon_k})\right|
		\leq L_Q d(\Phi_{\epsilon_k},\Phi'_{\epsilon_k}),
		\end{equation}}
		\mrev{where $L_Q$ is a problem-specific Lipschitz constant.}
	\end{assumption}
	\mrev{The local regularity condition in Assumption \ref{assump_2} is reasonable over the bounded operating region, given that the UAV dynamics and curriculum perturbations are continuous and the critic is represented by a continuous neural network with finite parameters.}
	
%	\begin{assumption}(\mrev{\textbf{Regularization of local value-function}}) \label{assump_2} 
%		\mrev{Considering Lipschitz rewards and transition, with the smooth system dynamics and response, the value function under optimal policy $\pi^{*}$ is Lipschitz continuous with respect to adversarial state $\Phi_{\epsilon_k}$ and the transition state $\Phi'_{\epsilon_k}$, given as,
%		\begin{equation} \label{eq:16}
%			\left | Q^{\pi^*}_k \left (\Phi_{\epsilon_k} \right ) - Q^{\pi^*}_k \left (\Phi'_{\epsilon_k} \right ) \right | \leq L_Q d \left (\Phi_{\epsilon_k}, \Phi'_{\epsilon_k} \right ).
%		\end{equation}
%	\end{assumption}
	%Considering the state evolution of UAV kinematics and environmental response like the obstacle motion are governed by smooth differentiable functions, hence small changes in the input state results in bounded changes in future state transition and value estimates. This validates the usage of assumption \ref{assump_2}. 
	Hence, considering the above assumptions and the Lipschitz continuity of the reward function we can write,
	\begin{equation} \label{eq:18}
		\left | r^{\epsilon_{k}} - r^{\epsilon_{k+1}} \right | \leq -L_r d \left (\Phi_{\epsilon_k}, \Phi_{\epsilon_{k+1}} \right ) \leq 0.
	\end{equation}
	Based on the definition of TD-error, the catastrophic forgetting is quantified as the $1$-Wasserstein distance $\mathcal{W}_1 \left ( \cdot \right )$ between the two TD-error distributions for the adversarial state $\Phi^{\epsilon}_{\textup{adv}}$ is shown as,
	\begin{equation} \label{eq:19}
		{f}^{\epsilon}_{\pi} = \inf_{\tilde{\gamma} \in \Gamma \left ( \mathbf{TD}_{\textup{DDPG}}, \mathbf{TD}^{\epsilon}_{\pi} \right )} \mathbb{E}_{\left ( x,y \right ) \sim \tilde{\gamma}} \left[ \left\| \mathbf{TD}_{\textup{DDPG}}- \mathbf{TD}^{\epsilon}_{\pi} \right\| _1 \right].
	\end{equation}
	\mrev{In the implementation, the TD samples used for distribution alignment are the normalized squared per-sample residuals computed in Algorithm \ref{alg:cata_forget} in Appendix \ref{adv_state} for state-transition sample $i$ i.e. $\mathbf{TD}_i = V_{\mathrm{tar}, i} - V_{\mathrm{adv}, i}$, where $V_{\mathrm{adv}, i}$ is the current critic estimate, and $V_{\mathrm{tar},i}$ is its one-step Bellman target. For equal-sized mini-batches, the empirical Wasserstein-1 distance is obtained by sorting the expert and adaptive TD samples and averaging the absolute differences between corresponding order statistics. During adaptation, the preceding-stage expert critic is frozen, while gradients of this Wasserstein loss are propagated only through the adaptive critic. No additional Bellman-regression loss is applied during curriculum adaptation.}
	\rev{\begin{remarknn}
		Wasserstein distance is used a preferred metric for defining the catastrophic forgetting, as it bounds the shift in both reward and transition models under Lipschitz continuity, consistent with the TD-space boundedness in Assumption \ref{assump_2}. In contrast, the Wasserstein-2 metric encodes second-order geometry \cite{liu2019wasserstein} and introduces quadratic sensitivity to tail-side distributions, making it more fragile to outliers and high-variance perturbations, induced by adversarial observations during adaptation. Wasserstein-1, being a first-order metric, provides stable and finite bounds even when the two distributions exhibit non-overlapping supports. Kullback-Leibler (KL) divergence, on the other hand, is asymmetric and becomes undefined when distributional supports do not overlap, which is frequently encountered in adversarial settings \cite{chen2020distributionally}. 
	\end{remarknn}}
%	The formal definition of an antifragile policy is provided below.
%	\begin{definition}[Antifragile Policy under Adversarial Perturbations]
%		\label{def:antifragile}
%		Let \( \pi^{\textup{af}} \) be a RL policy trained using domain-adaptive mechanisms (e.g., expert critics or adversarial curriculum). Then \( \pi^{\textup{af}} \) is said to be \emph{antifragile} if, for increasing adversarial perturbation levels \( \epsilon_1 < \epsilon_2 < \cdots < \epsilon_K \), the corresponding catastrophic forgetting values
%		\[
%		f^{\epsilon_k}_{\pi^{\textup{af}}} = \inf_{\gamma \in \Pi\left( \mathbf{TD}_{\textup{DDPG}}, \mathbf{TD}^{\epsilon_k}_{\pi^{\textup{af}}} \right)} \mathbb{E}_{(x, y) \sim \gamma} \left[ \left\| x - y \right\|_1 \right],
%		\]
%		are bounded and non-increasing up to a constant \( \delta > 0 \), such that:
%		\begin{equation} \label{eq:27}
%			\left| f^{\epsilon_{k+1}}_{\pi^{\textup{af}}} - f^{\epsilon_k}_{\pi^{\textup{af}}} \right| \leq \delta, \quad \forall k \in \{1, \dots, K-1\},
%		\end{equation}
%		and
%		\begin{equation} \label{eq:28}
%			f^{\epsilon_k}_{\pi^{\textup{af}}} < f^{\epsilon_k}_\pi,
%		\end{equation}
%		where \( \pi \) is a non-adaptive baseline policy (e.g., DDPG or robust RL) evaluated under the same perturbations. 
%	\end{definition}
	 Let $P_{k}$ denote the transition kernel associated with the adversarial observation process at adaptation stage $k$. Under this formulation, the following lemma characterizes the relationship between successive adversarial distributions.
	\begin{lemma} \label{lemma_1}
		The change in expected value function under different adversarial perturbations is controlled by Wasserstein 1-distance. It is expressed as,
		\begin{equation} \label{eq:20}
			\begin{aligned}
				&\left | \mathbb{E}_{\Phi'_{\epsilon_{k+1}}\sim P_{k+1}} \left [Q^{\pi^*}_{k+1} \left (\Phi'_{\epsilon_{k+1}} \right ) \right] - \mathbb{E}_{\Phi'_{\epsilon_{k}}\sim P_{k}} \left [Q^{\pi^*}_{k} \left (\Phi'_{\epsilon_{k}} \right ) \right] \right |  \\  &\leq L_Q \mathcal{W}_1 \left (P_{k+1}, P_k \right ).
			\end{aligned}
		\end{equation}
	\end{lemma}
	\begin{IEEEproof}
		Proof provided in Appendix \ref{lemma_1_proof}.
	\end{IEEEproof}
	Now we consider, the change in the value distributions with the adversarial state instead of transitioned state as follows.
	\begin{lemma} \label{lemma_2}
		The optimal value function does not change abruptly between adversarial perturbation levels
		\begin{equation} \label{eq:21}
			\left |Q^{\pi^*}_{k+1} \left (\Phi_{\epsilon_{k+1}} \right ) - Q^{\pi^*}_{k} \left (\Phi_{\epsilon_{k}} \right )  \right | \leq L_Q d \left (\Phi_{\epsilon_{k+1}}, \Phi_{\epsilon_{k}}\right ).
		\end{equation}
	\end{lemma}
	\begin{IEEEproof}
		Proof provided in Appendix \ref{lemma_2_proof}.
	\end{IEEEproof}
	Having established the relationship between the value function and the perturbed next-state distribution, we now investigate how the transition probability distribution itself evolves as the level of adversarial perturbation increases. This relationship is formalized in the following lemma.
	\begin{lemma} \label{lemma_3}
		If $\mathcal{W}_1 \left ( \cdot \right ) $ represents the 1-Wasserstein distance between two probability distributions $P_k$ and $P_{k+1}$ over the state space of $\mathcal{S}$, then we can consider for the states $\Phi'_{\epsilon_{k}} \sim P_{k}$, $\Phi'_{\epsilon_{k+1}} \sim P_{k+1}$, then we can consider,
		\begin{equation} \label{eq:22}
			\mathcal{W}_1(P_{k+1}, P_k) \leq L_T \mathcal{W}_1(\mathbf{TD}^{\pi^*}_{k+1}(\epsilon_{k+1}), \mathbf{TD}^{\pi^*}_k(\epsilon_k)),
		\end{equation}
		with $L_T$ being a Lipschitz function in the metric space.
	\end{lemma}
	\begin{IEEEproof}
		Proof provided in Appendix \ref{lemma_3_proof}.
	\end{IEEEproof}
	\begin{lemma} \label{lemma_4}
		If we consider the adversarial states $\Phi_{\epsilon_{k}}$ and $\Phi_{\epsilon_{k+1}}$, with $\epsilon_{k+1} > \epsilon_k $ then the distance between the adversarial states $d \left (\Phi_{\epsilon_{k}}, \Phi_{\epsilon_{k+1}} \right )$ is constrained by the distributional distance of the transition distribution $P_k$ and $P_{k+1}$, which is given by,
		\begin{equation} \label{eq:23}
			d \left (\Phi_{\epsilon_k}, \Phi_{\epsilon_{k+1}} \right ) \leq \mathcal{W}_1 \left (P_k, P_{k+1} \right ).
		\end{equation}
	\end{lemma}
	\begin{IEEEproof}
		Proof provided in Appendix \ref{lemma_4_proof}.
	\end{IEEEproof}
	Based on the above assumptions and lemmas, now the following theorem is provided to develop adaptation of the robust policy against incremental adversarial states.
	\begin{theorem} \label{theorem_main}
		Under the curriculum reward-degradation condition in Assumption \ref{assump_1} and the local Lipschitz regularity conditions on the reward and value function introduced in Assumption \ref{assump_2}, let $m$ be a problem-specific positive constant. For any consecutive curriculum stages $k, k+1 \in \left[K\right]$, the corresponding optimal policies $\pi^{k}$ and $\pi^{k+1}$ satisfy the following relationship with respect to the TD error:
		\begin{equation} \label{eq:24}
			\begin{aligned}
				&\mathbb{E} \left [ \mathbf{TD}^{\pi^*_{k+1}} \left ( \epsilon_{k+1}  \right ) - \mathbf{TD}^{\pi^*_{k}} \left ( \epsilon_{k+1}  \right ) \right] \\ &\leq m \mathcal{W}_1 \left ( \mathbf{TD}^{\pi^*_{k+1}} \left ( \epsilon_{k+1}  \right ), \mathbf{TD}^{\pi^*_{k}} \left ( \epsilon_{k}  \right )  \right ).
			\end{aligned}
		\end{equation}
	\end{theorem}
	\begin{IEEEproof}
		Proof provided in Appendix \ref{main_proof}.
	\end{IEEEproof}
	\begin{remarknn}
		\mrev{The proposed curriculum is driven by the adversarial perturbation budget $\epsilon$, rather than by exogenous noise in the collected experiences. Therefore, low-noise in the training data do not eliminate the curriculum effect, since successive stages induce distinct observation and TD-error distributions through increasing perturbation strength. Data variability instead affects the empirical estimation of these distributions: higher variability may require more samples to obtain a reliable Wasserstein-distance estimate, whereas excessively large replay buffers may retain experiences from earlier curriculum stages and mix distinct TD-error regimes. Accordingly, replay data should remain sufficiently representative of the current and neighbouring curriculum stages. Hence, the Wasserstein distance between successive TD-error distributions is used, rather than the raw variance of the dataset, as the criterion for curriculum progression, because it quantifies distributional changes in the critic's learning signal thus reflecting the combined influence of data variability and increasing adversarial difficulty.}
	\end{remarknn}
	Having derived the adaptation criterion, we next establish conditions under which robustness acquired during adaptation transfers to unseen adversarial attack distributions.
	\begin{theorem} \label{theorem:gen}
		We consider the robustness of adaptation derived in Theorem \ref{theorem_main}, with probability $1- \delta$ over $n$ episodes used at stage $K$. The test-train gap for $\pi^{*}_K$ against unseen attack $\mathfrak{A}$  satisfies point-wise certification as:
		\begin{equation}\label{eq:25}
			\begin{gathered}
				\left | \mathbb{E}\left[ R \left (\pi^*_K \text{ under } \mathfrak{A} \right ) \right]   - \frac{1}{n} \sum_{i=1}^n  R\left ( \pi^{*}_K  \right ) \right | \leq \textup{Rad}_K + C \delta_{\mathfrak{A}} \\ + O \left ( \frac{c}{\sqrt{n}} \sqrt{\log \frac{1}{\delta}} \right ).
			\end{gathered}
		\end{equation}
		2. Curriculum-aware certification from the source expert policy:
		\begin{equation} \label{eq:26}
			\begin{gathered}
				\left | \mathbb{E}\left[ R \left (\pi^*_K \text{ under } \mathfrak{A} \right ) \right]   - \frac{1}{n} \sum_{i=1}^n  R\left ( \pi^{*}_{\textup{exp}}  \right ) \right | \leq \textup{Rad}_1 + C \sum_{k=1}^{K-1}  \beta_k  +C \delta_{\mathfrak{A}} \\ + O \left ( \frac{c}{\sqrt{n}} \sqrt{\log \frac{1}{\delta}} \right ).
			\end{gathered}
		\end{equation}
		Here $\textup{Rad}_j$ represents the Rademacher complexity as defined in \cite{wang2019generalization} for the empirical return for the $j^{\textup{th}}$ curriculum stage. 
	\end{theorem}
	\begin{IEEEproof} 
		Proof provided in Appendix \ref{proof:gen}.
	\end{IEEEproof}
	\begin{remarknn}
		Theorem 2 provides a \mrev{conditional} robustness-transfer certificate for curriculum-guided adaptation. The first bound in (\ref{eq:25}) shows that the performance of the final policy under unseen attack is controlled by the discrepancy between the induced TD-error distribution \mrev{rather than specific attack mechanism. The certificate is formulated in terms of the induced TD-error distribution rather than a particular attack mechanism. Consequently, the analysis is not intrinsically restricted to a specific sensing modality.The bound in (\ref{eq:26}) further captures the cumulative adaptation discrepancy $\sum_k\beta_k$ across curriculum stages, thereby limiting value-function drift. Together, these bounds establish robustness transfer through TD-space adaptation rather than sensor- or attack-specific training.}
	\end{remarknn}
	\rev{The adaptation framework in Theorem \ref{theorem_main} and the generalization certificate in Theorem \ref{theorem:gen} \mrev{are} designed to control the trade-off between the nominal performance and robustness,  \mrev{thereby limiting excessive policy conservatism as the perturbation budget increases}. The performance-robustness trade-off is imposed subject to a bound on how much the underlying value estimates are allowed to change.} \mrev{Theorem \ref{theorem:gen} shows that, under the test-time error-distribution bound, improved robustness is associated with a provably bounded change in return relative to the expert policy. Thus, the performance–robustness trade-off is not eliminated but explicitly bounded.} The process to implement curriculum-adaptation is explained in Algorithm \ref{alg:adv_adaptation} as follows.
		\begin{algorithm}
				\caption{\mrev{TD-Error Guided Curriculum Adaptation}}
				\label{alg:adv_adaptation}
				
				{\mrev{\textbf{Input:} Expert TD-error samples $\mathbf{TD}_{\textup{exp}}$}
						\mrev{initial robust critic $Q_{\textup{ad}}$}, \mrev{initial adversarial strength
						$\epsilon$}, \mrev{curriculum increment $\Delta\epsilon$}, \mrev{maximum strength
						$\epsilon_{\textup{max}}$}, and \mrev{minibatch size $N$}.}\\
				{\mrev{\textbf{Output:} Curriculum-adapted robust policy and critic.}}
				
				\begin{algorithmic}[1]
						
						\While{\mrev{$\epsilon \leq \epsilon_{\textup{max}}$}}
						
						\State \mrev{Generate adversarial transitions at strength $\epsilon$
							using the relation in (\ref{eq:12}) and store them in
								$\mathcal{D}_{\textup{adv}}$.}
						
						\State \mrev{Sample $N$ transitions from $\mathcal{D}_{\textup{adv}}$
								and $N$ reference TD errors from $\mathbf{TD}_{\textup{exp}}$.}
						
						\State \mrev{Compute the adaptive critic TD errors
								$\mathbf{TD}_{\textup{ad}}$ for the sampled transitions using
								(\ref{eq:15}).}
						
						\State \mrev{Compute the TD-distribution alignment loss}
						\[
						\mrev{
								\mathcal{L}_{\textup{crit}}
								=
								\mathcal{W}_{1}
								\left(
								\mathbf{TD}_{\textup{exp}},
								\mathbf{TD}_{\textup{ad}}
								\right).
							}
						\]
						
						\State \mrev{Update $Q_{\textup{ad}}$ by minimizing
								$\mathcal{L}_{\textup{crit}}$, with
								$\mathbf{TD}_{\textup{exp}}$ treated as the fixed reference.}
						
						\State \mrev{Update the agent/adversarial policies to obtain CurAd policy and target networks
								using the action robust-RL update \cite{tessler2019action}.}
						
						\State \mrev{Repeat Steps 1-6 for the prescribed adaptation episodes
								at the current $\epsilon$.}
						
						\State \mrev{Set the adapted TD-error distribution as the reference
								for the next curriculum stage:
								$\mathbf{TD}_{\textup{exp}}
								\leftarrow \mathbf{TD}_{\textup{ad}}$.}
						
						\State \mrev{$\epsilon \leftarrow \epsilon+\Delta\epsilon$.}
						
						\EndWhile
						
						\State \mrev{\Return CurAd Policy.}
						
					\end{algorithmic}
			\end{algorithm}

	\section{Spoofing Attack Models for Evaluation}\label{spoofing_evaluation}
	\subsection{Fixed Spoofing Attack}
	 GNSS spoofing aims to manipulate the positional awareness of the UAV by injecting falsified satellite signals \cite{panda2024action}. These attacks operate at the signal layer and do not require any access to the policy network or its gradients, thus constituting a fully black-box threat model. The attacker exploits pseudorange manipulation to introduce consistent positional drift, misleading the UAV into paths based on erroneous geolocation data. \rev{The test-time spoofing scenario, considered in the analysis, models a persistent position bias produced by a fixed pseudorange offset at the receiver (constant code-phase bias) rather than continuously accelerating drift as provided in \cite{panda2025real}. The attack is feasible in practice when the spoofer:
			\begin{itemize}
				\item Maintains a small power advantage (1-3 dB) to capture without the automatic gain control (AGC) jumps.
				\item  Closely matches the carrier/Doppler ($\leq 0.1$ Hz) to avoid the phase locked loop disturbances.
				\item Injects a bounded instantaneous code-phase bias that keeps the receiver autonomous integrity monitoring (RAIM) residuals small. 
				\item Emulates the possible satellite configuration geometry to defeat the degree-of-arrival checks.
				\item Exploits the absence of navigation message authentication in civilian GNSS systems.
		\end{itemize}}
		\rev{Under the above signal-level constraints, the constant code-offset maps directly to a constant per-episode pseudorange bias and hence creating a persistent positional offset in the iterative least square navigation solution detailed in \cite{panda2025real}. Thus we can say that the modelling choice of persistent GNSS threat is practical based on the above parameters which can be implementable with a software defined radio (SDR) hardware.} GNSS spoofing aims to induce high-magnitude state-space distribution shifts that are not gradient-driven and are thus semantically and statistically distinct from algorithmic perturbations from the adversarial observations used for adaptation.
	\subsection{Dynamic Spoofing Attack}
	Dynamic GNSS spoofing represents a \mrev{stealth-constrained alternative to unconstrained position-bias injection described in the previous subsection}. Recent surveys  emphasize that spoofing detection against these attacks remains challenging when the attack is gradual, power-controlled, and consistent with expected receiver dynamics \cite{zeng2026gnss}. Here, a dynamic obstacle-luring spoofing attack is modelled where the objective of the adversary is not to immediately cause mission failure, but to distort the UAV's perception of obstacle clearance. If we consider the true UAV position $P^t$ at time $t$ while the nearest obstacle position, represented by $P^t_{\text{obs}}$, the attacker aims to construct a spoofing direction away from the obstacle given as 
	\begin{equation}\label{eq:28}
		u^t = \frac{P^t - P_{obs}^t} {\left\| P^t - P_{obs}^t \right\|}.
	\end{equation}
	Hence, the desired spoofing bias is then defined as $\mathfrak{b}^{t, *} = \zeta_p u^t$, where $\zeta_p$ denotes the maximum spoofing displacement permitted by the attacker. The key intuition is that the agent perceives to be farther from the obstacle than its true position. Consequently, the repulsive force generated by the obstacle appears weaker, reducing deconfliction behaviour and encouraging trajectories that pass closer to obstacles. A realistic GNSS spoofer cannot instantaneously shift the navigation solution by arbitrary amounts. Such abrupt changes would be detected by carrier tracking loops, Doppler consistency and  RAIM checks, GNSS-inertial navigation systems (INS) consistency verification, and motion-model validation. Hence, we can state that the gradual spoofing bias evolves as per 
	\[
	\dot{\mathfrak{b}}^{\,t,*} = \frac{b^{t,*} - b^t}{\Delta T},
	\]
	where the rate of change is constrained by $\dot{\mathfrak{b}}^{t} \leq r_{\text{max}}$ while the spoofing acceleration satisfies $\dot{\mathfrak{b}}^{t} - \dot{\mathfrak{b}}^{t-1} \leq a_{\text{max}} \Delta T $. The applied spoofing bias is updated as $\mathfrak{b}^{t+1} = \mathfrak{b}^t + \dot{\mathfrak{b}}^{t} \Delta T$ subject to $\left\| \mathfrak{b}^{t+1} \right\| \leq \zeta_p.$ These constraints emulate practical SDR-based spoofers that gradually manipulate pseudorange measurements while remaining unobservable to standard GNSS systems \cite{psiaki2016gnss}. To increase stealthiness, the attack is activated only when the UAV approaches an obstacle. The true obstacle clearance is given as 
	\[
	d^t = \left\| P^t - P_{obs}^t \right\|.
	\]
	If we define the activation gate as,
	\[
	g^t = \operatorname{clip}\!\left(
	1 - \frac{d^t - d_s}{d_a - d_s},
	\, 0, \, 1
	\right),
	\]
	where $d_s$ denotes the safety radius and $d_a$ denotes the attack activation radius. The attack implemented in the simulator directly incorporates several practical constraints used by modern GNSS spoofers i.e. position bias $\zeta_p$ of 8m, velocity bias $\zeta_v$ of $2 m/s$ , bounded bias rate of $r_{\text{max}} = 1.2$, delayed bias acceleration $a_{\text{max}} = 0.85$ while considering low-amplitude receiver noise $\sigma_{\text{rec}} = 0.05$.
	
	\section{Results and Discussion} 
	\label{section-V}
	\subsection{Numerical Simulations}
	The software implementation of the deep deterministic policy gradient algorithm (DDPG) for UAV navigation and deconfliction in environments with dynamic 3D obstacles was carried out as described in \cite{li2023uav}. During the training phase, the initial position of the UAV was randomly generated with a mean vector of approximately $\left( 0, 2, 5 \right)$ and a variance ranging from 0 to 1, while the target destination was set consistently at $\left( 10, 10, 5.5 \right)$. The aerial traffic 3D trajectories are generated according to the trajectory models described in \cite{panda6755038meta}. To promote policy generalization, a trajectory pattern is randomly sampled from the predefined set at the beginning of each training episode. To ensure realistic maneuverability, kinematic constraints were imposed, including a maximum ascent angle of $5 \pi / 9$ and a maximum descent angle of $-15 \pi / 36$. Furthermore, a safety buffer was implemented by enforcing a protective radius of 1.5 units around the UAV to mitigate collision risks. The reward function calculates the conflict penalty by augmenting the UAV's radius with an additional buffer of 0.4 units. 
	\subsection{Results using Action Robust Reinforcement Learning}
\begin{figure}[thpb]
		\centering
		\includegraphics[scale=0.18]{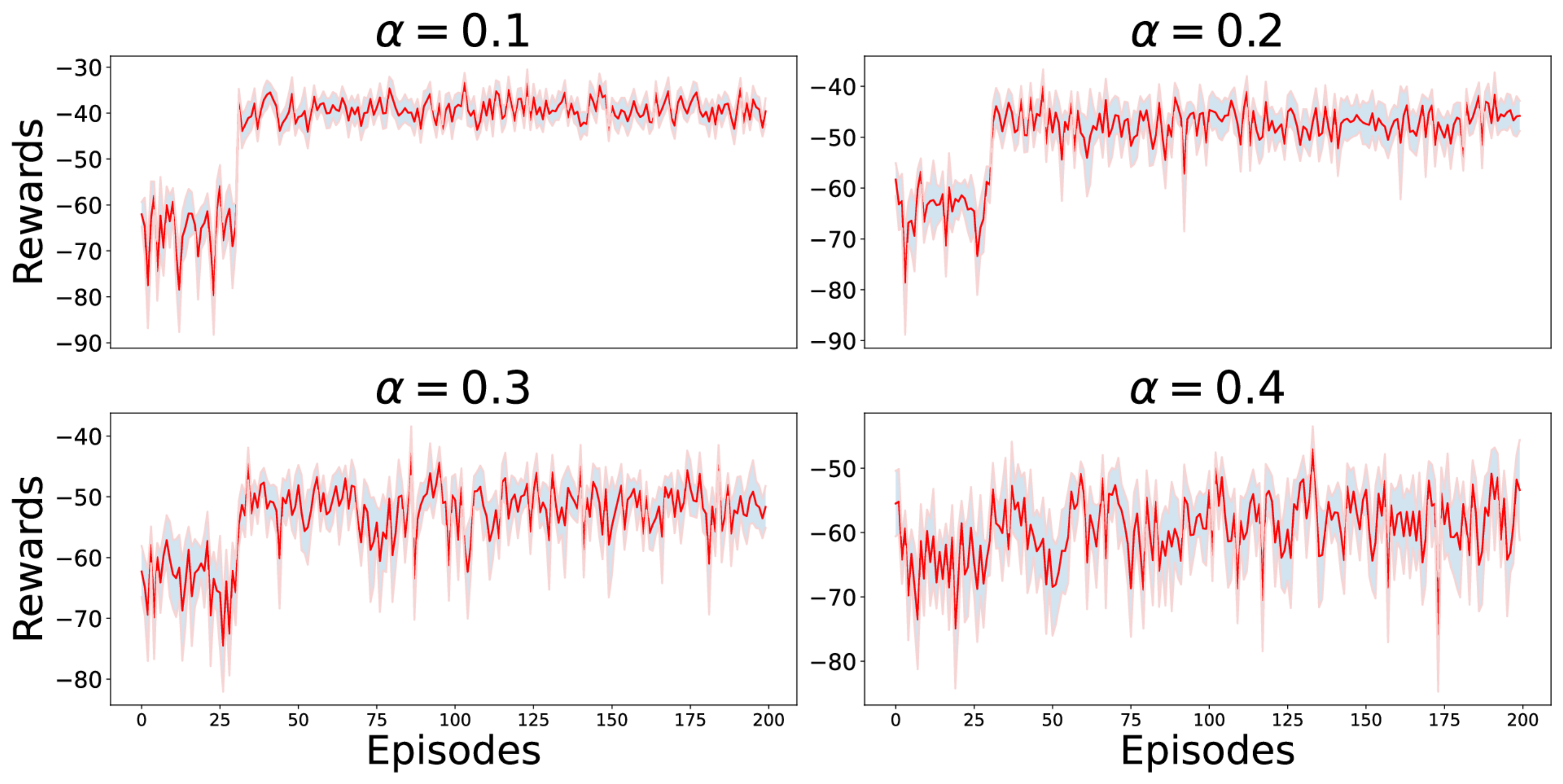}
		\caption{The mean and uncertainty of the convergence episodic reward plot for action robust RL with different $\alpha$ when run with 10 Monte Carlo runs.}
		\label{figure3_convergence}
	\end{figure}

\begin{figure}[thpb]
		\centering
		\includegraphics[scale=0.16]{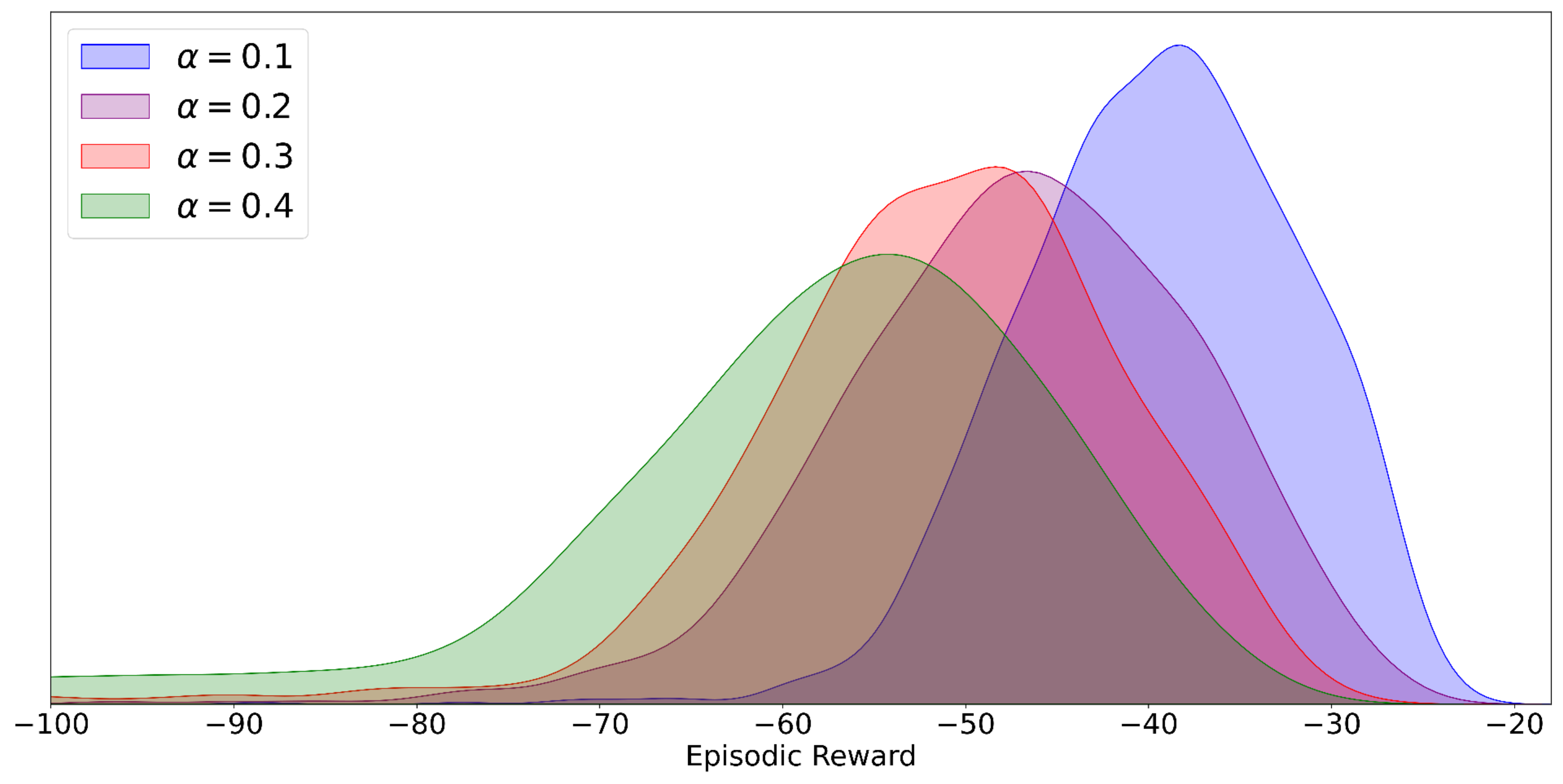}
		\caption{Density plot of the converged rewards from previous 100 episodes from different $\alpha$ from action robust RL algorithms over 10 Monte Carlo runs.}
		\label{figure4_kde_rewards}
	\end{figure}

	As illustrated in Figure \ref{figure3_convergence}, the episodic rewards are plotted against the training episodes for action-robust RL with $\alpha = { 0.1, 0.2, 0.3, 0.4 }$. It is evident that as the value of $\alpha$ increases, the episodic rewards exhibit a noticeable decline accompanied by increased uncertainty, particularly during the exploitation phase. A detailed examination of the reward distributions, as shown in Figure \ref{figure4_kde_rewards}, reveals that increasing $\alpha$ shifts the reward distribution to the left, while also increasing its variance.
	\begin{figure}[thpb]
		\centering
		\includegraphics[scale=0.18]{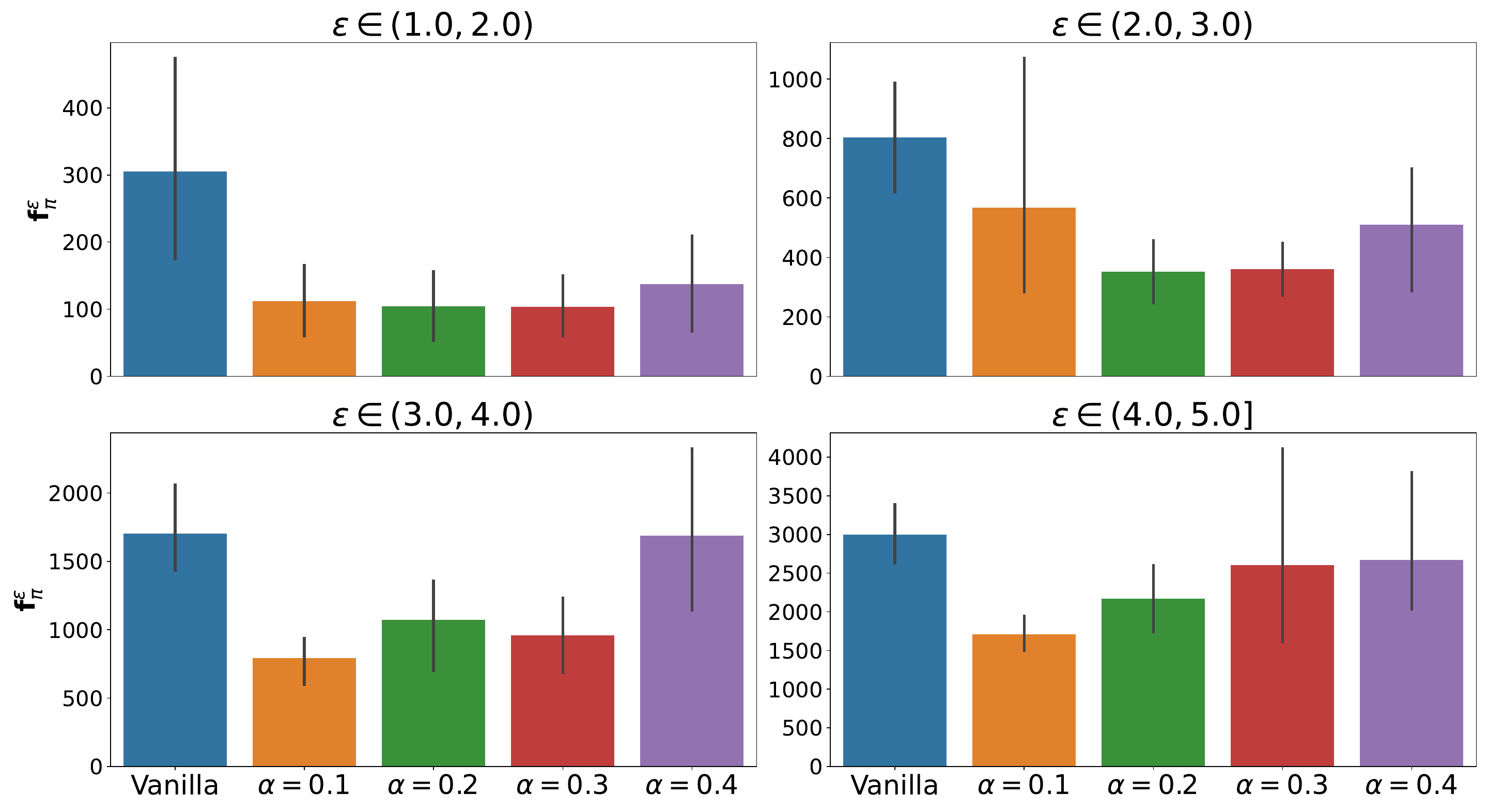}
		\caption{Forgetting value ${f}^{\epsilon}_{\pi}$ of vanilla and action robust RL with different $\alpha$ under PGD attack when the value of $\epsilon$ is varied from 1.0 to 5.0.}
		\label{figure5_forgetting_action_robust}
	\end{figure}
	
	\mrev{\subsection{Empirical Assessment of Lipschitz Smoothness}}
	
	\mrev{To examine whether the regularity conditions adopted in Assumption 2 are representative of the agent state-space distribution considered in this study, we empirically evaluate the local sensitivity of the reward function, transition dynamics, and learned critic. State-action samples are collected from trajectories of the trained robust and nominal policy, and each sample $i$ is paired with a neighbouring sample $j$ in the standardized state-action space. Standardization is applied to prevent variables with different physical scales from disproportionately influencing the Euclidean distances.}
	
	\mrev{The Lipschitz constant estimates are obtained using the inequality relations in (\ref{eq:16}), (\ref{eq:17}), and (\ref{eq:20}). Table \ref{tab:epsilon_comparison} summarizes the resulting estimates for Vanilla DDPG and action-robust DDPG under nominal conditions ($\epsilon=0$) and the adversarial curriculum condition ($\epsilon=0.25$). For the action-robust DDPG critic, the estimated constants under nominal conditions are $\widehat{L}_R=0.505$, $\widehat{L}_T=1.420$, and $\widehat{L}_Q=6.047$. Under $\epsilon=0.25$, these estimates decrease to $0.239$, $1.094$, and $5.584$, respectively. The corresponding $P{95}$ values of the critic state-gradient norm, $P{95}!\left(\lVert\nabla_s Q\rVert_2\right)$, are $8.069$ and $8.402$ for $\epsilon=0$ and $\epsilon=0.25$, respectively. Finite empirical local sensitivity is also observed for Vanilla DDPG, for which $\widehat{L}_R$, $\widehat{L}_T$, and $\widehat{L}_Q$ remain bounded over the sampled trajectories.}
	
	\begin{table}[t] 
		\centering 
		\caption{\mrev{Comparison of the Lipschitz constants for DDPG and action robust DDPG}} 
		\label{tab:epsilon_comparison} 
		\begin{tabular}{lccccc} 
			\toprule 
			\textbf{\mrev{Policy}}  
			& \mrev{$\epsilon$}  
			& \mrev{$\widehat{L}_R$}  
			& \mrev{$\widehat{L}_T$}  
			& \mrev{$\widehat{L}_Q$}  
			& \mrev{$P_{95}\!\left(\lVert\nabla_s Q\rVert_2\right)$} \\ 
			\midrule 
			\mrev{Vanilla DDPG} & \mrev{0.00} & \mrev{0.414} & \mrev{1.509} & \mrev{0.515} & \mrev{1.097} \\ 
			\mrev{Vanilla DDPG} & \mrev{0.25} & \mrev{0.616} & \mrev{1.127} & \mrev{0.437} & \mrev{0.995} \\ 
			\midrule 
			\mrev{Action Robust DDPG} & \mrev{0.00} & \mrev{0.505} & \mrev{1.420} & \mrev{6.047} & \mrev{8.069} \\ 
			\mrev{Action Robust DDPG} & \mrev{0.25} & \mrev{0.239} & \mrev{1.094} & \mrev{5.584} & \mrev{8.402} \\ 
			\bottomrule 
		\end{tabular} 
	\end{table}
	
	\subsection{Distributional Adaptation}
	The action-robust RL model with $\alpha = 0.1$ demonstrates the least catastrophic forgetting among robust frameworks, making it the preferred choice for adapting agents and adversarial policies against adversarial samples of increasing strength. These pre-trained policies are adapted following the procedures outlined in Algorithm \ref{alg:adv_adaptation}. \rev{ Overall the adaptation across curriculum took close to $8458.97$ seconds which equates to 2.35 hours and the simulation took place in a PC with the GPU NVIDIA RTX A2000 having the processor speed of 2904.0 MHz.} \mrev{This represents an offline adaptation cost and is expected to exceed that of fixed-distribution robust RL, since adaptation is repeated across successive perturbation levels ($\epsilon=2.0$-$5.0$, $\Delta\epsilon=0.25$), with adversarial-state generation, TD-error computation, Wasserstein alignment, and critic updates performed at each stage. For $K$ curriculum stages, $E$ episodes, and $T$ steps, the resulting complexity scales as $\mathcal{O}(KET(N_{\mathrm{PGD}}C_{\mathrm{adv}}+C_{\mathrm{TD}}+C_{\mathrm{W}}+C_{\mathrm{upd}}))$. This additional computation is incurred entirely offline and does not increase the computational complexity of policy execution during deployment; instead, it provides the improved robustness to unseen spoofing attacks demonstrated in the subsequent evaluation.}
	
	\mrev{The intermediate checkpoints at $\epsilon = 3.0, 4.0 $ and $5.0$ were examined, and the $\epsilon = 4.0$ checkpoint was retained as the deployment policy based on the deconfliction evaluation. Hence, the CurAd policy at $\epsilon = 4.0$ is compared against the baseline policies in the further subsections.}
	
\begin{figure}[thpb]
		\centering
		\includegraphics[scale=0.18]{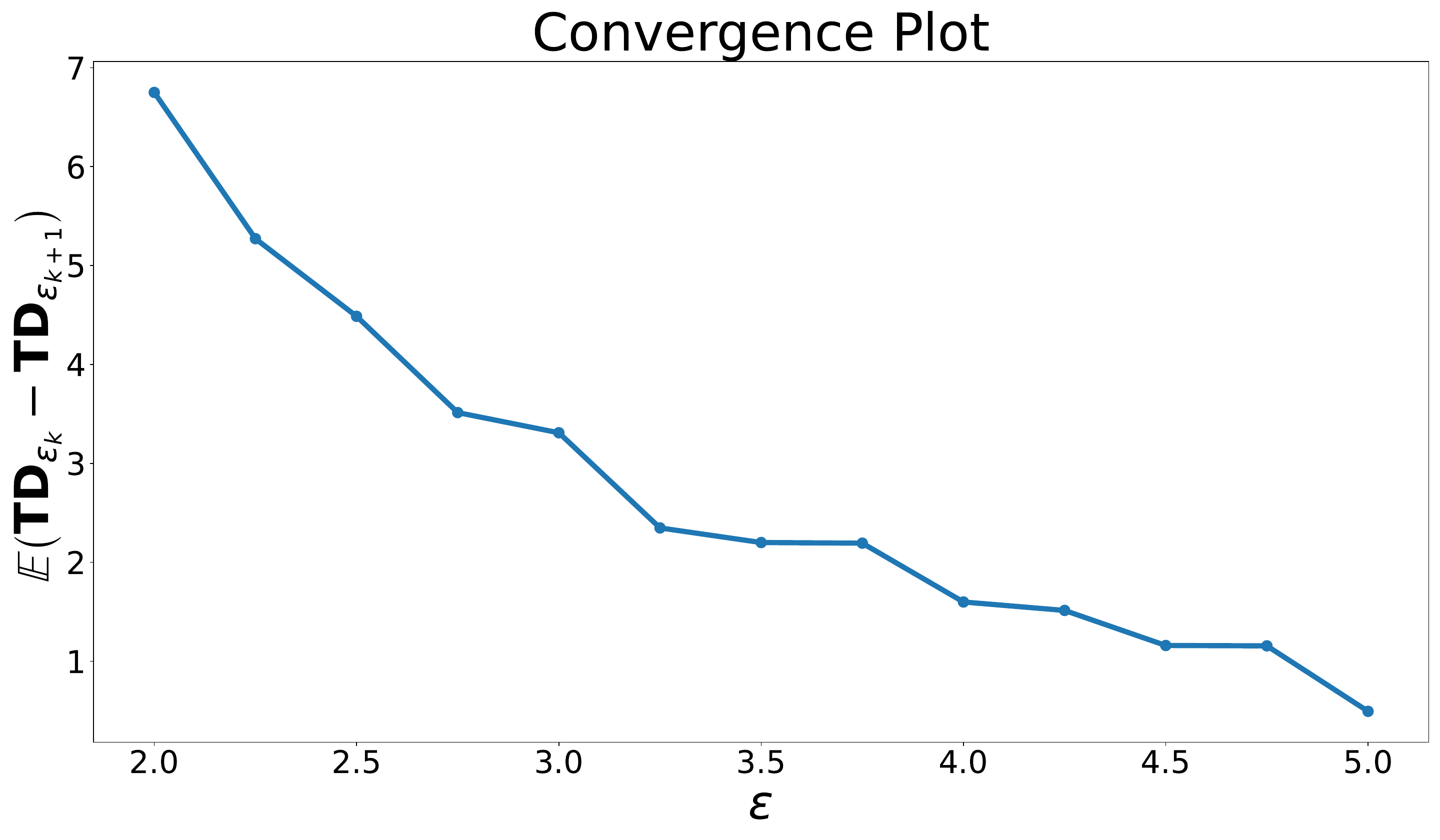}
		\caption{Convergence plot of the value distribution bound with the increase in $\epsilon$ to ensure adaptation.}
		\label{figure6_antifragile_convergence}
	\end{figure}

	The convergence characteristics for the algorithm governed by Theorem \ref{theorem_main}, is shown in Figure \ref{figure6_antifragile_convergence}. During curriculum adaptation, the algorithm maintains stable forgetting values even at high perturbation strengths ($\epsilon=5.0$); however, we conservatively evaluate generalization performance up to $\epsilon = 4.0$ during testing. This conservative approach is motivated by two factors. First, the $\epsilon = 4.0$ checkpoint was selected for the reported deployment evaluations, even though the curriculum adaptation was conducted up to $\epsilon = 5.0$. Second, test-time behavior is generally expected to remain within the certified robustness envelope established during training.  While the observed flatter forgetting curve at $\epsilon=5.0$  is encouraging, we cannot conclusively claim reliable generalization at higher perturbation level in the absence of additional  theoretical guarantees. To evaluate the forgetting value ${f}^\epsilon_{\pi}$, we compare the curriculum adapted RL agent and adversarial policies under PGD attacks with strengths varying from $\left[ 1.0, 5.0 \right]$ conducted in 10 steps. The forgetting value ${f}^\epsilon_{\pi}$ of the curriculum adapted policy is consistently lower than that of the action robust framework with $\alpha = 0.1$, the effect being particularly significant for $\epsilon \in \left( 2.0, 3.0 \right)$ and $\epsilon \in \left( 4.0, 5.0 \right]$. 
%	\begin{figure}[thpb]
%		\centering
%		\includegraphics[scale=0.18]{fragility_pic.pdf}
%		\caption{Distributional distance between the action robust critic with respect to the critic obtained with increasing perturbation of adversarial strength $\epsilon$.}
%		\label{figure7_fragility_pic}
%	\end{figure}
%	\begin{figure}[thpb]
%		\centering
%		\includegraphics[scale=0.18]{forgetting_adaptive_python.pdf}
%		\caption{Comparing the catastrophic forgetting value for curriculum-adapted with the robust framework for different intensities of adversarial attacks.}
%		\label{figure6_forgetting_adaptive_robust}
%	\end{figure}
	\subsection{Robustness Generalization against Unseen Spoofing Attacks}
	The curriculum adapted mechanism is compared against the benchmarking robust and adversarial meta learning algorithms as outlined in Appendix \ref{bench_algo}. We evaluate unseen threats by their TD-space proximity $\delta_{\mathfrak{A}}$ to the final curriculum stage; Theorem \ref{theorem:gen} applies whenever this proximity is bounded, regardless of the attack’s observation-space mechanism. We evaluate the effectiveness of the fixed and dynamic attacks with test rewards which signifies mission success rates, trajectory efficiency and navigation around the perceived obstacles. 
	\subsubsection{Fixed Spoofing Attacks}
		\begin{figure}[thpb]
			\centering
			\includegraphics[scale=0.18]{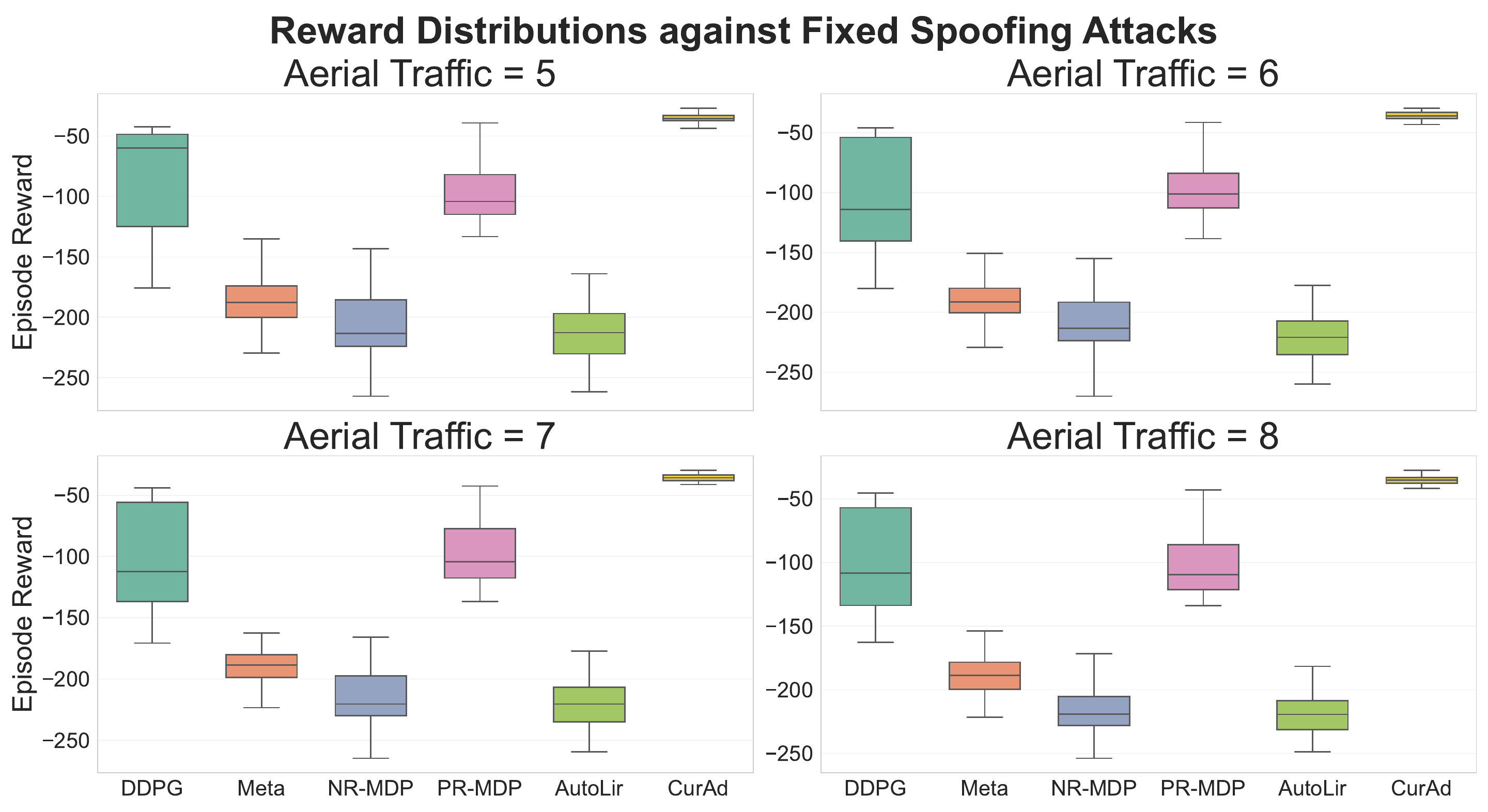}
			\caption{Reward distributions under fixed GNSS spoofing attacks, illustrating superior reward retention of the curriculum-adapted policy across increasing aerial traffic densities with the simulations conducted across 100 testing episodes.}
			\label{figure_reward_fixed}
		\end{figure}
		\begin{figure}[thpb]
		\centering
		\includegraphics[scale=0.18]{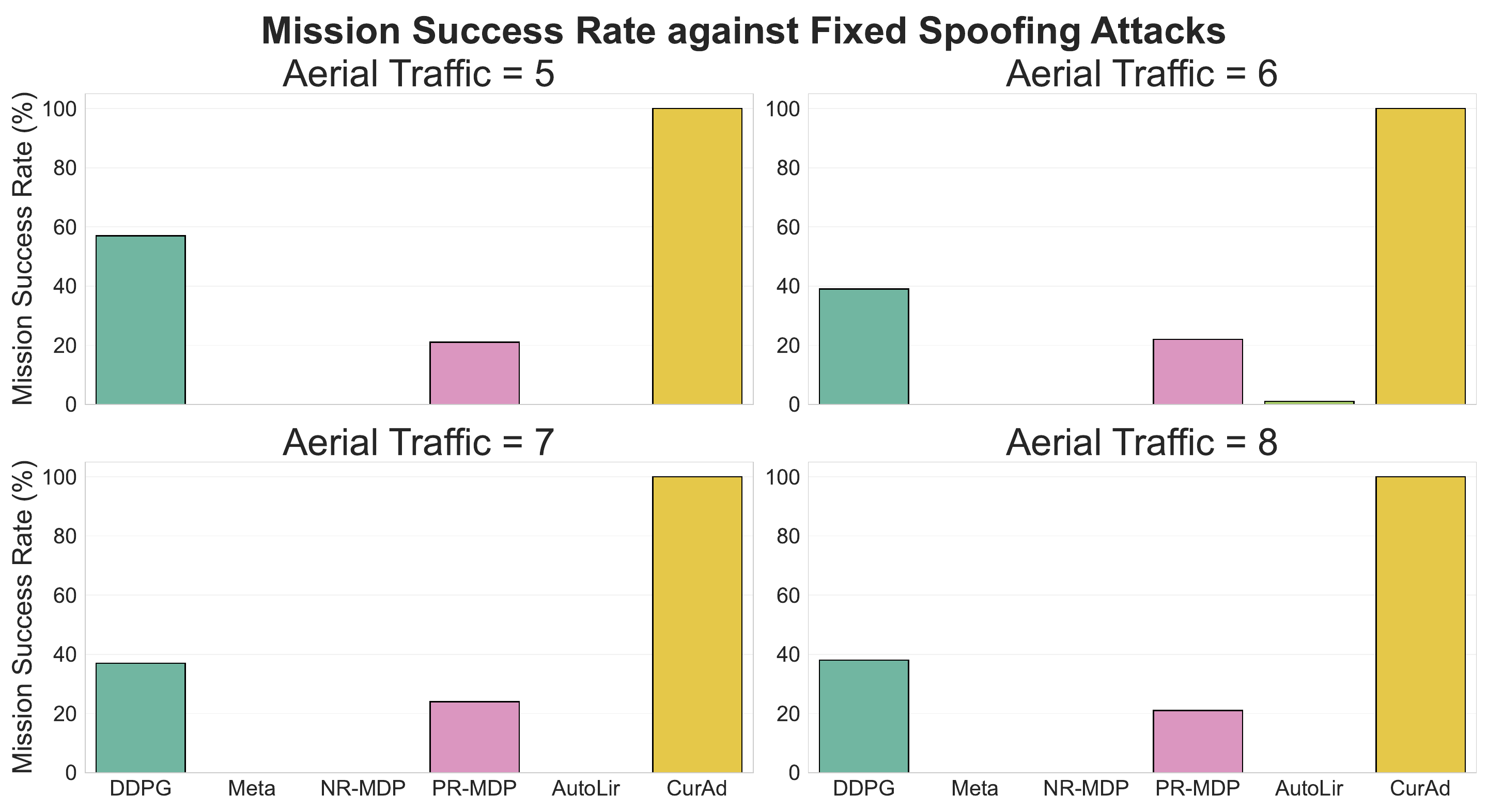}
		\caption{Mission success rates under fixed GNSS spoofing attacks with increasing aerial traffic, where failure corresponds to exhausting the 500-step flight budget before reaching the destination with simulations conducted across 100 testing episodes}
		\label{figure_mission_success_fixed}
		\end{figure}
	To evaluate robustness generalization beyond the synthetic perturbations used during adaptation, the vanilla, robust, meta-adversarial and curriculum adapted policies were tested under fixed GNSS spoofing attacks. In this scenario, the UAV receives a persistently biased position estimate throughout the mission, producing a deployment-time observation shift that is fundamentally different from the gradient-based perturbations used during curriculum adaptation. Each episode was limited to a maximum flight duration of 500 decision steps. A mission was considered successful only if the UAV reached its destination within this budget. This metric therefore quantifies goal completion within the finite mission horizon and should not be interpreted as a standalone collision- or conflict-free safety metric. Consequently, mission failure under fixed spoofing indicates that the navigation error induced by the attack exhausted the available flight time before the target could be reached.
	
	The results demonstrate a clear advantage for the proposed curriculum-adapted robust policy as per Figure \ref{figure_reward_fixed} and \ref{figure_mission_success_fixed}. Across all aerial-traffic levels, curriculum adapted policy consistently achieved the highest reward and maintained the lowest performance variability under fixed spoofing attacks. In contrast, the reward distributions of DDPG, Adversarial Meta, NR-MDP, PR-MDP, and AutoLiRPA deteriorated substantially as the persistent spoofed observations caused increasingly inefficient navigation trajectories. The mission-success results provide even stronger evidence of robustness generalization. CurAd achieved near-perfect mission completion across all traffic densities, whereas the baseline methods frequently failed to reach the destination within the 500-step flight budget. Notably, adversarial Meta, NR-MDP, and AutoLiRPA exhibited almost complete mission failure, while DDPG and PR-MDP achieved only limited success. It is also noteworthy that vanilla DDPG outperforms several robust baselines under fixed spoofing. This result indicates that robustness to the perturbation model used during training does not necessarily translate to improved performance under the structurally different spoofing attack considered here. \mrev{At the highest aerial-traffic density, the reduced availability of collision-free manoeuvres, compounded by spoofing-induced position errors, can produce longer avoidance trajectories that exhaust the 500-step mission horizon. Thus, residual failures primarily reflect the interaction between dense dynamic traffic, adversarial observation error, and the finite mission duration rather than necessarily indicating a collision.}
	
	These results are significant because the fixed spoofing attack was not used during curriculum adaptation. The proposed policy was adapted using progressively stronger gradient-based observation perturbations, yet it remains effective when evaluated against a black-box GNSS-induced position bias. This supports the central hypothesis of the paper: robustness can generalize across attack domains when adaptation preserves value-function consistency rather than merely fitting to a specific perturbation model. The comparatively narrow reward spread and consistently high mission-success rate suggest that TD-error distribution alignment helps stabilize value estimation under persistent observation bias. Instead of only improving resistance to the perturbations encountered during training, the proposed method enables the policy to maintain robust and goal-directed behaviour under the evaluated unseen cyber-physical attack. 
	
	\subsubsection{Dynamic Spoofing Attacks}
		\begin{figure}[thpb]
		\centering
		\includegraphics[scale=0.18]{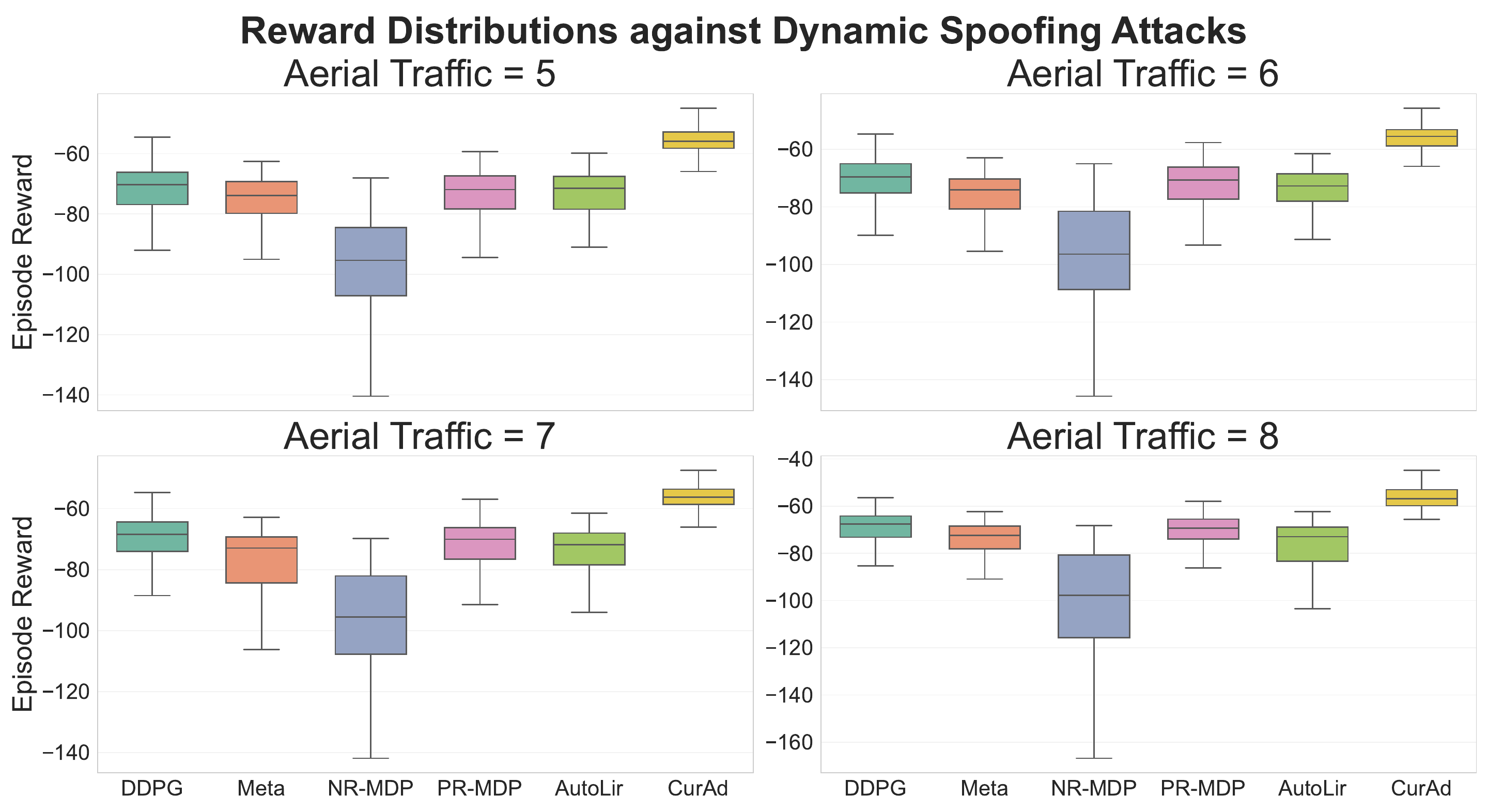}
		\caption{Reward distributions under dynamic obstacle-luring GNSS spoofing attacks across increasing aerial traffic densities  showing superior reward retention and lower performance variability for the curriculum-adapted policy with simulations conducted across 100 testing episodes.}
		\label{figure_reward_dynamic}
		\end{figure}
		\begin{figure}[thpb]
		\centering
		\includegraphics[scale=0.18]{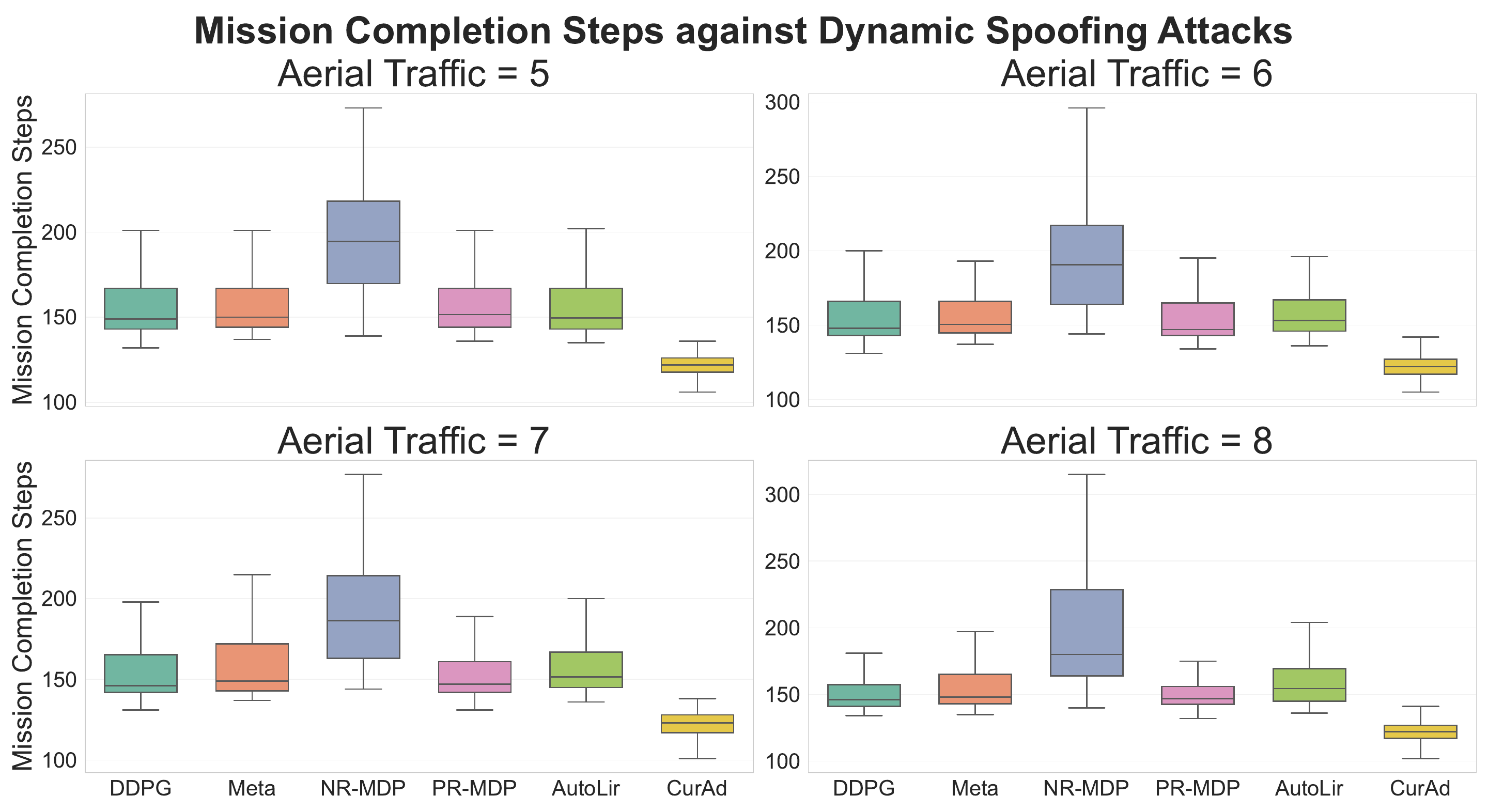}
		\caption{Mission completion steps under dynamic obstacle-luring GNSS spoofing attacks across increasing aerial traffic densities, where fewer steps indicate greater navigation efficiency with simulations conducted across 100 testing episodes.}
		\label{figure_steps_dynamic}
		\end{figure}
		In this section, the evaluation results after benchmarking the performance of curriculum adapted policy with vanilla, robust and meta policies under an adaptive obstacle-luring GNSS spoofing attack are analyzed. Unlike the fixed spoofing scenario, which introduces a persistent navigation bias, the dynamic attacker modifies the perceived UAV position according to the local obstacle geometry while remaining subject to constraints on spoofing magnitude, bias rate, and acceleration. Such constraints are consistent with practical carry-off spoofing attacks, where gradual manipulation is preferred to avoid triggering navigation integrity checks, sensor-consistency tests, or abrupt trajectory anomalies \cite{psiaki2016gnss}. Consequently, the attack is designed to distort obstacle perception rather than directly induce mission failure. Under these constraints, the agents remain capable of reaching the destination. Therefore, mission success alone is insufficient to characterize attack impact. Instead, the primary effects emerge through navigation efficiency and perception-reality mismatch. By biasing the perceived UAV position away from nearby obstacles, the attacker creates the illusion of increased obstacle clearance, potentially altering deconfliction behaviour and route selection.
		
		The reward distributions in Figure \ref{figure_reward_dynamic} indicate that the proposed curriculum-adapted policy consistently achieves the highest reward across all aerial-traffic densities while exhibiting comparatively low variability between episodes. In contrast, NR-MDP experiences the largest reward degradation and variance, suggesting greater sensitivity to the non-stationary observation errors introduced by the attack. DDPG, Meta, PR-MDP, and AutoLiRPA demonstrate intermediate performance but remain consistently below the curriculum-adapted policy. A clearer distinction emerges from the mission-completion step distributions as per Figure \ref{figure_steps_dynamic}. Although most controllers eventually reach the goal, substantial differences are observed in the number of steps required to complete the mission. The curriculum-adapted policy consistently requires the fewest steps across all traffic levels, whereas NR-MDP exhibits both the longest completion times and the highest variability. This behaviour suggests that dynamic spoofing primarily affects path efficiency by inducing suboptimal deconfliction decisions and unnecessary trajectory deviations rather than causing outright mission failure.
			\begin{figure}[thpb]
			\centering
			\includegraphics[scale=0.18]{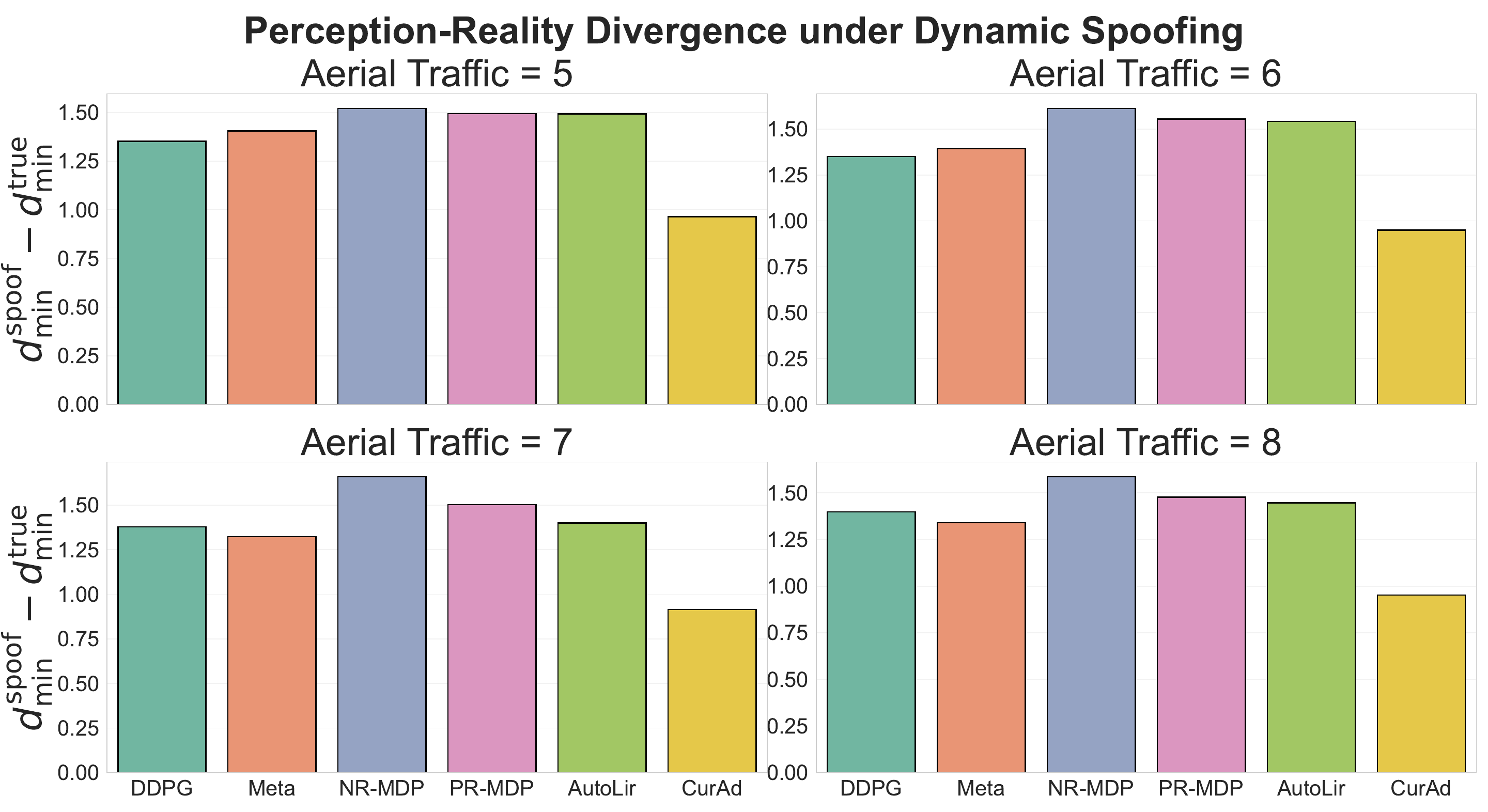}
			\caption{Perception-reality divergence under dynamic obstacle-luring GNSS spoofing attacks, measured as the difference between minimum perceived and true obstacle clearance, where lower values indicate greater resistance to spoofing-induced perception errors.}
			\label{figure_reality_divergence}
		\end{figure}
		\begin{figure}[thpb]
			\centering
			\includegraphics[scale=0.50]{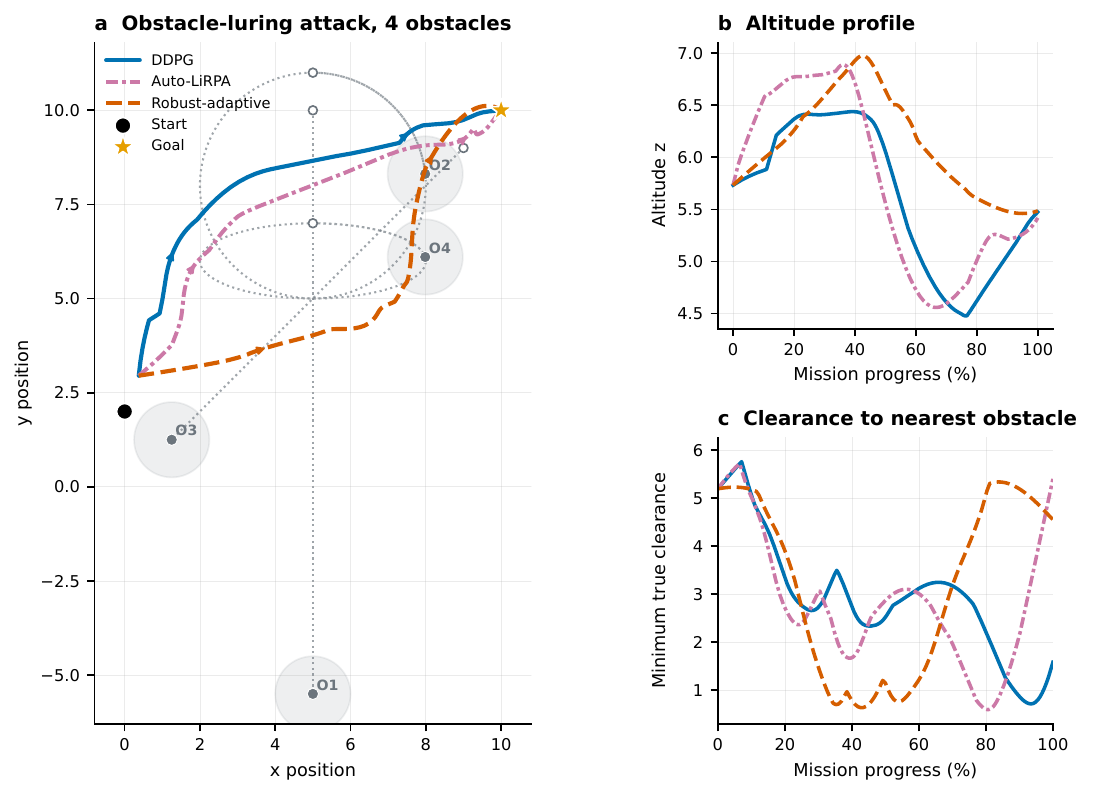}
			\caption{Representative UAV trajectories, altitude profiles, and minimum obstacle clearances under a dynamic obstacle-luring spoofing attack, illustrating the ability of the curriculum-adapted policy to maintain stable navigation despite corrupted obstacle perception.}
			\label{figure_dynamic_trajectory}
		\end{figure}		
		To directly quantify the attack objective, we measure the perception-reality divergence, defined as the difference between the minimum perceived obstacle clearance and the minimum true obstacle clearance shown in Figure \ref{figure_reality_divergence}. Larger values indicate that the agent believes it is farther from obstacles than it actually is. The curriculum-adapted policy consistently exhibits the smallest divergence across all traffic densities, whereas NR-MDP, PR-MDP, and AutoLiRPA display substantially larger discrepancies. This result suggests that the proposed adaptation mechanism is less susceptible to perception manipulation and maintains a more accurate internal representation of environmental risk despite corrupted observations.
		
		The trajectory analysis in Figure \ref{figure_dynamic_trajectory} provides further insight into this behaviour. Following attack activation, the baseline policies exhibit larger deviations in obstacle clearance and trajectory geometry, indicating stronger coupling between spoofed observations and navigation decisions. In contrast, the curriculum-adapted policy maintains smoother altitude profiles and more stable clearance margins throughout the mission, despite being exposed only to synthetic adversarial perturbations during training. These findings are particularly significant because the attack mechanism differs fundamentally from the perturbations used during adaptation. The policy is adapted using gradient-based observation perturbations, yet evaluated against a state-dependent GNSS spoofing attack operating at the sensing layer. Rather than merely increasing resistance to a specific perturbation model, the proposed framework maintains efficient navigation and accurate risk assessment under evolving perception-driven attacks, both of which are essential requirements for trustworthy autonomous operation in contested airspace.

%	\begin{table}[htbp]
%		\centering
%		\caption{The mean reward of the vanilla DDPG, NR-MDP, PR-MDP, SA-MDP, Adversarial Meta Learning, and Antifragile algorithms along with standard deviations for different adversarial and spoofing attacks.}
%		\label{reward_table}
%		\scalebox{0.7}{
%			\begin{tabular}{|l|c|c|c|}
%				\hline
%				\textbf{Models} & \textbf{PGD $l_{\infty}$ attack $\epsilon = 4.0$} & 
%				\textbf{PGD $l_{\infty}$ attack $\epsilon = 5.0$} & 
%				\textbf{Spoofing Attack} \\ 
%				\hline
%				Vanilla DDPG & $-52.90 \pm 12.44$ & $-50.56 \pm 9.36$ & $-71.22 \pm 8.29$ \\ 
%				\hline
%				NR-MDP \cite{tessler2019action} & $-89.01 \pm 51.02$ & $-73.16 \pm 20.15$ & $-307.1 \pm 51.32$ \\ 
%				\hline
%				PR-MDP \cite{tessler2019action} & $-74.27 \pm 23.07$ & $-72.54 \pm 22.44$ & $-248.0 \pm 39.28$ \\ 
%				\hline
%				SA-MDP \cite{zhang2020robust} & $-55.71 \pm 14.37$ & $-54.35 \pm 12.97$ & $-363.09 \pm 55.14$ \\ 
%				\hline
%				Adversarial Meta \cite{chen2021adaptive} & $-59.01 \pm 10.18$ & $-59.91 \pm 12.22$ & $-327.76 \pm 60.32$ \\ 
%				\hline
%				Antifragile & $\mathbf{-49.36 \pm 10.66}$ & $\mathbf{-47.80 \pm 8.86}$ & $\mathbf{-60.05 \pm 11.42}$ \\ 
%				\hline
%		\end{tabular}}
%	\end{table}
	\subsection{Ablation Study}
	\rev{\subsubsection{Conflicts Comparison with Various Curriculum Update}}
	\begin{figure}[thpb]
		\centering
		\includegraphics[scale=0.18]{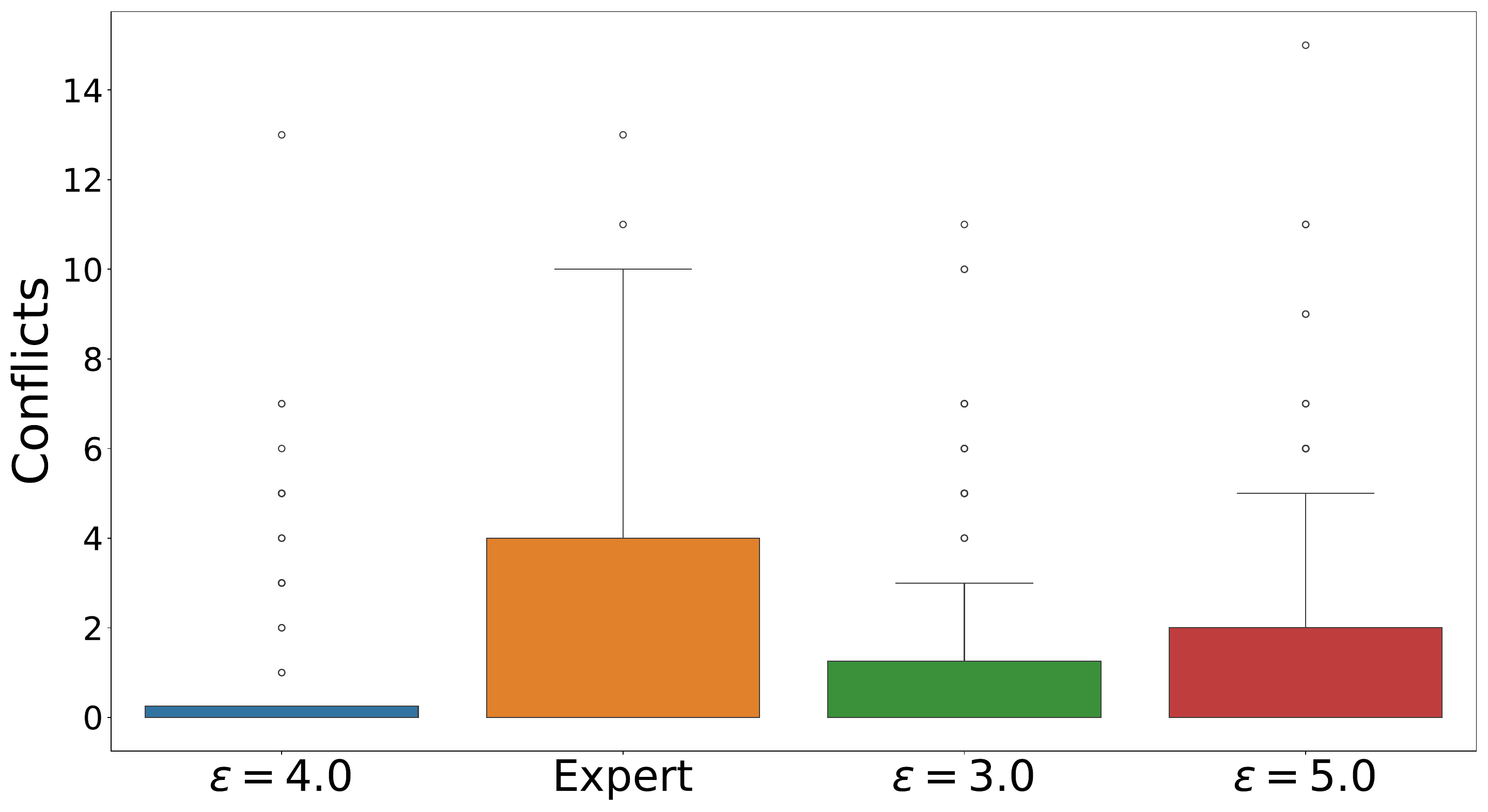}
		\caption{\rev{Comparing the conflict distribution with expert policy and the curriculum update stopping condition against spoofing attack and 8 dynamic obstacles across 100 episodes.}}
		\label{figure_ablation_conflicts}
	\end{figure}
	\rev{We compare the impact of different curriculum stopping conditions of $\epsilon$ using the adaptation scheme described in Theorem \ref{theorem_main} and Algorithm \ref{alg:adv_adaptation}. Figure \ref{figure_ablation_conflicts} reports the conflict distribution for the policies trained with stopping conditions $\epsilon = 3.0,\, 4.0,\, 5.0$, alongside the expert policy, under a constant spoofing bias of $0.05$ units and eight dynamic obstacles across $100$ testing episodes.} \rev{The expert policy exhibits both the highest average number of conflicts and the largest variance, indicating poor robustness against spoofing despite being optimal under nominal conditions. Although the mean conflict values for the policies trained at $\epsilon=3.0$ and $\epsilon=5.0$ remain close to the one obtained at $\epsilon=4.0$, however latter policy demonstrate lower variance and the lowest median conflict count and the narrowest interquartile range, confirming a more reliable deconfliction behaviour against obstacles. While the mean reward for the policies at $\epsilon=3.0,\, 4.0,\, 5.0$ remains comparable, the expert policy exhibits significantly lower and more volatile rewards, highlighting its vulnerability to spoofing attacks. Hence based on the deconfliction behaivour, the policy obtained at $\epsilon=4.0$ consistently outperforms the existing policies and these results validate the choice of $\epsilon=4.0$ as the optimal curriculum stopping point.}
	\subsection{Sensitivity Analysis}
	\rev{\begin{figure}[thpb]
			\centering
			\includegraphics[scale=0.18]{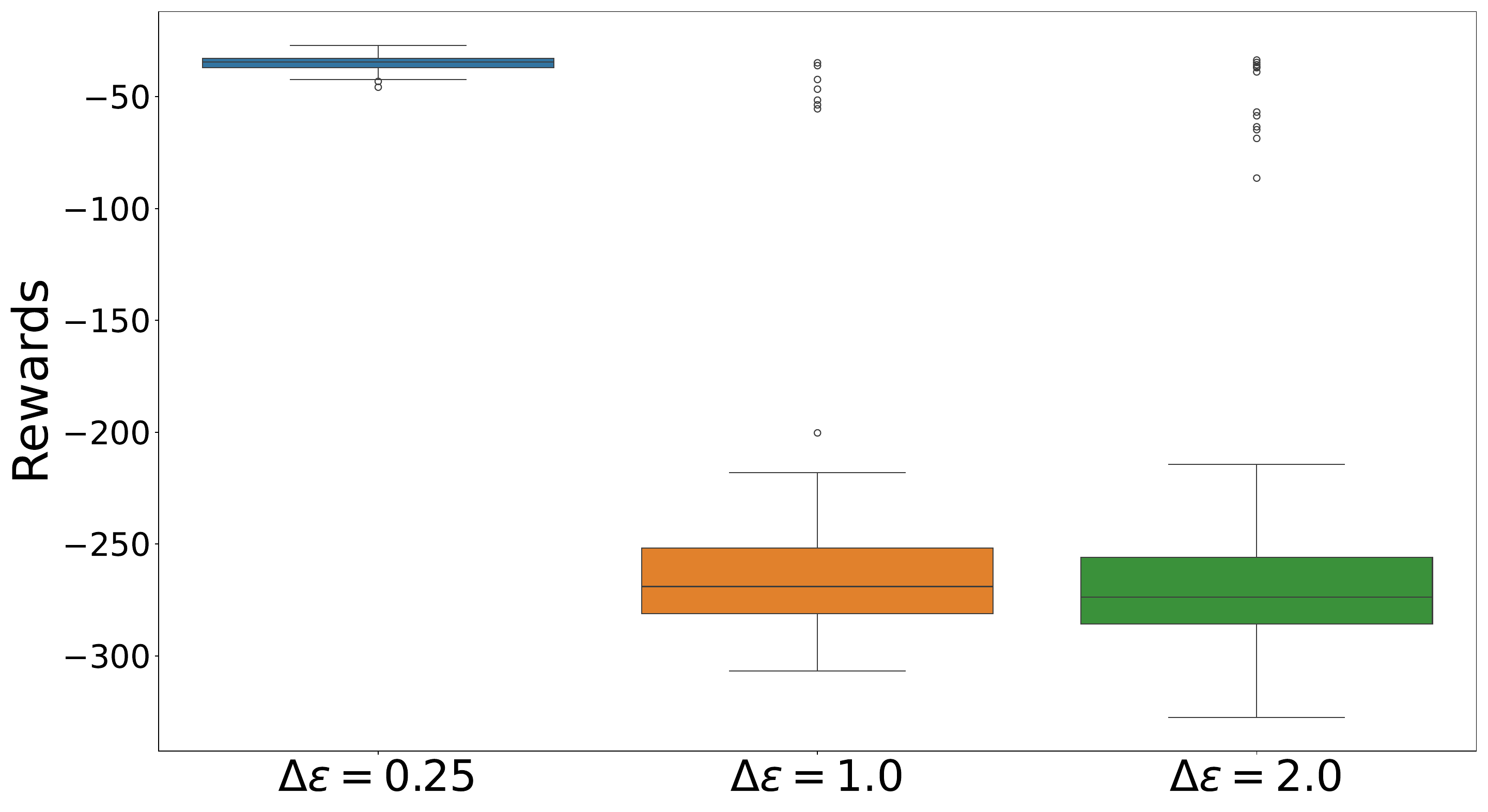}
			\caption{\rev{Sensitivity analysis comparing the performance of the curriculum adaptation trained with different adversarial strength. The analysis is conducted against spoofing attack over 100 testing episodes.}}
			\label{sensitivity_curriculum}
	\end{figure}}
	\rev{We conducted sensitivity analysis with respect to various curriculum stages, where $\Delta \epsilon = \left( 1.0, 2.0 \right ) $ while comparing with the existing curriculum of $\Delta \epsilon = 0.25$, and the trained curriculum is tested against unseen spoofing attack with the bias of 0.05 and 8 dynamic obstacles. We see the rewards in Figure \ref{sensitivity_curriculum} for the policies trained with different curriculum bounds. A higher reward for the policies obtained from $\Delta \epsilon = 0.25$, shows that smaller curriculum step size produces smoother TD-alignment between stages, with $\Delta \epsilon = 0.25$ yielding the highest and most stable rewards against unseen spoofing attacks. Larger increments ($\Delta \epsilon = 1.0, 2.0$) force the critic to align across widely separated perturbation regimes, causing abrupt TD-error shifts that violate the gradual progression required by Theorem \ref{theorem_main} and induce a mode-collapse-like behaviour where the value function collapses into overly conservative solutions. In contrast, $\Delta \epsilon = 0.25$ maintains small TD-space transitions, prevents value-mode collapse, and enables stable adaptation, explaining its superior reward performance.}

	\section{Conclusion} \label{concl_section}
	This paper investigated whether adaptation learned under one class of adversarial perturbations can generalize to previously unseen cyber-physical attacks. To address this challenge, we proposed a curriculum-guided adaptation framework that progressively exposes a robust RL policy to increasing adversarial perturbations while preserving value-function consistency through temporal-difference (TD) error distribution alignment. We established a TD-space generalization certificate linking adaptation performance to robustness transfer across attack domains. Experimental evaluation in a UAV deconfliction environment demonstrated that policies adapted using synthetic gradient-based perturbations achieve improved mission performance, navigation efficiency, and decondliction behaivour under previously unseen fixed and dynamic GNSS spoofing attacks. Collectively, the results suggest that preserving value-distribution consistency during adaptation provides a principled mechanism for improving the resilience of autonomous decision-making systems under unknown cyber-physical threats.
	\begin{itemize}
		\item Adversarial observation shifts can be interpreted through their effect on TD-error distributions, providing a value-distribution centric perspective of robustness in RL.
		\item Curriculum-guided adaptation reduces value-distribution drift across progressively stronger perturbations, mitigating catastrophic degradation under adversarial conditions.
		\item Robustness acquired from synthetic gradient-based perturbations can transfer to physically distinct GNSS spoofing attacks when TD-error distributions remain sufficiently aligned.
		\item In UAV deconfliction scenarios, the proposed framework consistently improves mission completion, navigation efficiency, and conflict avoidance under both fixed and dynamic spoofing attacks compared with standard and robust RL baselines.
	\end{itemize}
	The experimental conclusions are limited to the simulated UAV deconfliction environment and the fixed and dynamic GNSS spoofing attacks evaluated in this study. The comparative results also reflect the specific baseline implementations and training configurations considered here. Evaluation across additional sensing modalities, independently trained seeds, and alternative adaptation strategies remains an important direction for future work.
	
	%\bibliographystyle{IEEEtran}
	%\bibliography{bibliography}
	
	\printbibliography

@article{berkenkamp2017safe,
  title={Safe model-based reinforcement learning with stability guarantees},
  author={Berkenkamp, Felix and Turchetta, Matteo and Schoellig, Angela and Krause, Andreas},
  journal={Advances in neural information processing systems},
  volume={30},
  year={2017}
}

@article{yu2023safe,
  title={Safe model-based reinforcement learning with an uncertainty-aware reachability certificate},
  author={Yu, Dongjie and Zou, Wenjun and Yang, Yujie and Ma, Haitong and Li, Shengbo Eben and Yin, Yuming and Chen, Jianyu and Duan, Jingliang},
  journal={IEEE Transactions on Automation Science and Engineering},
  volume={21},
  number={3},
  pages={4129--4142},
  year={2023},
  publisher={IEEE}
}

@inproceedings{wang2021meta,
  title={Meta-adversarial inverse reinforcement learning for decision-making tasks},
  author={Wang, Pin and Li, Hanhan and Chan, Ching-Yao},
  booktitle={2021 IEEE international conference on robotics and automation (ICRA)},
  pages={12632--12638},
  year={2021},
  organization={IEEE}
}

@article{turchetta2016safe,
  title={Safe exploration in finite markov decision processes with gaussian processes},
  author={Turchetta, Matteo and Berkenkamp, Felix and Krause, Andreas},
  journal={Advances in neural information processing systems},
  volume={29},
  year={2016}
}

@article{panda6755038meta,
  title={Meta Policy Switching for Resilient UAV Navigation in Adversarial Airspace},
  author={Panda, Deepak Kumar and Guo, Weisi},
  journal={Available at SSRN 6755038}
}

@article{korbak2022reinforcement,
  title={On reinforcement learning and distribution matching for fine-tuning language models with no catastrophic forgetting},
  author={Korbak, Tomasz and Elsahar, Hady and Kruszewski, Germ{\'a}n and Dymetman, Marc},
  journal={Advances in Neural Information Processing Systems},
  volume={35},
  pages={16203--16220},
  year={2022}
}

@article{dasgupta2022sensor,
  title={A sensor fusion-based GNSS spoofing attack detection framework for autonomous vehicles},
  author={Dasgupta, Sagar and Rahman, Mizanur and Islam, Mhafuzul and Chowdhury, Mashrur},
  journal={IEEE Transactions on Intelligent Transportation Systems},
  volume={23},
  number={12},
  pages={23559--23572},
  year={2022},
  publisher={IEEE}
}

@article{psiaki2016gnss,
  title={GNSS spoofing and detection},
  author={Psiaki, Mark L and Humphreys, Todd E},
  journal={Proceedings of the IEEE},
  volume={104},
  number={6},
  pages={1258--1270},
  year={2016},
  publisher={IEEE}
}

@article{chen2020distributionally,
  title={Distributionally robust learning},
  author={Chen, Ruidi and Paschalidis, Ioannis Ch and others},
  journal={Foundations and Trends{\textregistered} in Optimization},
  volume={4},
  number={1-2},
  pages={1--243},
  year={2020},
  publisher={Now Publishers, Inc.}
}

@inproceedings{liu2019wasserstein,
  title={Wasserstein gan with quadratic transport cost},
  author={Liu, Huidong and Gu, Xianfeng and Samaras, Dimitris},
  booktitle={Proceedings of the IEEE/CVF international conference on computer vision},
  pages={4832--4841},
  year={2019}
}

@article{khetarpal2022towards,
  title={Towards continual reinforcement learning: A review and perspectives},
  author={Khetarpal, Khimya and Riemer, Matthew and Rish, Irina and Precup, Doina},
  journal={Journal of Artificial Intelligence Research},
  volume={75},
  pages={1401--1476},
  year={2022}
}

@article{wang2023curriculum,
  title={Curriculum reinforcement learning from avoiding collisions to navigating among movable obstacles in diverse environments},
  author={Wang, Hsueh-Cheng and Huang, Siao-Cing and Huang, Po-Jui and Wang, Kuo-Lun and Teng, Yi-Chen and Ko, Yu-Ting and Jeon, Dongsuk and Wu, I-Chen},
  journal={IEEE Robotics and Automation Letters},
  volume={8},
  number={5},
  pages={2740--2747},
  year={2023},
  publisher={IEEE}
}

@article{razzaghi2024survey,
  title={A survey on reinforcement learning in aviation applications},
  author={Razzaghi, Pouria and Tabrizian, Amin and Guo, Wei and Chen, Shulu and Taye, Abenezer and Thompson, Ellis and Bregeon, Alexis and Baheri, Ali and Wei, Peng},
  journal={Engineering Applications of Artificial Intelligence},
  volume={136},
  pages={108911},
  year={2024},
  publisher={Elsevier}
}

@article{klink2024benefit,
  title={On the benefit of optimal transport for curriculum reinforcement learning},
  author={Klink, Pascal and D'Eramo, Carlo and Peters, Jan and Pajarinen, Joni},
  journal={IEEE Transactions on Pattern Analysis and Machine Intelligence},
  volume={46},
  number={11},
  pages={7191--7204},
  year={2024},
  publisher={IEEE}
}

@inproceedings{nayyar2025autonomous,
  title={Autonomous option invention for continual hierarchical reinforcement learning and planning},
  author={Nayyar, Rashmeet Kaur and Srivastava, Siddharth},
  booktitle={Proceedings of the AAAI Conference on Artificial Intelligence},
  volume={39},
  number={18},
  pages={19642--19650},
  year={2025}
}

@article{malagon2025self,
  title={Self-composing policies for scalable continual reinforcement learning},
  author={Malagon, Mikel and Ceberio, Josu and Lozano, Jose A},
  journal={arXiv preprint arXiv:2506.14811},
  year={2025}
}

@article{rusu2016progressive,
  title={Progressive neural networks},
  author={Rusu, Andrei A and Rabinowitz, Neil C and Desjardins, Guillaume and Soyer, Hubert and Kirkpatrick, James and Kavukcuoglu, Koray and Pascanu, Razvan and Hadsell, Raia},
  journal={arXiv preprint arXiv:1606.04671},
  year={2016}
}

@article{elelimy2025rethinking,
  title={Rethinking the foundations for continual reinforcement learning},
  author={Elelimy, Esraa and Szepesvari, David and White, Martha and Bowling, Michael},
  journal={arXiv preprint arXiv:2504.08161},
  year={2025}
}

@inproceedings{pollatos2025corruption,
  title={On corruption-robustness in performative reinforcement learning},
  author={Pollatos, Vasilis and Mandal, Debmalya and Radanovic, Goran},
  booktitle={Proceedings of the AAAI Conference on Artificial Intelligence},
  volume={39},
  number={19},
  pages={19939--19947},
  year={2025}
}

@article{shi2024distributionally,
  title={Distributionally robust model-based offline reinforcement learning with near-optimal sample complexity},
  author={Shi, Laixi and Chi, Yuejie},
  journal={Journal of Machine Learning Research},
  volume={25},
  number={200},
  pages={1--91},
  year={2024}
}

@article{clavier2024near,
  title={Near-Optimal Distributionally Robust Reinforcement Learning with General $ L\_p $ Norms},
  author={Clavier, Pierre and Shi, Laixi and Le Pennec, Erwan and Mazumdar, Eric and Wierman, Adam and Geist, Matthieu},
  journal={Advances in Neural Information Processing Systems},
  volume={37},
  pages={1750--1810},
  year={2024}
}

@article{lu2024distributionally,
  title={Distributionally robust reinforcement learning with interactive data collection: Fundamental hardness and near-optimal algorithms},
  author={Lu, Miao and Zhong, Han and Zhang, Tong and Blanchet, Jose},
  journal={Advances in Neural Information Processing Systems},
  volume={37},
  pages={12528--12580},
  year={2024}
}

@inproceedings{duan2016benchmarking,
  title={Benchmarking deep reinforcement learning for continuous control},
  author={Duan, Yan and Chen, Xi and Houthooft, Rein and Schulman, John and Abbeel, Pieter},
  booktitle={International conference on machine learning},
  pages={1329--1338},
  year={2016},
  organization={PMLR}
}

@inproceedings{castro2020scalable,
  title={Scalable methods for computing state similarity in deterministic markov decision processes},
  author={Castro, Pablo Samuel},
  booktitle={Proceedings of the AAAI Conference on Artificial Intelligence},
  volume={34},
  number={06},
  pages={10069--10076},
  year={2020}
}

@article{ferns2011bisimulation,
  title={Bisimulation metrics for continuous Markov decision processes},
  author={Ferns, Norm and Panangaden, Prakash and Precup, Doina},
  journal={SIAM Journal on Computing},
  volume={40},
  number={6},
  pages={1662--1714},
  year={2011},
  publisher={SIAM}
}

@article{wang2022dirichlet,
  title={A Dirichlet process mixture of robust task models for scalable lifelong reinforcement learning},
  author={Wang, Zhi and Chen, Chunlin and Dong, Daoyi},
  journal={IEEE Transactions on Cybernetics},
  volume={53},
  number={12},
  pages={7509--7520},
  year={2022},
  publisher={IEEE}
}

@inproceedings{luo2018adaptive,
  title={Adaptive Gradient Methods with Dynamic Bound of Learning Rate},
  author={Luo, Liangchen and Xiong, Yuanhao and Liu, Yan and Sun, Xu},
  booktitle={International Conference on Learning Representations},
  year={2018}
}

@article{huang2021decentralized,
  title={Decentralized autonomous navigation of a UAV network for road traffic monitoring},
  author={Huang, Hailong and Savkin, Andrey V and Huang, Chao},
  journal={IEEE Transactions on Aerospace and Electronic Systems},
  volume={57},
  number={4},
  pages={2558--2564},
  year={2021},
  publisher={IEEE}
}

@article{liu2022industrial,
  title={Industrial UAV-based unsupervised domain adaptive crack recognitions: From database towards real-site infrastructural inspections},
  author={Liu, Kangcheng and Chen, Ben M},
  journal={IEEE Transactions on Industrial Electronics},
  volume={70},
  number={9},
  pages={9410--9420},
  year={2022},
  publisher={IEEE}
}

@article{leonardi2017ads,
  title={ADS-B jamming mitigation: a solution based on a multichannel receiver},
  author={Leonardi, Mauro and Piracci, Emilio and Galati, Gaspare},
  journal={IEEE Aerospace and Electronic Systems Magazine},
  volume={32},
  number={11},
  pages={44--51},
  year={2017},
  publisher={IEEE}
}

@inproceedings{ying2019detecting,
  title={Detecting ADS-B spoofing attacks using deep neural networks},
  author={Ying, Xuhang and Mazer, Joanna and Bernieri, Giuseppe and Conti, Mauro and Bushnell, Linda and Poovendran, Radha},
  booktitle={2019 IEEE conference on communications and network security (CNS)},
  pages={187--195},
  year={2019},
  organization={IEEE}
}

@article{kim2017ads,
  title={ADS-B vulnerabilities and a security solution with a timestamp},
  author={Kim, Yoohwan and Jo, Ju-Yeon and Lee, Sungchul},
  journal={IEEE Aerospace and Electronic Systems Magazine},
  volume={32},
  number={11},
  pages={52--61},
  year={2017},
  publisher={IEEE}
}

@article{zhang2020robust,
  title={Robust deep reinforcement learning against adversarial perturbations on state observations},
  author={Zhang, Huan and Chen, Hongge and Xiao, Chaowei and Li, Bo and Liu, Mingyan and Boning, Duane and Hsieh, Cho-Jui},
  journal={Advances in Neural Information Processing Systems},
  volume={33},
  pages={21024--21037},
  year={2020}
}

@inproceedings{wang2019generalization,
  title={On the generalization gap in reparameterizable reinforcement learning},
  author={Wang, Huan and Zheng, Stephan and Xiong, Caiming and Socher, Richard},
  booktitle={International Conference on Machine Learning},
  pages={6648--6658},
  year={2019},
  organization={PMLR}
}

@article{panda2025real,
  title={Real-Time Bayesian Detection of Drift-Evasive GNSS Spoofing in Reinforcement Learning Based UAV Deconfliction},
  author={Panda, Deepak Kumar and Guo, Weisi},
  journal={arXiv preprint arXiv:2507.11173},
  year={2025}
}

@article{panda2024action,
  title={Action Robust Reinforcement Learning for Air Mobility Deconfliction Against Conflict Induced Spoofing},
  author={Panda, Deepak Kumar and Guo, Weisi},
  journal={IEEE Transactions on Intelligent Transportation Systems},
  volume={25},
  number={12},
  pages={21343--21355},
  year={2024},
  publisher={IEEE}
}

@inproceedings{ganin2015unsupervised,
  title={Unsupervised domain adaptation by backpropagation},
  author={Ganin, Yaroslav and Lempitsky, Victor},
  booktitle={International conference on machine learning},
  pages={1180--1189},
  year={2015},
  organization={PMLR}
}

@article{huang2022curriculum,
  title={Curriculum reinforcement learning using optimal transport via gradual domain adaptation},
  author={Huang, Peide and Xu, Mengdi and Zhu, Jiacheng and Shi, Laixi and Fang, Fei and Zhao, Ding},
  journal={Advances in Neural Information Processing Systems},
  volume={35},
  pages={10656--10670},
  year={2022}
}

@inproceedings{xu2020forget,
  title={Forget me not: Reducing catastrophic forgetting for domain adaptation in reading comprehension},
  author={Xu, Ying and Zhong, Xu and Yepes, Antonio Jose Jimeno and Lau, Jey Han},
  booktitle={2020 International joint conference on neural networks (IJCNN)},
  pages={1--8},
  year={2020},
  organization={IEEE}
}

@article{bhardwaj2019adaptively,
  title={Adaptively preconditioned stochastic gradient Langevin dynamics},
  author={Bhardwaj, Chandrasekaran Anirudh},
  journal={arXiv preprint arXiv:1906.04324},
  year={2019}
}

@inproceedings{zou2019sufficient,
  title={A sufficient condition for convergences of adam and rmsprop},
  author={Zou, Fangyu and Shen, Li and Jie, Zequn and Zhang, Weizhong and Liu, Wei},
  booktitle={Proceedings of the IEEE/CVF Conference on computer vision and pattern recognition},
  pages={11127--11135},
  year={2019}
}

@inproceedings{xing2021domain,
  title={Domain adaptation in reinforcement learning via latent unified state representation},
  author={Xing, Jinwei and Nagata, Takashi and Chen, Kexin and Zou, Xinyun and Neftci, Emre and Krichmar, Jeffrey L},
  booktitle={Proceedings of the AAAI Conference on Artificial Intelligence},
  volume={35},
  number={12},
  pages={10452--10459},
  year={2021}
}

@article{long2018conditional,
  title={Conditional adversarial domain adaptation},
  author={Long, Mingsheng and Cao, Zhangjie and Wang, Jianmin and Jordan, Michael I},
  journal={Advances in neural information processing systems},
  volume={31},
  year={2018}
}

@inproceedings{sankaranarayanan2018generate,
  title={Generate to adapt: Aligning domains using generative adversarial networks},
  author={Sankaranarayanan, Swami and Balaji, Yogesh and Castillo, Carlos D and Chellappa, Rama},
  booktitle={Proceedings of the IEEE Conference on Computer Vision and Pattern Recognition},
  pages={8503--8512},
  year={2018}
}

@book{rostami2021transfer,
  title={Transfer learning through embedding spaces},
  author={Rostami, Mohammad},
  year={2021},
  publisher={CRC Press}
}

@article{rostami2021lifelong,
  title={Lifelong domain adaptation via consolidated internal distribution},
  author={Rostami, Mohammad},
  journal={Advances in Neural Nnformation Processing Systems},
  volume={34},
  pages={11172--11183},
  year={2021}
}

@inproceedings{zhang2021adaptive,
  title={Adaptive interfered fluid dynamic system algorithm based on deep reinforcement learning framework},
  author={Zhang, Yunfei and Wang, Honglun},
  booktitle={International Conference on Autonomous Unmanned Systems},
  pages={1388--1397},
  year={2021},
  organization={Springer}
}

@article{ince2024sense,
  title={Sense and Avoid Considerations for Safe sUAS Operations in Urban Environments},
  author={Ince, Bilkan and Martinez, Victor Celdran and Selvam, Praveen Kumar and Petrunin, Ivan and Seo, Minguk and Tsourdos, Antonios},
  journal={IEEE Aerospace and Electronic Systems Magazine},
  year={2024},
  publisher={IEEE}
}

@inproceedings{carlini2017towards,
  title={Towards evaluating the robustness of neural networks},
  author={Carlini, Nicholas and Wagner, David},
  booktitle={2017 IEEE Symposium on Security and Privacy},
  pages={39--57},
  year={2017},
  organization={Ieee}
}

@article{perrusquia2024uncovering,
  title={Uncovering drone intentions using control physics informed machine learning},
  author={Perrusqu{\'\i}a, Adolfo and Guo, Weisi and Fraser, Benjamin and Wei, Zhuangkun},
  journal={Communications Engineering},
  volume={3},
  number={1},
  pages={36},
  year={2024},
  publisher={Nature Publishing Group UK London}
}

@inproceedings{li2023uav,
  title={A UAV Path Planning Method in Three-Dimensional Urban Airspace based on Safe Reinforcement Learning},
  author={Li, Yan and Zhang, Xuejun and Zhu, Yuanjun and Gao, Ziang},
  booktitle={2023 IEEE/AIAA 42nd Digital Avionics Systems Conference (DASC)},
  pages={1--7},
  year={2023},
  organization={IEEE}
}

@inproceedings{hsieh2019finding,
  title={Finding mixed nash equilibria of generative adversarial networks},
  author={Hsieh, Ya-Ping and Liu, Chen and Cevher, Volkan},
  booktitle={International Conference on Machine Learning},
  pages={2810--2819},
  year={2019},
  organization={PMLR}
}

@article{kamalaruban2020robust,
  title={Robust reinforcement learning via adversarial training with langevin dynamics},
  author={Kamalaruban, Parameswaran and Huang, Yu-Ting and Hsieh, Ya-Ping and Rolland, Paul and Shi, Cheng and Cevher, Volkan},
  journal={Advances in Neural Information Processing Systems},
  volume={33},
  pages={8127--8138},
  year={2020}
}

@inproceedings{tessler2019action,
  title={Action robust reinforcement learning and applications in continuous control},
  author={Tessler, Chen and Efroni, Yonathan and Mannor, Shie},
  booktitle={International Conference on Machine Learning},
  pages={6215--6224},
  year={2019},
  organization={PMLR}
}

@article{ilahi2021challenges,
  title={Challenges and countermeasures for adversarial attacks on deep reinforcement learning},
  author={Ilahi, Inaam and Usama, Muhammad and Qadir, Junaid and Janjua, Muhammad Umar and Al-Fuqaha, Ala and Hoang, Dinh Thai and Niyato, Dusit},
  journal={IEEE Transactions on Artificial Intelligence},
  volume={3},
  number={2},
  pages={90--109},
  year={2021},
  publisher={IEEE}
}

@article{zeng2026gnss,
  title={GNSS Jamming and Spoofing Threats in UAV Navigation: Countermeasure Status and Challenges},
  author={Zeng, Yejia and Lu, Zukun and Zhao, Xiaoyu and Xiao, Zhu and Ni, Shaojie and Han, Zhu and Li, Keqin},
  journal={IEEE Communications Surveys \& Tutorials},
  year={2026},
  publisher={IEEE}
}
	\clearpage
	
	\appendices
%	\renewcommand\thesubsection{\thesection.\Roman{subsection}}
%	\counterwithin{equation}{section}
	\section{Notation Table}	
	\begin{center}
		\renewcommand{\arraystretch}{1.08}
		\setlength{\tabcolsep}{3pt}
		\footnotesize
		\begin{tabularx}{\columnwidth}{p{2.5cm} X}
			\toprule
			\mrev{\textbf{Symbol}} & \mrev{\textbf{Definition}} \\
			\midrule
			
			\multicolumn{2}{l}{\mrev{\textit{UAV Deconfliction Environment}}} \\
			\midrule
			
			\mrev{$P^t,P_g,P_{\mathrm{obs}}^t$}
			& \mrev{UAV, goal, and obstacle positions} \\
			
			\mrev{$v(P),\bar{v}(P),v_{\mathrm{obs}}(P)$}
			& \mrev{Goal-directed, modulated, and obstacle velocities} \\
			
			\mrev{$\Phi^t$}
			& \mrev{UAV observation/state} \\
			
			\mrev{$a^t=(\rho,\varsigma,\vartheta)$}
			& \mrev{Navigation action: attraction, repulsion, and heading} \\
			
			\mrev{$R^t$}
			& \mrev{Composite navigation reward} \\
			
			\mrev{$R_{\mathrm{col}},R_{\mathrm{path}},R_{\mathrm{threat}}$}
			& \mrev{Collision, path, and obstacle-threat rewards} \\
			
			\mrev{$\lambda_1,\lambda_2,\lambda_3$}
			& \mrev{Reward weighting coefficients} \\
			
			\midrule
			\multicolumn{2}{l}{\mrev{\textit{Expert Critic Design}}} \\
			\midrule
			
			\mrev{$\pi_{\theta}^{\mathrm{agent}},\pi_{\omega}^{\mathrm{adv}}$}
			& \mrev{Agent and adversarial policies} \\
			
			\mrev{$Q_{\phi},Q_{\mathrm{AR}}$}
			& \mrev{Nominal and action-robust critics} \\
			
			\mrev{$\alpha$}
			& \mrev{Action-robustness parameter} \\
			
			\mrev{$\gamma$}
			& \mrev{Discount factor} \\
			
			\mrev{$\mathcal{M}_2$}
			& \mrev{Two-player action-robust MDP} \\
			
			\mrev{$T_2,\mathcal{R},p_0$}
			& \mrev{Transition, reward, and initial-state distribution} \\
			
			\mrev{$J(\theta,\omega)$}
			& \mrev{Agent-adversary expected return objective} \\
			
			\midrule
			\multicolumn{2}{l}{\mrev{\textit{Curriculum-Guided Adaptation}}} \\
			\midrule
			
			\mrev{$\epsilon_k,\Delta\epsilon$}
			& \mrev{Perturbation strength and curriculum increment} \\
			
			\mrev{$\Phi_{\mathrm{adv}}^{\epsilon}$}
			& \mrev{Adversarially perturbed observation} \\
			
			\mrev{$\mathbf{TD}^{\pi}_{\epsilon}$}
			& \mrev{TD-error distribution under perturbation $\epsilon$} \\
			
			\mrev{$\mathcal{W}_1(\cdot,\cdot)$}
			& \mrev{Wasserstein-1 distance} \\
			
			\mrev{$f_{\pi}^{\epsilon}$}
			& \mrev{TD-space catastrophic-forgetting measure} \\
			
			\mrev{$P_k$}
			& \mrev{Transition kernel at curriculum stage $k$} \\
			
			\mrev{$\beta_k$}
			& \mrev{Inter-stage TD-distribution discrepancy} \\
			
			\mrev{$K$}
			& \mrev{Final curriculum stage} \\
			
			\midrule
			\multicolumn{2}{l}{%
				\mrev{\textit{Algorithmic Generalization and Test-Time Attack Evaluation}}} \\
			\midrule
			
			\mrev{$\mathfrak{A}$}
			& \mrev{Unseen test-time attack} \\
			
			\mrev{$\delta_{\mathfrak{A}}$}
			& \mrev{Test-to-curriculum TD-space discrepancy} \\
			
			\mrev{$\mathrm{Rad}_k$}
			& \mrev{Rademacher-complexity term} \\
			
			\mrev{$\delta$}
			& \mrev{Generalization confidence parameter} \\
			
			\mrev{$\mathfrak{b}^t,\dot{\mathfrak{b}}^t$}
			& \mrev{GNSS spoofing bias and bias rate} \\
			
			\mrev{$\zeta_p,\zeta_v$}
			& \mrev{Maximum position and velocity biases} \\
			
			\mrev{$d_s,d_a$}
			& \mrev{Safety and attack-activation radii} \\
			
			\bottomrule
		\end{tabularx}
	\end{center}
	
	\section{Interfered Fluid Dynamic System} \label{IFDS}
	The interfered flow field causes a change in the initial flow field due to the presence of dynamic obstacles. The 3D obstacle is designed as a standard convex polyhedron, as:
	\begin{equation} \label{eq:1A}
		\Gamma\left ( \mathbf{P} \right ) = \left (  \frac{x-x_0}{\breve{a}}\right )^{2\breve{p}} + \left (  \frac{y-y_0}{\breve{b}}\right )^{2\breve{q}} + \left (  \frac{z-z_0}{\breve{c}}\right )^{2\breve{r}}. 
	\end{equation}
	In this case, $\breve{p}$, $\breve{q}$, $\breve{r}$, $\breve{a}$, $\breve{b}$, $\breve{c}$ determine the shape of the convex polyhedron of the obstacle, and $\left ( x_0, y_0, z_0 \right )$ determine centre of the convex polyhedron. If we consider $N$ obstacles in the space, the interfered flow field is represented by a disturbance matrix as: 
	\begin{equation} \label{eq:2A}
		\overline{M}\left ( \mathbf{P} \right ) = \sum_{n=1}^{N} \mathfrak{w}_n \left ( \mathbf{P} \right ) M_n \left ( P \right ),
	\end{equation}
	where $\mathfrak{w}$ is the disturbance weighting factor for the $N$ obstacles defined as: 
	\begin{equation} \label{eq:3A}
		\mathfrak{w}_n \left ( \mathbf{P} \right ) = \left\{\begin{matrix}
			1 \; \; \; \;  N = 1, \\
			\prod_{i=1, i \neq n}^{N} \frac{\Gamma_i \left ( P \right ) - 1}{\Gamma_i \left ( P \right ) - 1 + \Gamma_n \left ( P \right ) - 1}.
		\end{matrix}\right. 
	\end{equation}
	Here we define the radial normal vector along the 3D obstacle $\mathbf{t}_n$ as:
	\begin{equation} \label{eq:4A}
		t_n \left ( \mathbf{P} \right ) = \begin{bmatrix}
			\frac{\partial \Gamma_n \left ( \mathbf{P} \right )}{\partial x} & \frac{\partial \Gamma_n \left ( \mathbf{P} \right )}{\partial y} & \frac{\partial \Gamma_n \left ( \mathbf{P} \right )}{\partial z}  \\
		\end{bmatrix}.
	\end{equation}
	We consider two vectors $h_{n,1} \left ( \mathbf{P} \right )$, $h_{n,2} \left ( \mathbf{P} \right )$ which are orthogonal to the radial vector $\mathbf{t}_n$ and also mutually orthogonal to each other, given as:
	\begin{equation} \label{eq:5A}
		\begin{gathered}
			h_{n,1} \left ( \mathbf{P} \right ) = 
			\begin{bmatrix}
				\frac{\partial \Gamma_n \left ( \mathbf{P} \right )}{\partial y} \\  
				-\frac{\partial \Gamma_n \left ( \mathbf{P} \right )}{\partial x}\\
				0  \\
			\end{bmatrix}^T, \\
			h_{n,2} \left ( \mathbf{P} \right ) = \begin{bmatrix}
				\frac{\partial \Gamma_n \left ( \mathbf{P} \right )}{\partial x}  \cdot \frac{\partial \Gamma_n \left ( \mathbf{P} \right )}{\partial z}  \\  
				\frac{\partial \Gamma_n \left ( \mathbf{P} \right )}{\partial x}  \cdot \frac{\partial \Gamma_n \left ( \mathbf{P} \right )}{\partial z}  \\
				-\left ( \frac{\partial \Gamma_n \left ( \mathbf{P} \right )}{\partial x}  \right )^2 -\left ( \frac{\partial \Gamma_n \left ( \mathbf{P} \right )}{\partial y}  \right )^2  \\
			\end{bmatrix}^T.
		\end{gathered}
	\end{equation}
	The plane $S$ formed by $h_{n,1}  \left ( \mathbf{P} \right )$, $h_{n,2}  \left ( \mathbf{P} \right )$ is perpendicular to $t_n$. Hence, we can write any unit vector on the plane $S$ as, 
	\begin{equation} \label{eq:6A}
		\begin{bmatrix}
			\cos  \vartheta_n & \sin \vartheta_n & 0  \\
		\end{bmatrix},
	\end{equation}
	where $\vartheta_n$ is the direction coefficient affecting the tangential disturbance. The disturbance matrix while considering a single obstacle is given as:
	\begin{equation} \label{eq:7A}
		\begin{gathered}
			\mathbf{M}_n \left ( \mathbf{P} \right ) = \mathbf{M}_{n1} \left ( \mathbf{P} \right )
			+ \mathbf{M}_{n2} \left ( \mathbf{P} \right ),  \\
			\mathbf{M}_n = \mathbf{I} - \frac{\mathbf{t}_n \left ( \mathbf{P} \right ) \mathbf{t}_n \left ( \mathbf{P} \right )^T}{\left| \Gamma_n \left ( \mathbf{P} \right ) \right|^{\frac{1}{\varrho_n}}\mathbf{t}_n \left ( \mathbf{P} \right )^T \mathbf{t}_n \left ( \mathbf{P} \right )} + \\ \frac{h_n \left ( \mathbf{P} \right ) \mathbf{t}_n \left ( \mathbf{P} \right )^T}{\left| \Gamma_n \left ( \mathbf{P} \right ) \right|^{\frac{1}{\varsigma_n}}h_n \left ( \mathbf{P} \right )^T \mathbf{t}_n \left ( \mathbf{P} \right )}.
		\end{gathered}
	\end{equation}
	Here we can define $\varrho_n$ and $\varsigma_n$ using the given formulas,
	\begin{equation} \label{eq:8A}
		\left\{\begin{matrix}
			\varrho_n = \varrho_0 \cdot \exp \left ( 1 - \frac{1}{\bar{d}\left ( P, P_d \right ) \bar{d}\left ( P, O_n \right )}  \right ),\\
			\varsigma_n = \varsigma_0 \cdot \exp \left ( 1 - \frac{1}{\bar{d}\left ( P, P_d \right ) \bar{d}\left ( P, O_n \right )}  \right ).
		\end{matrix}\right.
	\end{equation}
	Here, $d\left ( \mathbf{P}, O_n \right )$ represents the distance between the end of the path and the surface of the $N$-th obstacle, $\varrho_0$,$\varsigma_0$ represents the response coefficient of the obstacle. The dynamic obstacle speed threat is represented as
	\begin{equation} \label{eq:9A}
		v_{\textup{obs}} \left ( \mathbf{P} \right ) = \sum_{n=1}^{N} \mathfrak{w}_n \left ( \mathbf{P} \right ) \exp \left [ \frac{-\Gamma_n \left ( \mathbf{P} \right )}{\Upsilon} \right ] V_{\textup{obs}}^{n} \left ( \mathbf{P} \right ).
	\end{equation}
	where $V_n^{\textup{obs}}$ is considered as the moving speed of the $N$th obstacle, $\Upsilon$ is a constant greater than zero. 
	\section{Stochastic Gradient Langevin Dynamics for Action Robust RL} \label{SGLD_soln}	
	However, solving the mixed Nash equilibrium objective in (\ref{eq:11}) can result in the solution becoming trapped at a local saddle point, preventing it from reaching the global optimum. To address this, both the agent and adversarial policies are updated by incorporating isotropic gradient noise into the stochastic gradient descent (SGD) process, to minimize the loss functions during the training of the agent and adversarial networks. By adaptively scaling this noise with higher-order curvature measures, such as Fisher scoring, the noise can be preconditioned to enhance convergence properties \cite{bhardwaj2019adaptively}. When optimizing the loss function by tuning the parameter $\Theta$ using SGD, the update equation at each step is given by
	\begin{equation} \label{eq:10A}
		\hat{g_s}\left ( \Theta_n \right ) \leftarrow \nabla_{\Theta} \hat{\mathcal{L}}_s \left ( \Theta_n \right ),
	\end{equation}
	where,  $\hat{\mathcal{L}}_s \left ( \Theta_n \right )$ represents the stochastic estimate of the loss function. Hence, the update equation can be represented as
	\begin{equation} \label{eq:11A}
		\Theta_{n+1} \leftarrow \Theta_n - \eta \left ( \hat{g_s} \left ( \Theta_n \right ) \right ), 
	\end{equation}
	where $\eta>0$ is the learning rate. Incorporating a zero mean Gaussian noise $\mathcal{N} \left ( 0, \mathfrak{e} \right )$ with variance $\mathfrak{e}$, we obtain,
	\begin{equation} \label{eq:12A}
		\begin{gathered}
			\zeta_n \sim \mathcal{N} \left ( 0, \mathfrak{e}  \right ), \\
			\Theta_{n+1} \leftarrow \Theta_n - \eta \left ( \hat{g}_s \left ( \Theta_n \right )  + \zeta_n \right ).
		\end{gathered}
	\end{equation}
	Uniform noise scaling can result in improper parameter update scaling, potentially slowing the training process and consequently converging to suboptimal minima \cite{luo2018adaptive}. To address this issue, adaptive preconditioners such as RMSProp \cite{zou2019sufficient} are used, which approximates the inverse of the second-order moments of the gradient update through a diagonal matrix. Bhardwaj et al. Bhardwaj et al. \cite{bhardwaj2019adaptively} proposed an adaptive preconditioned noise approach that leverages a diagonal approximation of the second-order gradient moments, improving the training efficiency of first-order methods. Consequently, noise scaling is adjusted proportionally to accelerate the training as follows
	\begin{equation} \label{eq:13A}
		\begin{aligned} 
			\bm{C}_n \leftarrow&  \kappa_C \bm{C}_{n-1} + \left ( 1 - \kappa_C \right ) \left ( \hat{\bm{g}_s} \left ( \Theta_n \right ) - \bm{\mu}_n \right )\left ( \hat{\bm{g}_s} \left ( \Theta_n \right ) - \bm{\mu}_{n-1} \right ), \\ 
			&\zeta_n \sim \mathcal{N} \left ( \bm{\mu}_n, \bm{C}_n \right ), \\
			&\Theta_{n+1} \leftarrow \Theta_n - \eta \left ( \hat{\bm{g}}_s \left ( \Theta_n \right ) + \psi \zeta_n \right ).
		\end{aligned}
	\end{equation}
	In \eqref{eq:13A}, the noise covariance preconditioner scales the noise in proportion to the dimensions exhibiting larger gradients. This method facilitates escape from saddle points and enhances exploration of the policy space, especially when encountering broader loss minima, thereby accelerating convergence and yielding improved solutions. Consequently, the integration of adaptive noise within the loss function enables the simultaneous optimization of both the critic loss and the loss functions associated with the agent and adversarial policies. If $\alpha_i$ denotes the adversarial contribution, the general action $a_t$ is expressed as follows
	\begin{equation} \label{eq:14A}
		a^t = \left ( 1 - \alpha \right ) \cdot  \pi^{\textup{agent}}_{\theta}\left ( \Phi^t \right ) + \alpha \cdot \pi^{\textup{adv}}_{\omega} \left ( \Phi^t \right ).
	\end{equation}
	If we consider a batch of samples $B = \left\{ \left ( \Phi^t, a^t, r^t, {\Phi}^{t+1}, \mathfrak{d} \right ) \right\}$ from the replay buffer $\mathcal{D}$, then we compute the critic target $y_{\textup{targ}}$  as follows
	\begin{equation} \label{eq:15A}
		y_{\textup{targ}} =  r^t  + \gamma \left ( 1 - \mathfrak{d} \right ) Q_{\phi_{\textup{targ}}} \left ( \Phi^{t+1}, \left ( 1 - \alpha \right ) \cdot \pi^{\textup{agent}}_{\theta} + \alpha  \cdot \pi^{\textup{adv}}_{\omega} \right ).
	\end{equation}
	Here, $\phi_{\textup{targ}}$ represents the parameters of the target critic network and $\mathfrak{d}$ represents the terminal status of the state transitions. Therefore, the target $y_{\textup{targ}}$ can be utilized to formulate the expected critic loss function as, 
	\begin{equation}\label{eq:16A}
		L\left ( \phi \right ) = \frac{1}{N} \sum_{B = \left\{ \left ( \Phi^t, a^t, r^t, {\Phi}^{t+1}, \mathfrak{d} \right ) \right\}} \left ( y\left ( r^t, \Phi^{t+1}, \mathfrak{d} \right ) - Q_{\phi} \left ( \Phi^t, a^t \right )  \right )^2. 
	\end{equation}
	We can also compute the expectations of the loss functions of the  agent and adversary policy networks over the experience samples $B$ as,
	\begin{equation}\label{eq:17A}
		\begin{gathered}
			\nabla_{\theta} \widehat{J \left ( \theta, \omega_t \right )} = \frac{1 - \alpha}{N}  \sum_{\Phi,a \in B} \nabla_{\theta} \pi^{\textup{agent}}_{\theta} \left ( \Phi \right )  \nabla_a Q_{\phi} \left ( \Phi, a \right ) ,   \\
			\nabla_{\omega} \widehat{J \left ( \theta_t, \omega \right )} = \frac{\alpha}{N} \sum_{\Phi, a \in B} \nabla_{\omega} \pi^{\textup{adv}}_\omega \left ( \Phi \right ) \nabla_a Q_{\phi} \left ( \Phi, a \right ). 
		\end{gathered}
	\end{equation}
	To optimize the loss functions of the critic and policy, as specified in equations (\ref{eq:16A}) and (\ref{eq:17A}) respectively, we incorporate the noise into the update equations as specified in (\ref{eq:13A}). This approach is taken to update the parameters of the joint critic network $\phi$ and the agent and adversary policy network $\left\{ \theta, \omega \right\}$. Let $\bm{g} = \left \{ \nabla_{\theta} J, \nabla_{\omega} J   \right \} $, $\bm{\mu}_t = \left \{ \mu_{\theta}, \mu_{\omega}   \right \} $, $\bm{C}_t = \left \{ C_{\theta}, C_{\omega}   \right \}$ represent the gradient, the joint mean value of the parameters, and the covariance matrices, respectively. According to the dynamics of SGLD given in (\ref{eq:13}), the update equation for the agent and the adversary parameter to obtain the mixed Nash equilibrium for the $k^{\textup{th}}$ sample from the buffer at time $t$, is given as:
	\begin{equation}\label{eq:18A}
		\theta^{k+1}_{t} \leftarrow \theta^{k}_{t} + \eta \left ( \nabla_{\theta} \widehat{J\left ( \theta, \omega_t \right )} + \psi \xi_t  \right ),
	\end{equation}
	\begin{equation}\label{eq:19A}
		\omega^{k+1}_{t} \leftarrow \omega^{k}_{t} - \eta \left ( \nabla_{\omega} \widehat{J\left ( \theta, \omega_t \right )} \left (\Gamma_{t}  \right ) + \psi \xi_t  \right ).
	\end{equation}
	Here, $\psi$ represents the noise parameter. The parameters updated with the time $t$ is given by
	\begin{equation}\label{eq:20A}
		\bar{\omega}_t = \left ( 1 - \beta  \right ) \bar{\omega}_t + \beta \omega^{\left (k+1  \right )}_t ;
		\; \bar{\theta}_t = \left ( 1 - \beta  \right ) \bar{\theta}_t + \beta \theta^{\left (k+1  \right )}_t.
	\end{equation}
	Obtain the parameters for the next time duration as,
	\begin{equation}\label{eq:21A}
		\omega_{t+1} = \left ( 1 - \beta  \right ) \omega_t + \beta \bar{\omega}_t ;  \; \theta_{t+1} = \left ( 1 - \beta  \right ) \theta_t + \beta \bar{\theta}_t.
	\end{equation}
	\section{Algorithms} \label{adv_state}
	\subsection{Algorithm to Generate Adversarial States}
	\begin{algorithm}[H]
		\caption{Generate adversarial state $\Phi^{\textup{adv}}$ for the curriculum adaptation of action robust RL using projected gradient ascent}
		\label{alg:adv_state}  
		\textbf{Input} Current State $\Phi$,  maximum perturbation $\epsilon$, Step size $\hat{\alpha}= \epsilon / N$, Mean value of state $\mu_{\Phi}$, Standard deviation of the state $\sigma_{\Phi}$ . \\
		\textbf{Output} Adversarial State $\Phi^{\textup{adv}}$
		\begin{algorithmic}[1]  
			\State Initialize: $\delta_0 \sim \text{Uniform}([-\epsilon, \epsilon]^d)$.
			\State Set: $\Phi_0 = \Phi + \delta_0$.
			\For{$n = 0$ to $N_{\text{step}} - 1$}
			\State  Compute gradient: $g_n \leftarrow \nabla_{\Phi_n} \mathcal{L}_{\pi}(\Phi_n)$.
			\State Gradient ascent update: $\Phi_{n+1}' \leftarrow \Phi_n + \hat{\alpha} \cdot \operatorname{sgn}(g_n)$.
			\State Project to \( \ell_\infty \)-ball: $\Phi_{n+1} \leftarrow \Pi_{\epsilon}(\Phi_{n+1}' - \Phi) + \Phi$.
			\State $\Phi_n \leftarrow \Phi_{n+1}'. $
			\EndFor
			\State \Return $\Phi^{\textup{adv}} = \Phi_{N_{\text{step}}}$.
		\end{algorithmic} 
	\end{algorithm}
	\subsection{Algorithm to Compute TD Error}
	\begin{algorithm}[H]
		\caption{Temporal difference (TD) error of action robust RL Policy}
		\label{alg:cata_forget}  
		\textbf{Input} Adversarial state $\Phi^{\textup{adv}}_{\epsilon}$ with strength $\epsilon$ , Sampled Batch $B =  \left \{ \left (\Phi_{\textup{adv}}^{\epsilon}, a^{\epsilon}_{\textup{adv}}, r^{\epsilon}_{\textup{adv}}, {\Phi}', \mathfrak{d}  \right ) \right \}$,  Robust agent policy $\pi^{\textup{tar}}_{\textup{agent}}$, robust adversarial policy $\pi^{\textup{tar}}_{\textup{adv}}$, Critic for the robust RL $Q_{\textup{AR}}$, Critic target for the robust RL $Q^{\textup{tar}}_{\textup{AR}}$,   \\
		\textbf{Output} TD error for the robust policy $\mathbf{TD}^{\textup{AR}}_{\Phi_{\textup{adv}}}$ \\
		\begin{algorithmic}[1] 
			\State Initialize: $\Phi_{\textup{norm}}$.
			\For{ $j = 1$ to $N_{\textup{samples}}$}
			\State Obtain the value with adversarial state as  $V_{\textup{adv}} = Q_{\textup{AR}} \left ( \Phi_{\textup{adv}}^{\epsilon}, a^{\epsilon}_{\textup{adv}}  \right )$.
			\State Obtain the next action using the target policies $a^{\textup{next}} = \left ( 1 - \alpha \right ) \cdot \pi^{\textup{tar}}_{\textup{agent}} \left ( \Phi^{t+1} \right ) + \alpha \cdot \pi^{\textup{tar}}_{\textup{adv}} \left ( \Phi^{t+1} \right )$.
			\State Obtain the value based on next observation and action $V^{\textup{tar}} = Q^{\textup{tar}}_{\textup{AR}} \left ( \Phi^{t+1}, a^{\textup{next}} \right )$.
			\State Obtain the target value $V^{\textup{targ}} = r^{\epsilon}_{\textup{adv}} + (1- \mathfrak{d}) \cdot \gamma \cdot V^{\textup{next}}$.
			\State Obtain the TD error $\mathbf{TD}_{\Phi^{\epsilon}_{\textup{adv}}} = \left ( V_{\textup{tar}} - V_{\textup{adv}}  \right ) \cdot \left ( V_{\textup{tar}} - V_{\textup{adv}}  \right )/ \sigma^2.$
			\EndFor
		\end{algorithmic} 
	\end{algorithm}

	\section{Proofs for Curriculum Adaptation} \label{proofs}
	\subsection{Proof of Lemma \ref{lemma_1}} \label{lemma_1_proof}
	As per Kantorovic-Rubinstein duality, for two different transition probability distributions $P_k$ and $P_{k+1}$, we define the Wasserstein distance while considering the adversarial state $\Phi'_{\epsilon_{k+1}} \sim P_{k+1}$ and $\Phi'_{\epsilon_k} \sim P_{k}$  between them as,
	\begin{equation}\label{eq:23A}
		\mathcal{W}_1 \left (P_{k+1}, P_k \right ) = \sup_{\text{Lip} \left (f \right ) \leq 1} \left | \mathbb{E}_{} \left [f \left (\Phi'_{\epsilon_{k+1}} \right ) \right ] - 
		\mathbb{E}_{} \left [f \left (\Phi'_{\epsilon_{k}} \right ) \right ] \right |.
	\end{equation}
	As per Assumption on regularity condition in (\ref{eq:17}), the value function $Q$ is Lipschitz. Thus, if we consider $f\left(s \right) = \frac{1}{L_Q} Q \left (\cdot \right )$ , then we can write the following relation,
	\begin{equation}\label{eq:24A}
		\begin{aligned}
			&\left | \mathbb{E}_{\Phi'_{\epsilon_{k+1}}\sim P_{k+1}} \left [Q^{\pi^*}_{k+1} \left (\Phi'_{\epsilon_{k+1}} \right ) \right] - \mathbb{E}_{\Phi'_{\epsilon_{k}}\sim P_{k}} \left [Q^{\pi^*}_{k} \left (\Phi'_{\epsilon_{k}} \right ) \right] \right | \leq \\ 
			& \sup \left | \mathbb{E}_{\Phi'_{\epsilon_{k+1}}\sim P_{k+1}} \left [Q^{\pi^*}_{k+1} \left (\Phi'_{\epsilon_{k+1}} \right ) \right] - \mathbb{E}_{\Phi'_{\epsilon_{k}}\sim P_{k}} \left [Q^{\pi^*}_{k} \left (\Phi'_{\epsilon_{k}} \right ) \right] \right |, \\
			&\left | \mathbb{E}_{\Phi'_{\epsilon_{k+1}}\sim P_{k+1}} \left [Q^{\pi^*}_{k+1} \left (\Phi'_{\epsilon_{k+1}} \right ) \right] - \mathbb{E}_{\Phi'_{\epsilon_{k}}\sim P_{k}} \left [Q^{\pi^*}_{k} \left (\Phi'_{\epsilon_{k}} \right ) \right] \right |  \\  
			&\leq L_Q \mathcal{W}_1 \left (P_{k+1}, P_k \right ).
		\end{aligned}
	\end{equation}
	
	\subsection{Proof of Lemma \ref{lemma_2}} \label{lemma_2_proof}
	Using the Bellman optimality equation,
	\begin{equation}\label{eq:25A}
		Q^{\pi^*} \left(\Phi \right ) = \max_a \left( r + \gamma \mathbb{E}_{\Phi'} \left [ Q^{\pi^*} \left ( \Phi'\right ) \right ] \right)
	\end{equation}
	Let us consider the following inequality for two real vectors $a$ and $b$, which can be written as,
	\begin{equation}\label{eq:26A}
		\begin{gathered}
			\left | a \right | - \left | b \right | \leq \left | a -b \right | \\
			\max a - \max b \leq \max \left (a-b \right )
		\end{gathered}
	\end{equation}
	Hence, using the above inequalities, let us consider two adversarially perturbed states, $\Phi_{\epsilon_k}$ and $\Phi_{\epsilon_{k+1}}$, taking the difference between the value functions,
	\begin{equation}\label{eq:27A}
		\begin{aligned}
			&\left | Q^{\pi^*} \left ( \Phi_{\epsilon_k} \right ) - Q^{\pi^*} \left ( \Phi_{\epsilon_{k+1}} \right )\right| \leq \\
			& \max \left | r_{\epsilon_k} - r_{\epsilon_{k+1}} \right | +  \\ 
			& \gamma \max \left (\mathbb{E} \left [ Q^{\pi^*} \left ( \Phi'_{\epsilon_k}  \right ) \right]  - \mathbb{E} \left [ Q^{\pi^*} \left ( \Phi'_{\epsilon_{k+1}}  \right ) \right]\right )
		\end{aligned}
	\end{equation}
	Considering the Lipschitz property of the reward and the critic function for incremental adversarial state, we can write,
	\begin{equation}\label{eq:28A}
		\left |Q^{\pi^*}_{k+1} \left (\Phi_{\epsilon_{k+1}} \right ) - Q^{\pi^*}_{k} \left (\Phi_{\epsilon_{k}} \right )  \right | \leq L'_Q d \left (\Phi_{\epsilon_{k+1}}, \Phi_{\epsilon_{k}}\right ).
	\end{equation}
	where, $L'_Q = L_r + \gamma L_Q$.
	
	\subsection{Proof of Lemma \ref{lemma_3}}\label{lemma_3_proof}
	As per the definition of the TD error it is a functional transformation of the transition distribution. Hence if we define the functional as $T \left (\cdot \right )$, then we can write,
	\begin{equation}\label{eq:29A}
		\mathbf{TD}^{\pi^*}_k = T \left (P_k \right ),  \mathbf{TD}^{\pi^*}_{k+1} = T \left (P_{k+1} \right ).
	\end{equation}
	If we consider a general property of the optimal transport theory, then we can write if $T$ is a Lipschitz function from a metric space $\mathcal{X} \rightarrow \mathbb{R} $, and $\mu, \nu \in \mathcal{P} \left( \mathcal{X} \right ) $ are probability measures on that space, then,
	\begin{equation}\label{eq:30A}
		\mathcal{W}_1 \left (\mu, \nu \right ) \leq L_T \mathcal{W}_1 \left (T_{\kappa} \mu, T_{\kappa} \nu  \right ),
	\end{equation}
	
	where, $T_{\kappa} \mu$ is the pushforward of $\mu$ under $T$ and $L_T$ is the Lipschitz constant of T. Hence, the above relation becomes,
	\begin{equation}\label{eq:31A}
		\mathcal{W}_1 \left (\mu, \nu \right ) \leq L_T \mathcal{W}_1 \left (T_{\kappa} P_{k+1}, T_{\kappa} \nu  P_{k} \right )
	\end{equation}
	As we know, $T \left (P \right ) = \mathbf{TD}^{\pi^*} \left ( P \right ) $, hence we write,
	\begin{equation}\label{eq:32A}
		\mathcal{W}_1(P_{k+1}, P_k) \leq L_T \mathcal{W}_1(\mathbf{TD}^{\pi^*}_{k+1}(\epsilon_{k+1}), \mathbf{TD}^{\pi^*}_k(\epsilon_k)).
	\end{equation}
	
	\subsection{Proof of Lemma \ref{lemma_4}}\label{lemma_4_proof}
	Considering the dual form of Wasserstein-1 distance with $\Phi'_{\epsilon_k} \sim P_k$ and $\Phi'_{\epsilon_{k+1}} \sim P_{k+1}$,
	\begin{equation}\label{eq:33A}
		\mathcal{W}_1 \left (P_k, P_{k+1} \right )  = \sup_{f \in \text{Lip}_1} \left | \mathbb{E}_{} \left [ f \left (\Phi'_{\epsilon_k} \right ) \right] - \mathbb{E}_{} \left [ f \left (\Phi'_{\epsilon_{k+1}} \right ) \right]  \right |.
	\end{equation}
	If we define $f\left ( \cdot \right )$ be the distance between the perturbed state and the transition state as $f\left (\Phi' \right ) = d \left (\Phi', \Phi_{\epsilon_k} \right ) $. The function is 1-Lipschitz, hence for any state $\Phi'_1, \Phi'_2 \in \mathcal{S}$, we can write,
	\begin{equation}\label{eq:34A}
		\begin{gathered}
			\left | f \left( \Phi'_1 \right ) - f \left( \Phi'_2 \right )  \right | \\
			\left | d \left( \Phi'_1, \Phi_{\epsilon_k} \right ) - d \left( \Phi'_2, \Phi_{\epsilon_k} \right )   \right | \leq d \left (\Phi'_1, \Phi'_2 \right ).
		\end{gathered}
	\end{equation}
	Now if we plug the above relation in the dual form, we can write,
	\begin{equation}\label{eq:35A}
		\left | \mathbb{E} \left[ f \left (\Phi'_{k+1} \right )   \right] - \mathbb{E} \left[ f \left (\Phi'_{k} \right )   \right]  \right | \leq \mathcal{W}_1 \left (P_k, P_{k+1} \right ).
	\end{equation}
	We can the above terms, $\mathbb{E} \left[ f \left (\Phi'_{k} \right ) \right] \approx 0 $ as we are considering a smooth dynamics for the UAV, as it does not allow a drastic change in the state. And $\mathbb{E} \left[ f \left (\Phi'_{k+1} \right ) \right] \approx d \left (\Phi_{\epsilon_{k+1}}, \Phi_{\epsilon_{k}} \right ) $, as the distribution is centered around $\Phi_{\epsilon_{k+1}}$. Hence, using these approximations we can write,
	\begin{equation}\label{eq:36A}
		d \left (\Phi_{\epsilon_k}, \Phi_{\epsilon_{k+1}} \right ) \leq \mathcal{W}_1 \left (P_k, P_{k+1} \right ).
	\end{equation}
	\subsection{Proof of Theorem \ref{theorem_main}}\label{main_proof}
	If we consider for an adversarial perturbed state under the attack strength $\epsilon_1$ as $\Phi^{\epsilon_1}_{\text{adv}}$, then for adversarial strength $\epsilon_2 > \epsilon_1$ , we take the difference between the expectation of the TD error as,
	\begin{equation} \label{eq:37A}
		\begin{gathered}
			\mathbb{E} \left [ \mathbf{TD}^{\pi^*}_{k+1} \left ( \epsilon_{k+1} \right ) - \mathbf{TD}^{\pi^*}_{k+1} \left ( \epsilon_{k} \right ) \right ] \\ = r^{\epsilon_{k+1}} - r^{\epsilon_{k}} + \gamma \mathbb{E}_{\Phi'_{\epsilon_{k+1}} \sim P_{k+1}} \left[ Q^{\pi^*}_{k+1} \left (\Phi'_{\epsilon_{k+1}} \right )  \right] \\ - Q^{\pi^*}_{k+1} \left ( \Phi_{\epsilon_{k+1}} \right ) 
			- \left ( \gamma \mathbb{E}_{\Phi'_{\epsilon_{k}} \sim P_{k+1}} \left[ Q^{\pi^*}_{k} \left (\Phi'_{\epsilon_{k}} \right )  \right] - Q^{\pi^*}_{k} \left ( \Phi_{\epsilon_{k}} \right ) \right ) \\
			= \left ( r^{\epsilon_{k+1}} - r^{\epsilon_{k}}\right ) + \gamma  \mathbb{E}_{\Phi'_{\epsilon_{k+1}} \sim P_{k+1}} \left[ Q^{\pi^*}_{k+1} \left (\Phi'_{\epsilon_{k+1}} \right )  \right] \\ - \gamma \mathbb{E}_{\Phi'_{\epsilon_{k}} \sim P_{k}} \left[ Q^{\pi^*}_{k} \left (\Phi'_{\epsilon_{k}} \right )  \right] 
			- \left ( Q^{\pi^*}_{k+1} \left ( \Phi_{\epsilon_{k+1}} \right ) - Q^{\pi^*}_{k} \left ( \Phi_{\epsilon_{k}} \right ) \right ).
		\end{gathered}
	\end{equation}
	If we consider the Assumption \ref{assump_1} and \ref{assump_2} and Lemma \ref{lemma_1} and \ref{lemma_2} and apply the bounds, to the relation derived  in (\ref{eq:37A}), we obtain,
	\begin{equation}\label{eq:38A}
		\begin{aligned}
			&\mathbb{E} \left [ \mathbf{TD}^{\pi^*}_{k+1} \left ( \epsilon_{k+1} \right ) - \mathbf{TD}^{\pi^*}_{k+1} \left ( \epsilon_{k} \right ) \right ] \leq  \\ &\gamma L_v \mathcal{W}_1(P_{k+1}, P_k) + L_v d(\Phi_{\epsilon_{k+1}}, \Phi_{\epsilon_k}).
		\end{aligned}
	\end{equation}
	If we use the relation in lemma \ref{lemma_4}, 
	\begin{equation} \label{eq:39A}
		d(\Phi_{\epsilon_{k+1}}, \Phi_{\epsilon_k}) \leq \mathcal{W}_1(P_{k+1}, P_k),
	\end{equation}
	We obtain the following relation,
	\begin{equation}\label{eq:40A}
		\begin{aligned}
			&\mathbb{E}  \left [ \left| \mathbf{TD}^{\pi^*}_{k+1} \left ( \epsilon_{k+1} \right ) - \mathbf{TD}^{\pi^*}_{k+1} \left ( \epsilon_{k} \right ) \right| \right ] \leq  \\ &L_V (1 + \gamma) \mathcal{W}_1 (P_{k+1}, P_k).
		\end{aligned}
	\end{equation}
	Now, we per the lemma \ref{lemma_3}, the distributional distance between the state distributions $P_k$ and $P_{k+1}$, which differ in Wasserstein distance based on $\epsilon_k$ and $\epsilon_{k+1}$, as
	\begin{equation}\label{eq:41A}
		\mathcal{W}_1(P_{k+1}, P_k) \leq \mathcal{W}_1(\mathbf{TD}^{\pi^*}_{k+1}(\epsilon_{k+1}), \mathbf{TD}^{\pi^*}_k(\epsilon_k)).
	\end{equation}
	Hence, using the relation of the distance between the state distribution in lemma \ref{lemma_3}, the final relation becomes,
	\begin{equation}\label{eq:42A}
		\begin{aligned}
			&\mathbb{E}  \left [ \left| \mathbf{TD}^{\pi^*}_{k+1} \left ( \epsilon_{k+1} \right ) - \mathbf{TD}^{\pi^*}_{k+1} \left ( \epsilon_{k} \right ) \right| \right ] \leq \\
			&  m \mathcal{W}_1 \left( \mathbf{TD}_{k+1}^{\pi^*} (\epsilon_{k+1}), \mathbf{TD}_{k}^{\pi^*} (\epsilon_k) \right),
		\end{aligned}
	\end{equation}
	where, $m=L_V L_T \left (1 + \gamma \right )$.
	\subsection{Proof of Theorem \ref{theorem:gen}} \label{proof:gen}
	If we decompose the LHS of the expression in Theorem \ref{theorem:gen}, we can write,
	\begin{equation}\label{eq:43A}
		\begin{gathered}
			\mathbb{E} \left[ R \left (\pi^*_K \left (\mathfrak{A} \right ) \right ) \right ] -   \mathbb{E} \left[ R \left (\pi^*_K \left (K \right ) \right ) \right ]  + \mathbb{E} \left[ R \left (\pi^*_K \left (K \right ) \right ) \right ] - \\ \frac{1}{n} \sum_{i=1}^n R \left ( \pi^*_K \right ).
		\end{gathered}
	\end{equation}
	The first two terms refer to the \textbf{distribution shift in TD-space}, while the last two terms refer to \textbf{intrinsic generalization}. Now we will deal with these terms separately to prove the relation in (\ref{eq:42A}). We consider $\mathcal{R} \left (\pi; P \right )$ as the expected returnpolicy $\pi$ under the transition kernel $P$. At stage $K$,  let us consider the transition $P_K$ for empirical evaluation. Hence, we can define the intrinsic generalization gap for policy $\pi^*_{K}$ is
	\begin{equation}\label{eq:44A}
		\textup{Gap}_{\text{int}} \left ( \pi^*_K; P_K \right ) =  \left | \mathbb{E}_{\tau \sim P_K, \pi^*_K} \left [R \left (\tau \right ) \right ] -\frac{1}{n} \sum^n_{i=1} R \left (\tau_i \right ) \right |,
	\end{equation}
	where $\tau_i$ are the $n$ iid trajectories generated by executing $\pi^*_K$ in $P_K$. From the intrinsic generalization result in Lemma 2 in \cite{wang2019generalization}, for a reparameterizable policy class $\Pi$ and any fixed environment $P$ with probability $1-\delta$, we can write, 
	\begin{equation}\label{eq:45A}
		\begin{gathered}
			\sup_{\pi \in \Pi} 
			\left| \mathbb{E}_{\tau \sim P_{\pi}}[R(\tau)] 
			- \frac{1}{n} \sum_{i=1}^{n} R(\tau_i) \right|
			\leq 2\, \mathrm{Rad}_n(\mathcal{F}_{P}) \\
			+ c \sqrt{\frac{\log(2/\delta)}{2n}},
		\end{gathered}
	\end{equation}
	where, $\mathrm{Rad}_n$ is the empirical Rademacher complexity of the return function class $\mathcal{F}_P := \{ \tau \mapsto R \left (\tau \right ): \pi \in \Pi \}$ under environment $P$, $c$ is the upper bound on the per-trajectory return. Since $\pi^*_K \in \Pi$ the same bound applies for $P = P_K$ and fixed $\pi^*_K$.  We define $\text{Gap}_{\text{int}}(\pi^{\star}_{K}; P_{K})$ as, 
	\begin{equation}\label{eq:46A}
		\text{Gap}_{\text{int}}(\pi^{\star}_{K}; P_{K}) 
		\leq \mathrm{Rad}_{K} + O\!\left( \frac{c}{\sqrt{n}} \sqrt{\log(1/\delta)} \right),
	\end{equation}
	where we denote $\textup{Rad}_K = 2 \text{Rad}_K \left ( \mathcal{F}_{P_K} \right ).$ As the return function $R \left (\tau \right )$ is $L_r$ -Lipschitz in the state-action sequence and the policy $\pi$ is $L_{\pi}$ Lipschitz in its parameters as per Lemma 3 and 4 in \cite{wang2019generalization}, then $\textup{Rad}_n \left (\mathcal{F}_{P_K} \right )$ is bounded as per Rademacher complexity of the underlying policy parameterization class $\Theta$, which is shown as, 
	\begin{equation}\label{eq:47A}
		\textup{Rad}_n \left ( \mathcal{F}_{P_K} \right ) \leq L_r L_\pi \textup{Rad}_{n} \left ( \Theta \right ).
	\end{equation}
	It scales linearly with the given Lipschitz constraints. Hence, combining the above relations, we finally obtain,
	\begin{equation}\label{eq:48A}
		\boxed{
			\text{Gap}_{\text{int}}(\pi^{\star}_{K}; P_{K}) 
			\leq \mathrm{Rad}_{K} + O\!\left( \frac{c}{\sqrt{n}} \sqrt{\log(1/\delta)} \right)
		}
	\end{equation}
	Now we will focus on the case of distribution shift in the TD-space.  Here, we wish to bound the performance difference of $\pi^*_K$ when executed under two different transition laws:
	\begin{itemize}
		\item $P_{\mathfrak{A}}$: the transition kernel under an unseen adversarial attack $\mathfrak{A}$,
		\item $P_{K}$: the transition kernel corresponding to the last curriculum stage with attack intensity $\epsilon_K$.
	\end{itemize}
	The ultimate aim is to bound $\Delta_{\mathsf{A},K}$ which is expressed as,
	\begin{equation}\label{eq:49A}
		\Delta_{\mathsf{A},K} \ :=\ 
		\bigg|\, \mathbb{E}_{\tau \sim P_{\mathsf{A}},\,\pi^*_K}\!\left[R(\tau)\right] 
		- \mathbb{E}_{\tau \sim P_{K},\,\pi^*_K}\!\left[R(\tau)\right] \bigg|.
	\end{equation}
	Let $V^{\pi^*_K}_{P}\left  (\Phi \right  )$ be the value function of policy $\pi^*_K$ under transition law $P$.  
	By the Bellman equation:
	\begin{equation}\label{eq:50A}
		V^{\pi^*_K}_{P}(\Phi) 
		\ =\ r(\Phi,\pi^*_K(\Phi)) 
		+ \gamma\,\mathbb{E}_{\Phi' \sim P(\cdot \mid \Phi,\pi^*_K(\Phi))}\!\left[ V^{\pi^*_K}_{P}(\Phi') \right].
	\end{equation}
	For any two transition laws $P$ and $P'$, subtracting their Bellman equations and applying Lemma~\ref{lemma_1} to the $Q$-term yields:
	\begin{equation}\label{eq:51A}
		\begin{gathered}	
			\big|\, V^{\pi^*_K}_{P}(\Phi) - V^{\pi^*_K}_{P'}(\Phi) \,\big|
			\ \le\ \\ L_Q\, \mathcal{W}_1\!\left(P(\cdot \mid \Phi,\pi^*_K(\Phi)),\ P'(\cdot \mid \Phi,\pi^*_K(\Phi))\right),
		\end{gathered}
	\end{equation}
	where $L_Q$ is the Lipschitz constant of $Q^{\pi^*_K}$ with respect to its state argument. Iterating this bound over a finite horizon $T$ and using the Lipschitz property of the reward function $r$ with constant $L_r$ (Assumption \ref{assump_1}), we obtain:
	\begin{equation}\label{eq:52A}
		\Delta_{\mathfrak{A},K} \ \le\ C_0 \, \mathcal{W}_1(P_{\mathfrak{A}}, P_K),
	\end{equation}
	where,
	\begin{equation}\label{eq:53A}
		C_0 \ :=\ L_r \sum_{t=0}^{T-1} \gamma^t + L_Q \sum_{t=1}^{T} \gamma^t.
	\end{equation}
	It is a problem-dependent constant aggregating the per-step Lipschitz effects of $r$ and $Q$ over the horizon. By Lemma~\ref{lemma_3}, the Wasserstein-1 distance between the two transition kernels is dominated by the Wasserstein-1 distance between their induced TD-error distributions when evaluated under the same policy:
	\begin{equation}\label{eq:54A}
		\mathcal{W}_1(P_{\mathfrak{A}}, P_K) 
		\ \le\ \mathcal{W}_1\!\left(\mathbf{TD}^{\pi^*_K}(\mathfrak{A}),\ \mathbf{TD}^{\pi^*_K}(\epsilon_K)\right),
	\end{equation}
	where $\text{TD}^{\pi^*_K}(\cdot)$ denotes the distribution of TD-errors computed from the given transition law. Let us define:
	\begin{equation}\label{eq:55A}
		\delta_{\mathfrak{A}} \ :=\ \mathcal{W}_1\!\left(\mathbf{TD}^{\pi^*_K}(\mathfrak{A}),\ \mathbf{TD}^{\pi^*_K}(\epsilon_K)\right).
	\end{equation}
	Combining the above steps we can write:
	\begin{equation}\label{eq:56A}
		\Delta_{\mathfrak{A},K}
		\ \le\ C_0\, \mathcal{W}_1(P_{\mathfrak{A}}, P_K)
		\ \le\ C_0\, \delta_{\mathfrak{A}}.
	\end{equation}
	Setting $C := C_0$ gives the claimed bound:
	\begin{equation}\label{eq:57A}
		\big|\, \mathbb{E}[R(\pi^*_K\ \text{under }\mathfrak{A})]
		- \mathbb{E}[R(\pi^*_K\ \text{under }\epsilon_K)] \,\big|
		\ \le\ C\, \delta_{\mathfrak{A}}.
	\end{equation}
	We now compare the expected return of $\pi^*_K$ at stage $K$ to that of $\pi^*_1$ at stage $1$ by inserting and subtracting the intermediate \emph{stage expectations}:
	\begin{equation}\label{eq:58A}
		\begin{gathered}
			\mathbb{E}\!\left[ R(\pi^*_K\ \text{at stage }K) \right]
			- \mathbb{E}\!\left[ R(\pi^*_1\ \text{at stage }1) \right] \\
			\quad = \sum_{k=1}^{K-1}
			\bigg(
			\mathbb{E}\!\left[ R(\pi^*_{k+1}\ \text{at stage }k+1) \right]
			- \mathbb{E}\!\left[ R(\pi^*_{k}\ \text{at stage }k) \right]
			\bigg).
		\end{gathered}
	\end{equation}
	Consider a single increment in (\ref{eq:58A}):
	\[
	\Delta_k := \mathbb{E}\!\left[ R(\pi^*_{k+1}\ \text{at stage }k+1) \right]
	- \mathbb{E}\!\left[ R(\pi^*_{k}\ \text{at stage }k) \right].
	\]
	We decompose this by holding the \emph{attack level} fixed at $\epsilon_{k+1}$:
	\begin{equation} \label{eq:59A}
		\begin{gathered}
			\Delta_k
			= \underbrace{
				\mathbb{E}\!\left[ R(\pi^*_{k+1},\ \epsilon_{k+1}) \right]
				- \mathbb{E}\!\left[ R(\pi^*_{k},\ \epsilon_{k+1}) \right]
			}_{\text{policy change at fixed attack level}}
			\\
			\quad +\ \underbrace{
				\mathbb{E}\!\left[ R(\pi^*_{k},\ \epsilon_{k+1}) \right]
				- \mathbb{E}\!\left[ R(\pi^*_{k},\ \epsilon_{k}) \right]
			}_{\text{attack level change at fixed policy}}.
		\end{gathered}
	\end{equation}
	By \textbf{Theorem~\ref{theorem_main}}, the \emph{policy change term} in (\ref{eq:59A}) is bounded in absolute value by
	\begin{equation}\label{eq:60A}
		\begin{gathered}
			\big|\, \text{policy change term} \,\big| 
			\ \le\ m\, \mathcal{W}_1\!\left(\mathrm{TD}^{\pi^*_{k+1}}(\epsilon_{k+1}),\ \mathrm{TD}^{\pi^*_{k}}(\epsilon_{k})\right) \\
			\ =\ m\,\beta_k,		
		\end{gathered}
	\end{equation}
	where $\beta_k$ is the stagewise TD misalignment.  For the \emph{attack-level change term}, Lemma~\ref{lemma_1}  bounds the return change by a constant $L_Q$ times the Wasserstein distance between the corresponding transition laws; Lemma~\ref{lemma_3} then upper-bounds that by the same $\beta_k$ via TD-error pushforward. Lemma~\ref{lemma_4} further ensures that the state-distribution perturbation is also controlled by $\beta_k$. Thus there exists a constant $C_1>0$ (depending on $L_Q$, $L_r$, $\gamma$) such that
	\begin{equation}\label{eq:61A}
		\big|\, \text{attack-level change term} \,\big| \ \le\ C_1\,\beta_k.	
	\end{equation}
	Combining both terms in (\ref{eq:60A}), we have
	\[
	|\Delta_k| \ \le\ C\,\beta_k,
	\]
	where $C := m + C_1$ absorbs all problem-dependent constants from Theorem~\ref{theorem_main}, Lemma~\ref{lemma_1}, and Lemma~\ref{lemma_4}. Applying this bound in \eqref{eq:58A} yields,
	\begin{equation}\label{eq:62}
		\bigg|\ \mathbb{E}[R(\pi^*_K\ \text{at stage }K)]
		- \mathbb{E}[R(\pi^*_1\ \text{at stage }1)]\ \bigg|
		\ \le\ C \sum_{k=1}^{K-1} \beta_k.
	\end{equation}
	\section{Further Ablation Studies}\label{abla_policies}
	\begin{figure}[!t]
		\centering
		\includegraphics[scale=0.18]{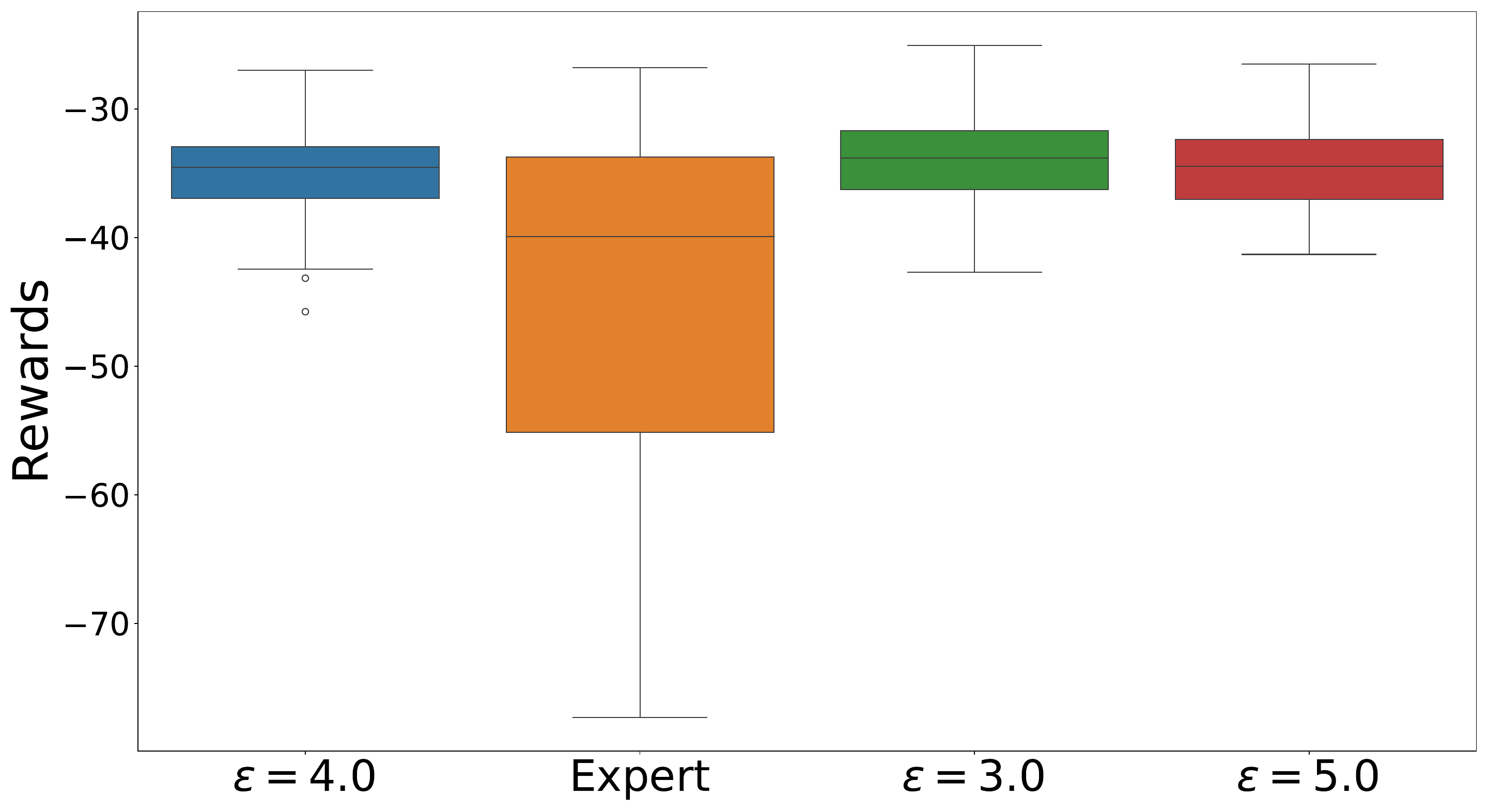}
		\caption{Comparing the test-time reward distribution with expert policy and the curriculum update stopping condition against spoofing attack and 8 dynamic obstacles across 100 episodes.}
		\label{figure_ablation_rewards}
	\end{figure}
	\begin{figure}[!t]
		\centering
		\includegraphics[scale=0.18]{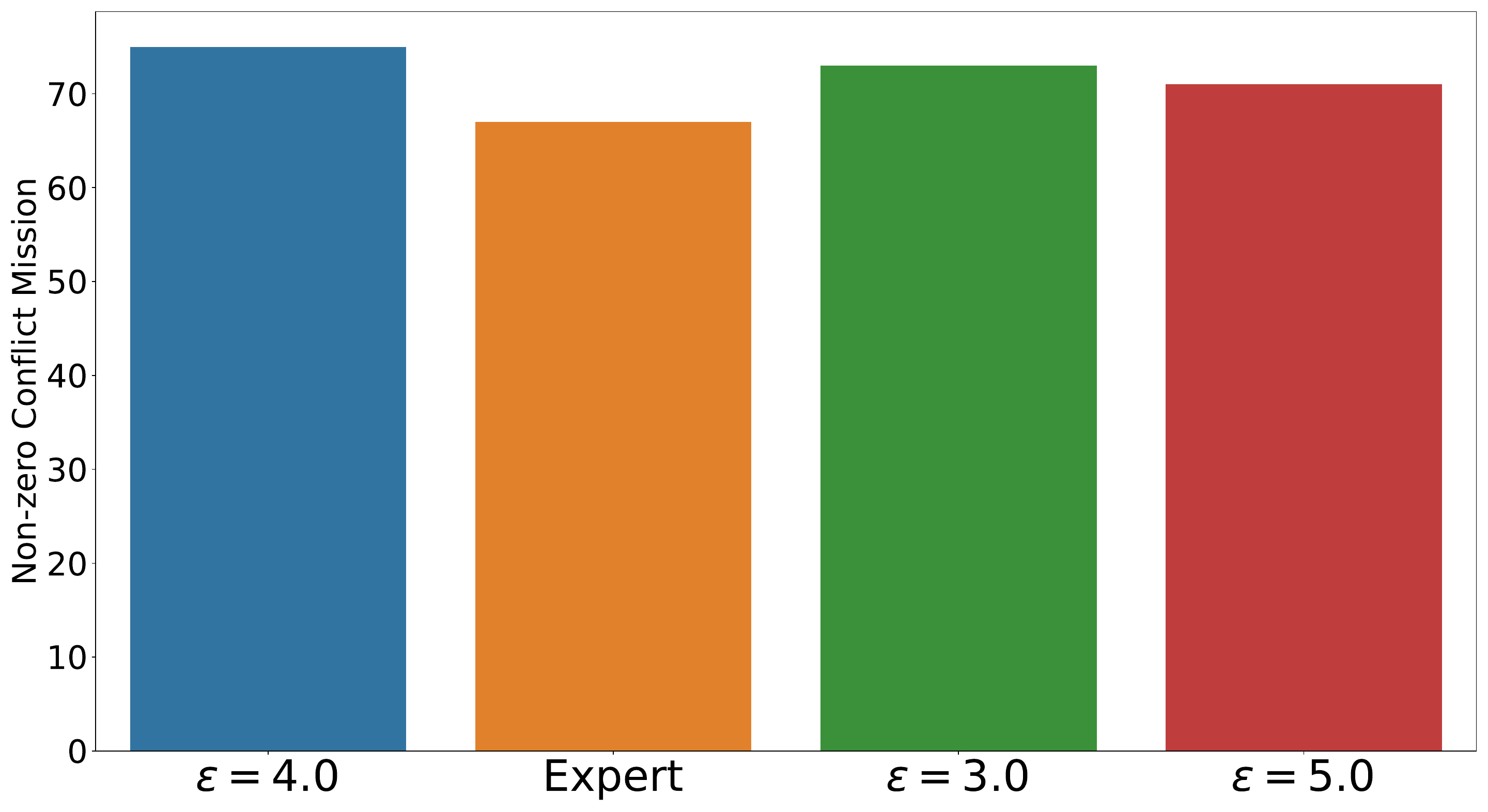}
		\caption{Comparing the total number of non zero conflicts obtained from expert policy and the curriculum update stopping condition against spoofing attack and 8 dynamic obstacles across 100 episodes.}
		\label{figure_ablation_non_zero}
	\end{figure}
	\section{Computational Costs for Curriculum Adaptation}\label{comp_cost}
	\begin{figure}[!t]
		\centering
		\includegraphics[scale=0.18]{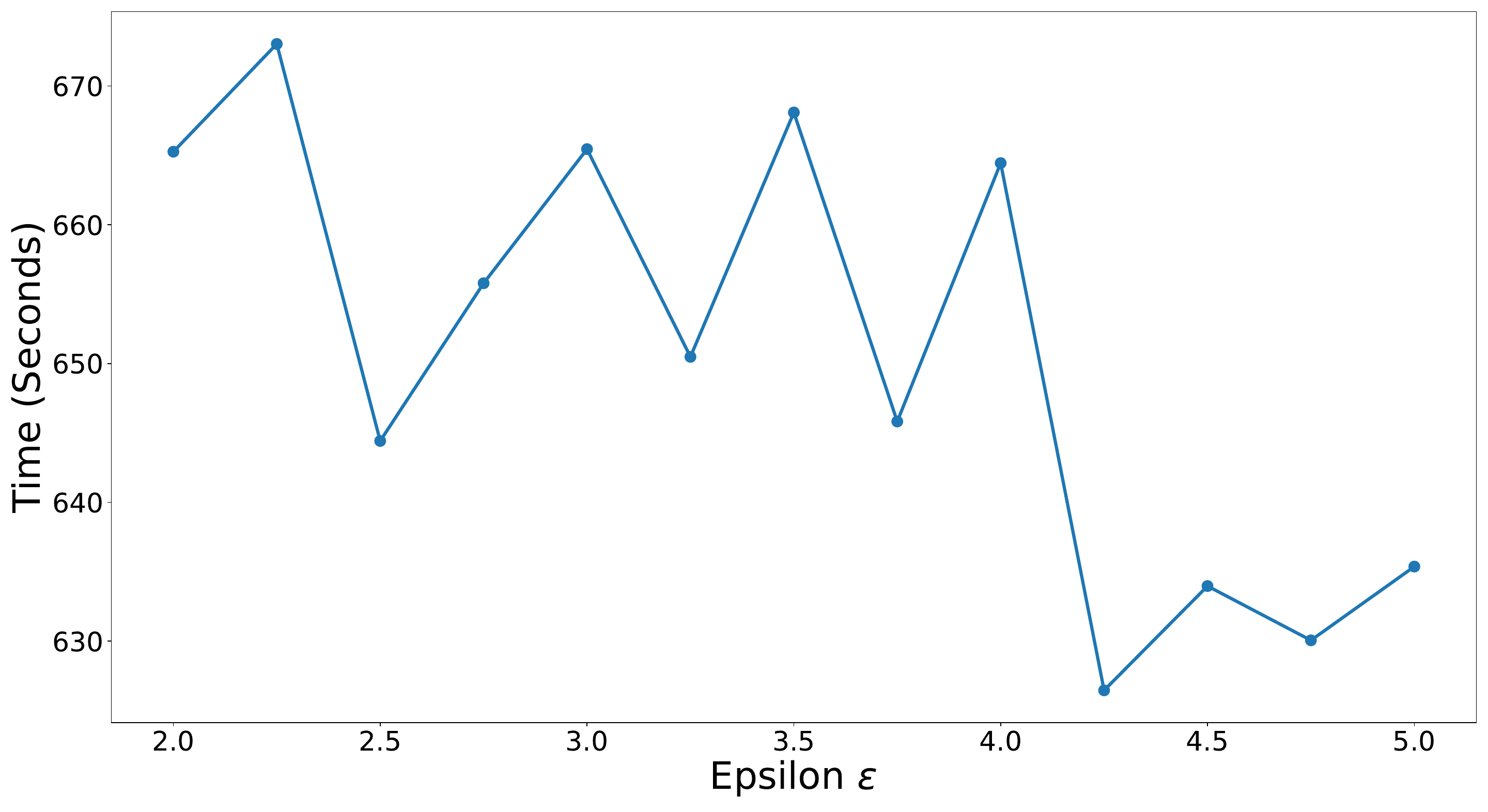}
		\caption{Computational time for each curriculum stage (at each $\epsilon$) across all episodes in seconds.}
		\label{figure_time_antifragile}
	\end{figure}
	\begin{figure}[!t]
		\centering
		\includegraphics[scale=0.18]{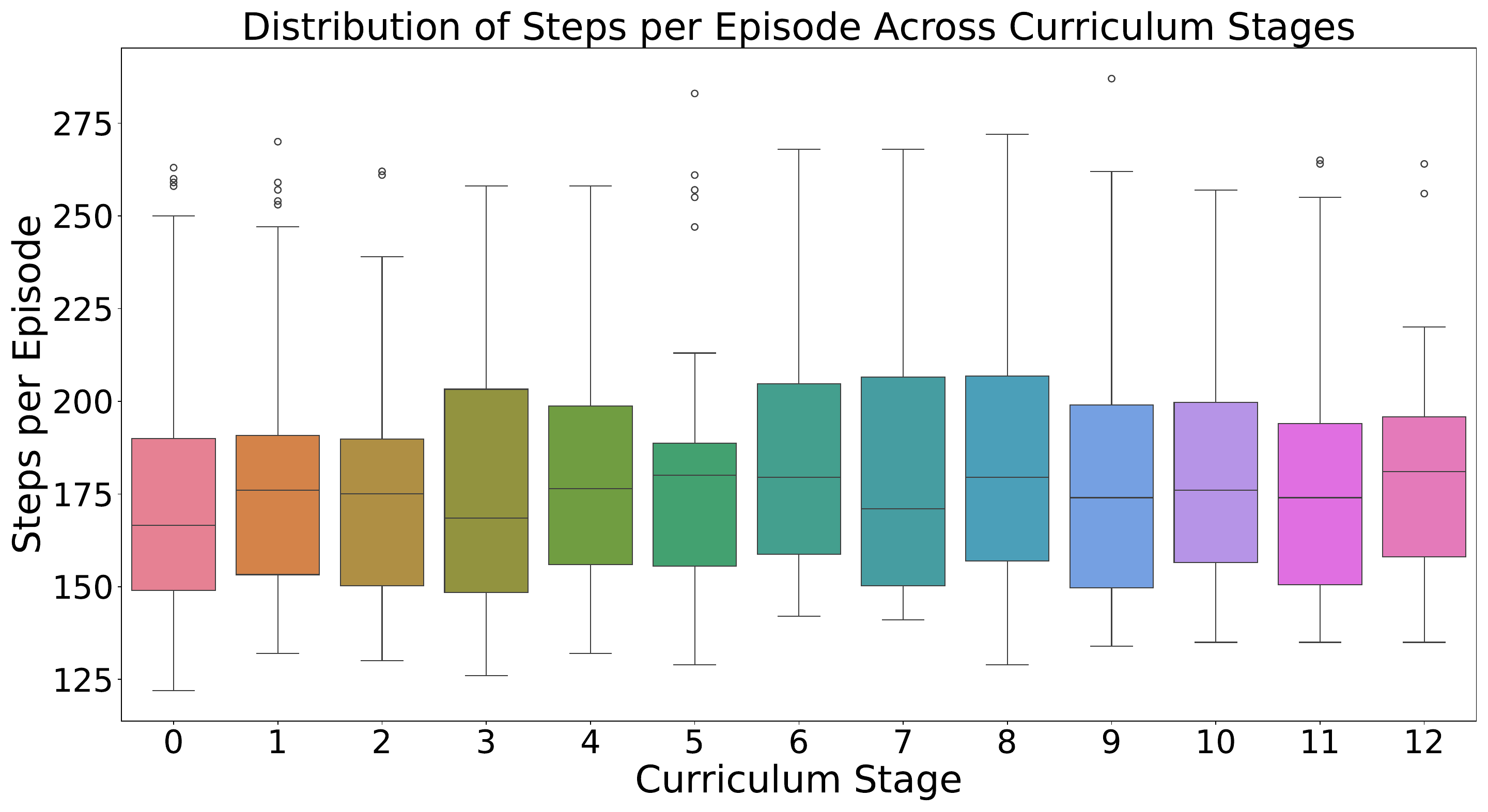}
		\caption{Boxplot of the steps the episodes for each curriculum stage.}
		\label{figure_steps_antifragile}
	\end{figure}
	\section{Baseline Algorithms} \label{bench_algo}
	Benchmarking is performed by evaluating the standard policy gradient algorithm, such as the DDPG, along with its robust variants. An approach involves an agent who intends to execute an action $\mathbf{a}$, but an alternative adversarial action $\bar{\mathbf{a}}$ is implemented with probability $\alpha$. Another variant, conceptually similar to the approach described in Section \ref{adv_ml}, employs soft probabilistic robust policy iteration (PR-PI) as proposed in \cite{tessler2019action}, rather than solving the mixed Nash equilibrium objective in (\ref{eq:11}). The final benchmarking strategy involves training the agent to optimize the rewards without directly modifying the policy during training. Instead, an adversarial buffer is utilized, allowing the agent to learn the optimal value function in the presence of adversarial states. A brief explanation of these algorithms is provided in the subsequent subsections.
	\subsection{Deep Deterministic Policy Gradient (DDPG)}
	The action robust RL is replaced by DDPG considering $\alpha = 0$. Hence, the critic target $y_{\textup{targ}}$ in (\ref{eq:15A}) is modified as follows,
	\begin{equation} \label{eq:63A}
		y_{\textup{targ}} = r_t + \gamma \left (1-d \right ) Q_{\phi_{\textup{targ}}} \left ( \Phi', \mu_{\theta}  \right ).
	\end{equation}
	Similarly the policy loss with $\alpha = 0$ to compute the policy parameter $\theta$, can be described as,
	\begin{equation} \label{eq:64A}
		\nabla_{\theta} \widehat{J \left ( \theta\right )} = \frac{1}{N}  \sum_{\Phi \in \mathcal{D}} \nabla_{\theta} \mu_\theta \left ( \Phi \right )  \nabla_a Q_{\phi} \left ( \Phi, a \right ).
	\end{equation}
	Hence, the min-max objective in (\ref{eq:11}) is replaced with a pure max objective which can be solved using traditional policy gradient methods as outlined in \cite{duan2016benchmarking}. 
	\subsection{Probabilistic Robust Markov Decision Process}
	According to \cite{tessler2019action}, PR-MDP is framed as a zero-sum game between an agent and an adversary. In this framework, an optimal probabilistic robust policy is defined with a probability $\alpha$, where the adversary assumes control and executes the most detrimental actions to account for potential system control limitations and undesirable outcomes. PR-MDP addresses stochastic perturbations within the policy space.  When we consider $\alpha \in \left [ 0, 1 \right ]$ and the 5-tuple MDP outlined in Section \label{adv_ml}, the probabilistic joint policy $\pi^{\textup{mix}}_{P,\alpha}$ is considered as 
	\begin{equation} \label{eq:65A}
		\pi^{\textup{mix}}_{P,\alpha} \left ( \left.\begin{matrix}
			\mathbf{a} 
		\end{matrix}\right| \Phi  \right )  \equiv \left ( 1 - \alpha \right ) \pi \left ( \left.\begin{matrix}
			\mathbf{a} 
		\end{matrix}\right| \Phi \right ) + \alpha \bar{\pi} \left ( \left.\begin{matrix}
			\mathbf{a} 
		\end{matrix}\right| \Phi \right ). 
	\end{equation}
	The $\pi^{\textup{mix}}_{P, \alpha}$ to learn the optimal value function for PR-MDP is obtained using PR-PI as shown in Algorithm 1 in \cite{tessler2019action}.
	\subsection{Noisy Robust Markov Decision Process}
	According to \cite{tessler2019action}, NR-MDP also models a zero-sum game between an agent and an adversary, but incorporates perturbations in the action space rather than in the policy space as in PR-MDP. Given $\pi$ and $\bar{\pi}$ as the policies of the agent and the adversary, respectively, their joint policy $\pi^{\textup{mix}}_{P,\alpha}$ as, 
	\begin{equation} \label{eq:66A}
		\pi^{\textup{mix}}_{N, \alpha} \left ( \left.\begin{matrix} \mathbf{a}  \end{matrix}\right| \Phi \right ) 
		= \mathbb{E}_{\mathbf{b} \sim \pi \left ( .| \Phi \right ),  \bar{\mathbf{b}} \sim \bar{\pi} \left ( .| \Phi \right )} \left [ \vec{\mathbf{1}}_{\mathbf{a} = \left ( 1 - \alpha \right ) \mathbf{b} + \alpha \bar{\mathbf{b}} } \right ].
	\end{equation}
	The NR-MDP is analogous to the Markov decision process (MDP) used to develop the ensembles, except for the addition of SGLD noise to the loss function,  to obtain the mixed Nash equilibrium solution for the objective. The solution to the min-max objective considered here is according to the soft PR-PI method as described in Algorithm 2 in \cite{tessler2019action}.
	\subsection{State Adversarial MDP}
	The SA-MDP formulation benchmarks the curriculum-adapted robust policy against a theoretically grounded robust learning framework that models adversarial perturbations on the state observations \cite{zhang2020robust}. In this setting, the observed state is no longer the true state $s$, but a perturbed version $\nu(s)$ constrained within a bounded perturbation set $B(s)$. The agent’s policy operates on the adversarially distorted observation $\nu(s)$, while the environment transitions according to the unperturbed state dynamics.
	SA-MDP defines a modified MDP tuple $(S, A, B, R, p, \gamma)$, where $B(s)$ restricts the adversarial perturbations. The core objective is to regularize the policy such that action distributions under perturbed and unperturbed states remain close. The policy regularization term minimizes the total variation or KL divergence between $\pi(a|s)$ and $\pi(a|\nu(s))$, thus promoting robustness to adversarial state shifts. This regularizer is incorporated into standard actor-critic algorithms like DDPG and PPO, using techniques such as convex relaxation or stochastic Langevin dynamics to solve the inner maximization over perturbation sets.
	Unlike methods that directly inject adversarial states into the replay buffer, SA-MDP systematically aligns the policy over the perturbation set, avoiding unstable training and catastrophic forgetting. In our experiments, we implement SA-MDP using convex relaxation as in \cite{zhang2020robust}, and benchmark it against curriculum-adapted robust RL to evaluate its capacity for generalization under severe adversarial conditions, including high-step PGD and spoofing attacks.

	\subsection{Meta-Adversarial Reinforcement Learning}
	We benchmark our curriculum-adapted robust policy against a meta-learning based approach that incorporates adaptive adversarial perturbations during policy training, as proposed in \cite{wang2021meta}. This method extends the model-agnostic meta-learning (MAML) framework by  integrating adversarial task variations into the inner-loop adaptation. Each task corresponds to a distinct adversarial perturbation level $\epsilon$, forming a curriculum of adversarial environments.
	In this framework, the agent is trained over a distribution of tasks $\mathcal{T} = \{\epsilon_1, \epsilon_2, ..., \epsilon_K\}$, where each task applies a bounded observation-space perturbation sampled via a generative adversarial network (GAN) conditioned on the current state. For a given task $\epsilon_k$, the inner loop optimizes the policy parameters $\theta$ on a perturbed environment using policy gradient updates. The updated policy $\theta_k'$ is then evaluated on the same task, and the meta-objective is constructed by aggregating performance across all tasks:
	\begin{equation}
		\min_{\theta} \sum_{\epsilon_k \in \mathcal{T}} \mathbb{E}_{\Phi \sim \mathcal{D}_{\epsilon_k}} \left[ \mathcal{L} \left( \pi_{\theta_k'} \right) \right], \quad
		\theta_k' = \theta - \hat{\beta} \nabla_\theta \mathcal{L} \left( \pi_\theta, \mathcal{D}_{\epsilon_k} \right),
	\end{equation}
	where $\hat{\beta}$ is the inner-loop learning rate, and $\mathcal{D}_{\epsilon_k}$ denotes the adversarially perturbed data distribution for task $\epsilon_k$.
	The adversarial perturbations are generated by a task-conditioned GAN, trained to minimize the expected Q-value of the perturbed state, thereby producing strong but feasible adversarial samples. Unlike static adversarial training methods, this approach dynamically adapts both the agent and the adversary, encouraging robustness through meta-level generalization. The final meta-update aggregates gradients across tasks to optimize the initial parameters $\theta$, enabling fast adaptation to unseen perturbations at test-time.
	In our implementation, we adopt a first-order approximation of MAML for computational efficiency and train the adversarial generator alongside the agent in alternating steps. This meta-adversarial RL framework is evaluated under both PGD and spoofing attacks to assess its generalization capability under severe and structured perturbations.
%	\FloatBarrier
	\section{\rev{Hyperparameter for Curriculum Adaptation and Robust RL Baselines}}
	\rev{For reproducibility, the principal training parameters used for the proposed method and benchmarking algorithms are summarized here. The expert action-robust policy used to initialize the curriculum adaptation was obtained using the same network architecture and training hyperparameters as our previous robust RL implementation \cite{panda2024action}. Unless otherwise specified, the DDPG-based methods use a minibatch size of (N=128), a maximum episode horizon of 500 steps, Gaussian exploration noise with standard deviation (0.3), and parameter updates every 50 environment steps. The baseline-specific parameters were retained according to their respective implementations. In particular, the SA-MDP/AutoLiRPA baseline uses an ($L_\infty$)-bounded perturbation radius of ($\epsilon=4.0$) and 40 training episodes. The adversarial meta-RL baseline considers five adversarial tasks uniformly distributed over ($\epsilon\in[0,5]$), with a meta-batch size of 3, five inner adaptation steps, 100 meta-training episodes, and a meta-learning rate of ($10^{-3}$).}
	
	\rev{For the proposed curriculum adaptation, the TD errors generated by the expert critic are computed separately and retained as the reference distribution for the corresponding curriculum stage. At each critic update, a minibatch of (N=128) transitions is sampled from the adversarial replay buffer and an equally sized sample is drawn from the stored expert TD-error values. The TD errors of the current critic are then evaluated for the sampled adversarial transitions, and the adaptation loss is computed as the empirical Wasserstein-1 distance between the expert and current TD-error minibatches,}
	\[
	\rev{
		\mathcal{L}_{\mathrm{crit}}
		=
		\frac{1}{N}
		W_1\!\left(
		\{\mathbf{TD}_i^{\mathrm{exp}}\}_{i=1}^{N},
		\{\mathbf{TD}_i^{\mathrm{ad}}\}_{i=1}^{N}
		\right).
	}
	\]
	\rev{The resulting loss is differentiated with respect to the current critic parameters while the expert TD-error samples are treated as the fixed reference for that curriculum stage. After completion of a stage, the TD errors generated by the adapted critic are retained as the expert reference for the next stage. Adversarial observations are generated using 10 projected-gradient steps, and the perturbation strength is progressively increased using a curriculum increment of ($\Delta\epsilon=0.25$). The hyperparameters for the proposed framework and the robust baselines used in the manuscript is summarized as follows.}
	
%	\noindent\rev{\textbf{Key Hyperparameters and Training Settings:}}
%	
	\begin{itemize}
		
		\item \rev{\textbf{Shared DDPG / Robust Backbone:}
			Batch size of 128, training horizon of 500,
			exploration noise of 0.3, and update interval of 50.}
		
		\item \rev{\textbf{Expert Action-Robust RL:}
			The architecture, optimizer, and training settings
			follow those reported in [5], with the adversarial
			mixing parameter $\alpha$ selected accordingly.}
		
		\item \rev{\textbf{Curriculum Adaptation:}
			Batch size of 128, 10 PGD steps,
			curriculum increment $\Delta\epsilon = 0.25$,
			and Wasserstein-1 ($W_1$) distance for
			TD-error distribution alignment.}
		
		\item \rev{\textbf{Adversarial Meta-RL:}
			Five tasks sampled over $[0,5]$,
			meta-batch size of 3, five inner adaptation steps,
			meta-learning rate of $10^{-3}$,
			and 100 training episodes.}
		
		\item \rev{\textbf{SA-MDP / AutoLiRPA:}
			$L_{\infty}$ perturbation bound of
			$\epsilon = 4.0$, CROWN-based bounds,
			and 100 training episodes.}
		
		\item \rev{\textbf{DDPG:}
			Shared DDPG training settings with
			adversarial mixing parameter $\alpha = 0$.}
		
		\item \rev{\textbf{PR-MDP / NR-MDP:}
			Shared DDPG backbone with method-specific
			adversarial mixing parameters.}
		
	\end{itemize}

	\appendices
\end{document}